# Confounding-Robust Fitted-Q-Iteration under Observed Markovian Marginals

David Bruns-Smith
Stanford University, causal@stanford.edu

Angela Zhou
University of Southern California, zhoua@usc.edu

**Abstract.** Offline reinforcement learning is important in domains such as medicine, economics, and e-commerce where online experimentation is costly, dangerous or unethical, and where the true model is unknown. However, most methods assume sequential ignorability, that all covariates used in the behavior policy's action decisions are observed. Though observational data likely has *unobserved confounders* (UC), in the big-data regime, these UCs are likely less informative than observed confounders, motivating sensitivity analysis. We study robust policy evaluation and policy optimization under a sensitivity model assuming that the *observed marginal transition probabilities are Markovian*. We introduce a test for this assumption, which we show is practically equivalent to *memoryless UCs*. Our test also informs how many lags to add to the state to handle higher-order UCs. We propose and analyze orthogonal robust fitted-Q-iteration, based on our derived loss function whose solution is the robust $Q$ function. Orthogonality reduces dependence on quantile estimation error. We provide sample complexity bounds, insights, and show effectiveness both in simulations and on real-world longitudinal healthcare data of treating sepsis. Our model of sequential unobserved confounders yields an online Markov decision process, rather than a partially observed Markov decision process: we illustrate how this can enable warm-starting optimistic reinforcement learning algorithms with valid robust bounds from observational data.

**Key words:** offline reinforcement learning, causal reinforcement learning, sequential decision-making under ambiguity

## 1. Introduction

Sequential decision-making problems in medicine, economics, and e-commerce require the use of historical observational data when online experimentation is costly, dangerous or unethical. Given the rise of big data, these observational datasets are increasingly large and widely available, with great potential to improve decisions based on personalizing treatments to those who most benefit. The recent literature on offline reinforcement learning (RL) has made extensive progress on evaluating and optimizing sequential decision rules given only historical datasets of observed trajectories. In particular, methods that target estimation of the Q function leveraging black-box regression, such as fitted-Q-evaluation and fitted-Q-iteration (FQE/FQI), have gained popularity due to their computational ease and scalability (Voloshin et al. 2019).

However, offline RL methods almost universally require that all covariates used to make the historical decisions are recorded in the observational dataset. Unfortunately, there are usually factors not observed in the data that jointly influence the historical decisions and the outcomes. The presence of these unobserved confounders introduces spurious correlation, biasing the estimates from offline RL algorithms, and potentially resulting in harmful policies.

Our goal is to learn optimal sequential decision policies that are robust to a potentially *restricted* extent of unobserved confounding in the observational dataset. To analyze the impacts of unobserved confounding on key MDP estimands, we build on sensitivity analysis techniques developed in the causal inference literature. Sensitivity models parameterize the strength of unobserved confounding via how it affects the probability of selection into treatment (Robins et al. 2000, Rosenbaum 2004, VanderWeele and Ding 2017). Choosing a proposed maximum level of confounding yields an ambiguity set for robust $Q$-functions. A practitioner can sweep the sensitivity parameter from no confounding to very strong confounding, and if the worst-case value of a proposed policy is consistently better than baseline, this provides good evidence for robustness. Typically strong confounding is quantified relative to observed covariates. For example, in lung cancer prevention treatments, it is very unlikely that there exists an unobserved variable more important than the patient's observed smoking status. We adopt the "marginal sensitivity model" (MSM) of Tan (2012), a variant of Rosenbaum's sensitivity model (Rosenbaum 2004), which has been widely used for offline single-timestep policy optimization (Aronow and Lee 2013, Miratrix et al. 2018, Zhao et al. 2019, Yadlowsky et al. 2018, Kallus et al. 2018, Kallus and Zhou 2020b).

Dynamic robust optimization over absolutely general unobserved confounding is difficult, and likely conservative. We begin with an assumption that the *observed marginal transition probabilities (i.e. over observed states alone) are Markovian* (i.e. conditionally independent of prior history given current state). In turn, this testable assumption places substantial restrictions on the underlying data-generating process on unobserved confounders: we show later on that when the underlying causal graph is the same from timestep to timestep, the only remaining unobserved confounding is so-called *memoryless*. Memoryless unobserved confounders $U_t$ cannot not have any direct causal influence from prior unobserved confounders; but they can be causally influenced by the current observed state $S_t$. This still permits autocorrelation and persistence in $U_t$; i.e. $U_t$ and $U_{t-1}$ can be statistically *correlated* with each other, due to state dependence; but not directly causally dependent. In stark contrast to general models of unobserved confounding that result in POMDPs, which are statistically and computationally harder to solve, under observed Markov marginals, we can solve for a confounding-robust optimal policy with robust MDPs. These bounds from sensitivity analysis remain informative as the horizon grows and therefore admit standard MDP formulations on observed states alone.

Importantly, often decision problems in operations may only be *higher*-order Markov, in which case adding lagged history information to the observed state can restore the Markovian property to it (Howard 1960). Analogously, adding lagged history to the state variable can generalize our framework to handle richer unobserved confounders: adding one lag can address Markovian $U_t$, and so on. We summarize this in our more general *k-order memoryless UC* assumption. This mirrors general strategies in operations to move from non-Markovian formulations to MDP by augmenting the state variable with (part of) the history (Howard 1960). Of course, a crucial question is validating our assumption of observed Markov marginals holds, or otherwise determining the appropriate number of lags. We also provide a heuristic conditional independence testing procedure to test whether the observed Markov property is satisfied for some number of lags. First, we give some detailed examples as to how unobserved confounders might arise in practice, whether memoryless or Markovian (within our framework by augmenting the state variable).

Example 1 (Memoryless UCs might be state-driven behavioral biases). Systematic behavioral biases in operational decision-making are often strongly explained by state information. In retail operations, Caro et al. (2010) found that managers typically used heuristic pricing strategies that focused on inventory run-out time; even after the rollout of a revenue-maximizing decision-support system, these *inventory state-driven behavioral biases* persisted. Managers might be particularly concerned about stockouts or rewarded for total sales that are not considered by central profit maximization, therefore increasing prices while also taking other unrecorded actions or adjusting in-person selling efforts, affecting future sales. Memoryless unobserved confounders can model systematic behavioral biases that are driven by differences in observed state – here, inventory/stockout salience. Su (2008) studies quantal-response models of bounded rationality in newsvendor decisions; such random errors would be examples of memoryless UCs. Kremer et al.

(2010) posit simple state-dependent behavioral biases like mean anchoring, demand chasing, or inventory error minimization; which depend on simple sufficient statistics like mean demand, prior demand or order quantities that could augment the observed state. Managerial deviations from algorithm-recommended prices can be driven by these state-dependent behavioral biases (Feng and Zhang 2017, Wachtel and Dexter 2010), also signaling other unobserved in-store actions (like promoting other items in-person) that affect profits.

Example 2 (MDPs in healthcare with simple state descriptions calibrated from observational data Markov Decision Processes are powerful tools in healthcare operations management for chronic and longitudinal conditions (Denton 2018). But Denton (2018) notes that the source data '.. reside largely in observational data sources, such as electronic health records, ... claims ..., and other forms of data that are collected routinely as part of the healthcare delivery process". This data is observational, not randomized, and hence vulnerable to unobserved confounders. Often researchers simplify the state space, in part due to the curse of dimensionality, further omitting potential *unobserved confounders* affecting both treatment selection and state evolution. For example, Zhang et al. (2014) calibrate a MDP for type-II diabetes control from administrative claims data, and their state definition is hbA1c discretized into 10 levels. For example, KDIGO 2022 clinical guidelines for diabetes management in chronic kidney disease (Navaneethan et al. 2023) highlight how kidney disease progression and status affect treatment choice and diabetes progression (Kumar et al. 2023). Treatment guidelines of Navaneethan et al. (2023) change based on eGFR, a standard clinical measure of kidney functioning, which may not be measured as standard practice for everyone.

These examples illustrate unobserved confounders that are strongly explained by observed state, with limited dependence on the prior history of unobserved confounders.

Researchers often only have access to longitudinal data that is vulnerable to the presence of unobserved confounders. The use of observational data for learning sepsis management policies to illustrate offline reinforcement learning (Raghu et al. 2017b, Komorowski et al. 2018) is another example in healthcare, with some recent controversies around (un)reliability of the data where unobserved confounders are of utmost concern (Gottesman et al. 2019)[1].

We develop a blueprint for practitioners to derive safe insights (whether via bounds on policy values or robust policies) from observational data. At the end of our paper, we revisit sepsis data from MIMIC-III. We deliberately design our method to hew closely to typical fitted-Q-evaluation/iteration methods previously used in the literature, namely based on a series of regressions. We walk through our approach, from analyzing Markovianity of observed transitions to determining the number of lags to add to the state, calibrating ambiguity sets, to developing robust value estimates and robust policies. Comparing robust vs. nominal value functions can provide insight or, even if not deployed, inform future investigation and data collection.

[1] More broadly, the FDA has recognized a growing need for methods that assess the "robustness and resilience of these [clinical decision support] algorithms to withstand changing clinical inputs and conditions" (FDA 2021).

Our paper proceeds as follows: In Section 2 we introduce the problem setup and our key assumptions and characterizations thereof. We anchor on the testable assumption that the *transitions between observed states and actions are Markovian*, and the *faithfulness* assumption from causal discovery. In Section 3 we introduce our method: we translate sensitivity models from causal inference restricting the strength of unobserved confounding to robust MDPs. The robust $Q$ function is a conditional expected-shortfall/conditional CVaR function: we estimate by regressing on a transformed orthogonal target that we introduce. Our regression-based method is in line with popular fitted-Q evaluation/iteration (Fu et al. 2021, Le et al. 2019) paradigms. Section 4 discusses how to instantiate the method by connecting underlying models on UCs to observed data, such as checking assumptions and our testing procedure. Section 5 develops provable guarantees for our method, where the key benefit of orthogonality is in reducing dependence of estimating the conditional expected shortfall/conditional CVaR on the estimation of the conditional quantile function. Without our orthogonal adjustment, the conditional quantile would need to be estimated at fast parametric $O_p(n^{-1/2})$ rates to ensure $O_p(n^{-1/2})$ convergence of policy evaluation and optimization; but with our orthogonal adjustment, the conditional quantile only needs to be estimated at a slower $o_p(n^{-1/4})$ rate. Section 6 contains our empirical experiments: 1) in simulations, we demonstrate the benefits of our orthogonal approach (orders of magnitude reduction in MSE), 2) in simulations, we demonstrate improvements in policy optimization, how adding lags to the state can handle UCs and robustness of our method otherwise, and 3) a complete end-to-end real-world case study using electronic medical records from the MIMIC-III critical care database for the sepsis management task. Our case study demonstrates a blueprint for practitioners from validating assumptions, to learning robust policies with insights that line with high-level clinical findings (Silversides et al. 2017). Section 7 includes a warm-starting extension that demonstrates the significance of the marginal MDP characterization: it enables warmstarting online learning with bounds from confounded data.

### 1.1. Related Work

We first discuss offline reinforcement learning in general, and other approaches for unobserved confounders besides ours based on robustness. Then we discuss other topics such as orthogonalized estimation, robust Markov decision processes, and robust offline reinforcement learning; before summarizing how our work is at the intersection of and relates to these areas.

*Policy learning with unobserved confounders in single-timestep and sequential settings.* The rapidly growing literature on offline reinforcement learning with unobserved confounders can broadly be divided into three categories. We briefly discuss central differences from our approach to these three broad groups and include an expanded discussion in the appendix. First, some work assumes point identification is available via instrumental variables (Wang et al. 2021)/latent variable models (Bennett and Kallus 2019)/front-door identification (Shi et al. 2022b). Although point identification is nice if available, sensitivity analysis can be used when assumptions of point identification (instrumental-variables, front-door adjustment) *are not true*, as may be the case in practice. Second, a growing literature considers proximal causal inference in POMDPs from temporal structure (Tennenholtz et al. 2019, Bennett et al. 2021, Uehara et al. 2022, Shi et al. 2022a) or additional proxies (Miao et al. 2022). Proximal causal inference imposes additional (unverifiable) completeness assumptions on the latent variable structure and is a statistically challenging ill-posed inverse problem. Furthermore, we study a more restricted model of memoryless unobserved confounders that has important qualitative differences from generically unstructured POMDPs: the online counterpart is a marginal MDP, enabling warmstarting approaches. Third, a few approaches compute no-information partial identification (PI) bounds based only on the structure of probability distributions and no more (Han 2022, Chen and Zhang 2021). These can generally be much more conservative than sensitivity analysis, which relaxes strong assumptions.

Overall, developing a *variety* of identification approaches further is crucial both for analysts to use appropriate estimators/bounds, and methodologically to support falsifiability analyses. Other works include (Fu et al. 2022, Liao et al. 2021, Saghafian 2021). In our work, we consider the marginal sensitivity model. Extending to other sensitivity analysis models may also be of interest (Robins et al. 2000, Scharfstein et al. 2018, Yang and Lok 2018, Bonvini and Kennedy 2021, Bonvini et al. 2022, Scharfstein et al. 2021, Chernozhukov et al. 2022).[2]

[2] Both the state-action conditional uncertainty sets and the assumption of memoryless unobserved confounders are particularly crucial in granting state-action rectangularity (for binary treatments), and avoiding decision-theoretic issues with time-inconsistent preferences in multi-stage robust optimization (Delage and Iancu 2015). On the other hand, the exact functional form (subject to these structural assumptions) could readily be modified.

*Off-policy evaluation in offline reinforcement learning* An extensive line of work on off-policy evaluation (Jiang and Li 2016, Thomas et al. 2015, Liu et al. 2018, Tang et al. 2019) in offline reinforcement learning studies estimating the policy value of a posited evaluation policy when only data from the behavior policy is available. Most of this literature, implicitly or explicitly, assumes sequential ignorability/sequential unconfoundedness. Methods for policy optimization are also different in the offline setting than in the online setting. Options include direct policy search (which is quite sensitive to functional specification of the optimal policy) (Zhao et al. 2015), off-policy policy gradients which are either statistically noisy (Imani et al. 2018) or statistically debiased but computationally inefficient (Kallus and Uehara 2020b), or fitted-Q-iteration (Le et al. 2019, Ernst et al. 2006). Of these, fitted-Q-iteration's ease of use and scalability make it a popular choice in practice. It is also theoretically well-studied (Duan et al. 2021). A marginal MDP also appears in Kallus and Zhou (2022) but in a different context, without unobserved confounders.

*Orthogonal estimation.* Double/debiased machine learning seeks so-called Neyman-orthogonalized estimators of statistical functionals so that the Gateaux derivative of the statistical functional with respect to nuisance estimators is 0 (Newey 1994, Chernozhukov et al. 2018, Foster and Syrgkanis 2019). Nuisance estimators are intermediate regression steps (i.e. the conditional quantile) that are not the actual target function of interest (i.e. the robust $Q$ function). Orthogonalized estimation reduces the dependence of the statistical estimator on the estimation rate of the nuisance estimator. See Kennedy (2022) for tutorial discussion and Jordan et al. (2022) for a computationally-minded tutorial. There is extensive literature on double robustness/semiparametric estimation in the longitudinal and MDP setting (Laan and Robins 2003, Robins et al. 2000, Orellana et al. 2010, Bibaut et al. 2019, Kallus and Uehara 2020a, Singh and Syrgkanis 2022, Lewis and Syrgkanis 2020).

Recent work studies orthogonality/efficiency for partial identification and in other sensitivity models than the one here (Bonvini and Kennedy 2021, Bonvini et al. 2022, Scharfstein et al. 2021, Chernozhukov et al. 2022). More specifically, Semenova (2017), Olma (2021) study orthogonal partial identification or conditional expected shortfall, and we directly apply the orthogonal moment given in Olma (2021). Other works study orthogonality under the Rosenbaum model (Yadlowsky et al. 2018) with limited extension to the sequential setting (Namkoong et al. 2020) via a *single-worst timestep* restriction. Other works study related variants of CVaR estimands (Jeong and Namkoong 2020, Dorn and Guo 2022). We discuss more closely in Section 3.4. Overall, in contrast to these works, for policy optimization we require the entire robust Q function, which is important for rectangularity, motivating different estimation.

*Robust Markov decision processes and offline reinforcement learning.* Elsewhere, in the robust Markov-decision process framework (Nilim and El Ghaoui 2005), the challenge of *rectangularity* has been classically recognized as an obstacle to efficient algorithms although special models may admit non-rectangularity and computational tractability (Goyal and Grand-Clement 2022). Many recent algorithmic improvements are tailored for special structure of ambiguity sets (Behzadian et al. 2021, Ho et al. 2021). Recent work in distributionally robust RL (Zhou et al. 2021, Wang et al. 2023b, 2024) studies sample complexity in the offline setting under a generative model for tabular MDPs (Yang et al. 2022) or linear function approximation (Ma et al. 2022), or the online setting (Wang et al. 2023a). Our work relates sensitivity analysis in sequential causal inference to this line of literature and focuses on algorithms for policy evaluation based on a robust fitted-Q-iteration. Panaganti et al. (2022) also proposes a robust fitted-Q-iteration algorithm; we consider a different uncertainty set from their $\ell_1$ set, and further introduce orthogonalization. We focus on the conditional expected shortfall (conditional CVaR); prior works study marginal CVaR and variants (Lobo et al. 2020, Chow et al. 2015).

Importantly, robust RL doesn't directly handle the problem of causal *ambiguity*. It's more plausible and credible for practitioners to analyze restrictions on the underlying selection process from unobserved confounders. That is where we begin. The alternative, choosing the ambiguity set on transition probabilities directly (i.e., beginning in robust RL), would "assume the consequent". Although the "pessimism" principle in offline reinforcement learning is well-studied as a tool to relax strong concentrability assumptions (Jin et al. 2021), it relies on robustness sets motivated by statistical uncertainty, calibrated to probabilistic confidence levels. While we analyze the conditional CVaR reformulation, the resulting quantile level depends on instead on the *analyst-specified ambiguity*.

## 2. Problem Setup and Characterization

### 2.1. Problem Setup with Unobserved State

We consider a finite-horizon MDP on a full-information state space, summarized as the tuple $\mathcal{M} = (\mathcal{S} \times \mathcal{U}, \mathcal{A}, r, P, \chi, T)$. We let the product state space of observed and unobserved confounders, $\mathcal{S}, \mathcal{U}$, be continuous, and assume the action space $\mathcal{A}$ is finite (but possibly very large). We consider a finite horizon of length $T$, with time periods $t = 0, \ldots, T-1$. We provide an extension to the discounted infinite-horizon case in the appendix. Let $\Delta(X)$ denote probability measures on a set $X$. The set of time $t$ transition functions $P$ is defined with elements $P_t : \mathcal{S} \times \mathcal{U} \times \mathcal{A} \to \Delta(\mathcal{S} \times \mathcal{U})$;

$r$ denotes the set of time $t$ reward maps[3], $r_t : \mathcal{S} \times \mathcal{A} \times \mathcal{S} \to \mathbb{R}$; the initial state distribution is $\chi \in \Delta(\mathcal{S} \times \mathcal{U})$. A policy, $\pi$, is a set of maps $\pi_t : \mathcal{S} \times \mathcal{U} \to \Delta(\mathcal{A})$, where $\pi_t(a \mid s, u)$ is the probability of taking actions given states and unobserved confounders. The MDP dynamics under policy $\pi$ induce the random variables, $S_0, U_0 \sim \chi$, and for all $t$, $A_t \sim \pi_t(\cdot \mid S_t, U_t)$, $S_{t+1}, U_{t+1} \sim P_t(\cdot \mid S_t, U_t, A_t), R_t = r_t(S_t, A_t, S_{t+1})$. We will use $P_\pi$ and $\mathbb{E}_\pi$ to denote the joint probabilities (and expectations thereof) of the random variables $S_t, U_t, A_t, \forall t$ in the underlying MDP running policy $\pi$.

We consider a confounded offline setting: data is collected via an arbitrary behavior policy $\pi^b$ that potentially depends on $U_t$, but in the resulting data set, the $\mathcal{U}$ part of the state space is unobserved. That is, although the underlying dynamics follow a standard Markov decision process generating the history $\{(S_t^{(i)}, U_t^{(i)}, A_t^{(i)}, S_{t+1}^{(i)})_{t=0}^{T-1}\}_{i=1}^n$, the observational dataset omits the unobserved confounder. We observe $n$ trajectories, $\mathcal{D}_{obs} := \{(S_t^{(i)}, A_t^{(i)}, S_{t+1}^{(i)})_{t=0}^{T-1}\}_{i=1}^n$. When referring to the distribution under the behavior policy $\pi^b$, we will write $P_{\text{obs}}$, $\mathbb{E}_{\text{obs}}$ to emphasize the distribution of variables in the observational dataset.

As in standard offline RL, we study policy evaluation and optimization for target policies $\pi^e$ using data collected under $\pi^b$. In our confounded setting, we only consider $\pi^e$ that are a function of the observed state $S_t$ alone. Our objects of interest will be the observed state Q function and value function for the target policy $\pi^e$:

$$Q_t^{\pi^e}(s,a) := \mathbb{E}_{\pi^e}\left[\textstyle\sum_{j=t}^{T-1} R_j | S_t = s, A_t = a\right], \;\; V_t^{\pi^e}(s) := \mathbb{E}_{\pi^e}[Q_t^{\pi^e}(S_t, A_t) | S_t = s].$$

We would like to find a policy $\pi^e$ that is a function of the observed state alone, maximizing $V_0^{\pi^e}$. The key challenge is that with unobserved confounders, we cannot directly evaluate the true expectations above due to biased estimation.

### 2.2. The Observed-State Markov Property and Marginal MDP

First we establish that under observed Markovian marginals, the online decision problem is an MDP over just $S_t$ and $A_t$ (instead of a much more difficult generic POMDP), although unobserved confounding implies we don't know the true marginal transition probabilities. This enables us to focus on estimating robust $Q$ functions later on. We adopt the following central assumption:

[3] Note, we specify the reward as a function of only observables, $S_t, A_t, S_{t+1}$. This is essentially without loss of generality as we illustrate in Section 3.5.

Assumption 1 **(Observed-State Markov Property).** *Let* $H_t := (S_{t-1}, A_{t-1}, ..., S_0, A_0)$ *be the history of observed variables before time* $t$*. Then for all* $s, a, h, t$*:*

$$P_{obs}(S_{t+1}|S_t = s, A_t = a, H_t = h) = P_{obs}(S_{t+1}|S_t = s, A_t = a),$$
$$P_{obs}(A_t|S_t = s, H_t = h) = P_{obs}(A_t|S_t = s).$$

This condition is testable from observables as we discuss in Section 4.1. Assumption 1 alone is insufficient, because it holds on the observed data collected under policy $\pi^b$, but potentially not under other policies (such as evaluation or optimized policies). It is possible to construct adversarial policies $\pi^b$ such that the observed-state Markov property holds for $\pi^b$ but not for other policies. We require an assumption from graphical causal inference (Spirtes et al. 2000, Pearl et al. 2016) called *faithfulness* to link observed conditional independence with the underlying causal structure, ruling out such adversarially chosen $\pi^b$. Informally, faithfulness asserts that variables are probabilistically independent *only* if they are causally independent in the underlying causal graph, rather than circumstantial cancellation of parameters. The graphical language of d-connecting and d-separated paths (Pearl et al. 2016) makes this precise; we keep our discussion in the main text intuitive, see the Appendix for the full graphical causal inference framework.

Assumption 2 **(Faithfulness, Informal).** *Two random variables* $X_1$ *and* $X_2$ *in the underlying MDP are conditionally independent given a set of variables* $X_S$ *if and only if there are no unblocked backdoor paths from* $X_1$ *to* $X_2$ *given* $X_S$*.*

Faithfulness assumes that if variables are not directly or indirectly causally connected, then there shouldn't be any observed correlations between them that aren't already explained by other causal relationships. It is a relatively weak technical condition that is necessary in the causal discovery literature (Pearl et al. 2016, Spirtes et al. 2000).

A faithfulness violation would require exact cancellation of causal relationships and has measure zero (Pearl 2009), so it's generally expected to hold. To give some concrete examples of a faithfulness violation, consider an A/B test of a marketing campaign that increases direct revenues by \$10$k$ but cannibalizes the *exact* same revenue from brick-and-mortar channels, reducing brick-and-mortar revenues by \$10$k$. If the only observed outcome data is the *aggregate* net profit of \$0, without finer-grained data on sales, naively assessing the effect of the marketing campaign on profits incorrectly concludes no causal effects on profits, due to the faithfulness violation (exact cancellation of causal relationships). If this sounds like a knife-edge situation to you, that's because it is – and the faithfulness assumption protects against these *exact* cancellations that are unlikely to show up in real-world, non-homeostatic systems.

Assumptions 1 and 2 *together* imply that the online decision problem for $\pi^e$ depending only on observed state is a MDP, rather than a POMDP, ensuring the validity of our robust MDP approach.

PROPOSITION 1 **(Marginal MDP).** *Let $\chi^m \in \Delta(\mathcal{S})$ be the marginal distribution of $S_0$ under $\chi$. Given Assumptions 1 and 2, there exists transition probabilities $P^m_t : \mathcal{S}\times\mathcal{A} \to \Delta(\mathcal{S})$ on the observed state such that for any $\pi^e$ that does not depend on $U_t$, the full information MDP running the policy $\pi^e$ is equivalent to the marginal MDP, $(\mathcal{S}, \mathcal{A}, r, P^m, \chi^m, T)$, running the policy $\pi^e$. That is, in both the underlying and marginal MDPs, $S_0 \sim \chi^m, A_t \sim \pi^e_t(\cdot|S_t), S_{t+1} \sim P^m_t(\cdot|S_t, A_t), R_t = r_t(S_t, A_t, S_{t+1})$.*

Proposition 1 only guarantees the existence of such a true marginal transition probability $P^m_t$, but unobserved confounding implies that transition probabilities estimated from observed confounded data alone will be biased, and therefore different. *Both* Assumptions 1 and 2 are required to conclude the marginal MDP from properties of the observed data distribution alone. For now, we state the conclusion, deferring the analysis to Section 4.

(Informal) Given Assumptions 1 and 2, for all policies $\pi^e$ that do not depend on $U_t$, the observed-state Markov property holds for $\pi^e$.

### 2.3. An Important Case: Memoryless Confounders

An important case of our general setting is that of "memoryless unobserved unconfounders." Some version of the following condition is used in both Kallus and Zhou (2020a), Bruns-Smith (2021):

DEFINITION 1 (MEMORYLESS UNOBSERVED CONFOUNDERS). The full-information MDP has memoryless unobserved confounders if $U_t$ is independent of $S_{t-1}, U_{t-1}, A_{t-1}$ given $S_t$.

Memoryless confounding does not preclude the unobserved state from being strongly auto-correlated, but dependence in $U_t$ over time has to be fully mediated by $S_t$. We find that Assumptions 1 and 2 are, for practical purposes, nearly equivalent to memoryless confounding.

PROPOSITION 2.

1. *Assuming memoryless confounding alone (without Assumptions 1 and 2) is a sufficient condition for the results on the Marginal MDP (Proposition 1), offline confounding (Theorem 1), and all results in Section 3 and Section 5.*
2. *Assume that there is non-trivial confounding in every period, i.e. for all t, $S_{t+1}$ is not independent of $U_t$ conditional on $S_t, A_t$. Then, Assumptions 1 and 2 imply memoryless confounding.*

However, a variety of non-memoryless edge cases exist in our general setting (complicating the proofs), especially when the transitions are time-inhomogeneous — we give an exhaustive characterization in Section B. Many other works assume memoryless UCs (Wang et al. 2022, Shi et al. 2024, Bennett et al. 2024, Fu et al. 2022, Kausik et al. 2024, Xu et al. 2023b) *because* they admit observed Markov marginals; here we further characterize the exact relationship.

Though memoryless confounding can appear restrictive, our framework fully allows higher-order generalizations. A $k$th order memoryless unobserved confounder allows $U_t$ to additionally depend on its $k-1$ prior values only. Thus adding $k-1$ lagged history observations to the state recovers Markovian transitions (Assumption 1) on the augmented state space. When $k=1$ this is equivalent to Definition 1. If $k=2$, augmenting the state with the prior state/action enables robustness to Markovian unobserved confounders.

Definition 2 ($k$-order memoryless Unobserved Confounders). The full-information MDP has $k$-order memoryless unobserved confounders if $U_t$ is independent of $\{S_{\tilde{t}}, U_{\tilde{t}}, A_{\tilde{t}}\}_{\tilde{t} \leq t-k}$ given $S_t$ and $\{S_{t'}, U_{t'}, A_{t'}\}_{t'=t-k+1}^{t-1}$.

We summarize some settings where memoryless or $k$-order memoryless confounding might hold (although we emphasize Assumption 1 is a testable condition).

- Measurement error, reverse measurement error, or mixed measurement scopes: For example, consider the perspective of continuous glucose monitoring: these devices provide continuous noisy measurements of blood glucose ($S_t$) to patients who may occasionally take costly finger-stick measurements of glucose ($U_t$) and correspondingly adjust medication dosing which affects future blood glucose dynamics. Measurement errors between CGM and fingersticks are idiosyncratic or $S_t$-dependent (Clarke et al. 2008). The CGM monitor does not see the patient's measurement of $U_t$ but may try to improve dosing recommendations from historical data.
- Memoryless UCs for MDPs with exogenous arrivals: Many sequential decision problems in operations are driven by contextual exogenous arrival processes, i.e. transitions factorize into an unknown single-timestep random quantity (e.g. a demand observation) and a known stateful system transition function (e.g. inventory levels decrease if a sale is made).[4] Exogenous arrivals imply memoryless UCs in terms of additional contextual information associated with *each* arrival, independent of those of other arrivals. For example, in ridesharing systems, *exogenous customer arrivals* are attached to *personally identifiable information (PII)* such as geolocation, device data, or even third-party marketing data. This PII can affect historical system decisions (Uber Privacy Center), but may be unavailable in downstream analysis due to privacy or licensing restrictions.

[4] When actions affect the random quantity, this is stateful single-timestep off-policy evaluation (rather than full MDP) (Kallus and Zhou 2022); when actions don't affect the random quantity, Sinclair et al. (2023) calls these "MDPs with exogenous inputs".

We next introduce our method given these assumptions. We devote Section 4 afterwards to connect these model assumptions to data by testing for Assumption 1.

# 3. Method

## 3.1. Bias Characterization from Unobserved Confounding

So far, we established via Proposition 1 that the oracle decision problem remains a Markov decision process under our Assumptions 1 and 2. However, it is not possible to get unbiased estimates of the true marginal transition probabilities given data collected under $\pi^b$ when $U_t$ is unobserved. We now characterize the exact bias from confounding for estimating conditional expectations in the online marginal MDP, using the observational data distribution.

THEOREM 1 **(Confounding for Regression).** *Define the marginal behavior policy,* $\pi_t^b(a|s) :=$ $P_{obs}(A_t = a|S_t = s)$. *Let* $f(s, a, s')$ *be any function. Given Assumptions 1 and 2, for all* $s, a, t$,

$$\mathbb{E}_{P_t^m}\left[f(S_t, A_t, S_{t+1})|S_t = s, A_t = a\right] = \mathbb{E}_{obs}\left[\frac{\pi_t^b(A_t|S_t)}{\pi_t^b(A_t|S_t, U_t)} f(S_t, A_t, S_{t+1}) \mid S_t = s, A_t = a\right].$$

As a corollary, applying Theorem 1 with $f$ as the indicator for the next state bounds the bias between confounded observed-state transitions and the true marginal transitions. [az: should this be reversed order?] Define $\text{Bias}_t(s', s, a) := P_t^m(S_{t+1} = s'|S_t = s, A_t = a) - P_{\text{obs}}(S_{t+1} = s'|S_t = s, A_t = a)$. Given Assumptions 1 and 2, for all $s, a, s', t$:

$$\text{Bias}_t(s', s, a) = \mathbb{E}_{\text{obs}}\left[\left(1 - \frac{\pi_t^b(A_t|S_t)}{\pi_t^b(A_t|S_t, U_t)}\right)\mathbb{I}\{S_{t+1} = s'\} \mid S_t = s, A_t = a\right].$$

Note that the bias is potentially unbounded since $U_t$ is unobserved; we next discuss deriving robust MDPs from restrictions on the strength of unobserved confounding.

## 3.2. Sensitivity analysis to robust $Q$ function

We approach robustness to unobserved confounders following the sensitivity analysis literature from causal inference. We begin with a commonly-used sensitivity model (Tan 2012):

ASSUMPTION 3 **(Marginal Sensitivity Model).** *There exists* $\Lambda$ *such that* $\forall t, s \in \mathcal{S}, u \in \mathcal{U}, a \in \mathcal{A}$,

$$\Lambda^{-1} \leq \left(\frac{\pi_t^b(a \mid s, u)}{1 - \pi_t^b(a \mid s, u)}\right) / \left(\frac{\pi_t^b(a \mid s)}{1 - \pi_t^b(a \mid s)}\right) \leq \Lambda. \tag{1}$$

Under Assumption 3, we will derive upper and lower bounds on the value function for any given policy and develop a corresponding robust policy optimization. However, the sensitivity parameter $\Lambda$ is unknown and cannot be identified directly from the data; we discuss *how* to choose $\Lambda$ by calibrating from observed covariates in Section 4.2.

Next we introduce our key estimands – the robust Q and value functions. Assumption 3 implies an uncertainty set for the true observed-state transition probabilities $P_t(s'|s,a)$. First, note that Assumption 3 implies the following bounds on the unobserved ratio from Theorem 1:

$$\alpha_t(S_t,A_t) \leq \frac{\pi_t^b(A_t|S_t)}{\pi_t^b(A_t|S_t,U_t)} \leq \beta_t(S_t,A_t) \tag{2}$$

where $\alpha_t(S_t,A_t) := \pi_t^b(A_t|S_t)+\Lambda^{-1}(1-\pi_t^b(A_t|S_t))$ and $\beta_t(S_t,A_t) := \pi_t^b(A_t|S_t)+\Lambda(1-\pi_t^b(A_t|S_t))$. Though Assumption 3 holds on the behavior policy, combining (2) with Theorem 1 gives the following proposition: optimizing $\pi_b$ over the MSM ambiguity set is equivalent to optimizing transition probabilities over a corresponding interval in the next proposition.

Proposition 3. *Define the set:*

$$\mathcal{P}_t^{s,a} := \left\{ \bar{P}_t(\cdot \mid s,a) \colon \alpha_t(s,a) \leq \frac{\bar{P}(s_{t+1} \mid s,a)}{P_{obs}(s_{t+1} \mid s,a)} \leq \beta_t(s,a), \forall s_{t+1};\ \int \bar{P}_t(s_{t+1} \mid s,a) ds_{t+1} = 1, \right\}$$

*and let $\mathcal{P}_t$ be the set of transition probabilities for all $s,a$ defined as the product set over the $\mathcal{P}_t^{s,a}$. Under Assumptions 1, 2, and 3, we have that $P_t^m \in \mathcal{P}_t$.*

Obtaining the worst-case values of $Q_t^{\pi^e}$ and $V_t^{\pi^e}$ over transition probabilities in the uncertainty set, $\bar{P}_t \in \mathcal{P}_t$ is therefore an (s,a)-rectangular Robust Markov decision process (RMDP) problem (Iyengar 2005). Sharpness holds for $|\mathcal{A}| = 2$ actions but for higher-cardinality actions, this is a relaxation.[5]. Denote the robust Q and value functions $\bar{Q}_t^{\pi^e}$ and $\bar{V}_t^{\pi^e}$. Results of Iyengar (2005) allow us to define the following operators:

Definition 3 (Robust Bellman Operators). For any function $Q : \mathcal{S} \times \mathcal{A} \to \mathbb{R}$,

$$(\bar{\mathcal{T}}_t^{\pi^e} Q)(s,a) := \inf_{\bar{P}_t \in \mathcal{P}_t} \mathbb{E}_{\bar{P}_t}[R_t + Q(S_{t+1}, \pi_{t+1}^e)|S_t = s, A_t = a], \tag{3}$$

$$(\bar{\mathcal{T}}_t^{*} Q)(s,a) := \inf_{\bar{P}_t \in \mathcal{P}_t} \mathbb{E}_{\bar{P}_t}[R_t + \max_{A'}\{Q(S_{t+1}, A')\}|S_t = s, A_t = a]. \tag{4}$$

[5] We comment on the tightness of the robust operator. For a fixed $s$ and $a$, $\mathcal{P}_t^{s,a}$ is exactly the set of transition probabilities consistent with Assumption 3 and the observational data distribution — see Kallus and Zhou (2020a) and Bruns-Smith (2021) for a derivation. However the $s,a$-rectangular product set $\mathcal{P}_t$ does not explicitly enforce the density constraint on $\pi_t^b$ across actions, and is therefore potentially loose. In the special case where there are only two actions, Dorn et al. (2021) show that the different minima over $\mathcal{P}_t^{s,a}$ across actions are simultaneously achievable, and thus the robust bounds are tight and we get equalities in Proposition 4. For $|\mathcal{A}| > 2$, the infimum in eq. (3) is not generally simultaneously realizable (see Section 7.2 for a counter-example). Nonetheless, the robust Bellman operator corresponds to an $s,a$-rectangular relaxation of the RMDP, Proposition 4 will hold with lower bounds instead of equalities, and our results are still guaranteed to be robust.

Proposition 4 (**Robust Bellman Equation**). *Let $|\mathcal{A}| = 2$ and let Assumptions 1, 2, and 3 hold. Then applying the results in Iyengar (2005), gives that for any $\pi$, $\bar{Q}_t^\pi(s,a) = \bar{\mathcal{T}}_t^\pi \bar{Q}_{t+1}^\pi(s,a)$ and $\bar{V}_t^\pi(s) = \mathbb{E}_{A\sim\pi_t(s)}[\bar{Q}_t^\pi(s,A)]$.*

Solving the optimization problems in Equation (3) and Equation (4) for each $s,a$ pair isn't feasible for large state and action spaces. In this section, we state a closed-form expression for the minimum, which can be extracted from Rockafellar et al. (2000), Dorn et al. (2021). (Analogous results hold for the maximum). Define $\tau := \Lambda/(1+\Lambda)$. For any function $Q : \mathcal{S}\times\mathcal{A}\to\mathbb{R}$, we define the Bellman target and observational $(1-\tau)$-level conditional quantile of $Y_t(Q)$:

$$
\begin{aligned}
Y_t(Q) &:= R_t + \max_{a'}\,[Q(S_{t+1},a')] \qquad (5)\\
Z_t^{1-\tau}(Y_t(Q)\mid s,a) &:= \inf_z\{z\colon P_{\text{obs}}(Y_t(Q)\ge z \mid S_t=s, A_t=a)\le 1-\tau\}.
\end{aligned}
$$

Proposition 5 (**Conditional expected shortfall closed form solution**). *Equation (4) admits the closed-form solution:*

$$
(\bar{\mathcal{T}}_t^* Q)(s,a) = \mathbb{E}_{obs}\left[\alpha_t Y_t(Q) + \frac{1-\alpha_t}{1-\tau} Y_t(Q)\mathbb{I}\left[Y_t(Q)\le Z_{t,a}^{1-\tau}\right] \mid S_t=s, A_t=a\right]. \qquad (6)
$$

where we omitted some functional dependence when clear from context: $Z_{t,a}^{1-\tau}$ for $Z_t^{1-\tau}(Y_t(Q)\mid s,a)$, $\alpha_t$ for $\alpha_t(S,A)$, $\beta_t$ for $\beta_t(S,A)$. The solution function to Equation (4) is a superquantile (also called conditional expected shortfall, or covariate-conditional CVaR), which is the conditional expectation of exceedances of a random variable beyond its conditional quantile.

### 3.3. Estimation

Having introduced the key estimands of interest, we now introduce our estimation strategy, a robust analog of Fitted-Q Iteration (FQI).

The observational dataset $\mathcal{D}_{obs}$ comprises of $n$ trajectories of length $T$, was collected from the underlying MDP under an unknown behavior policy $\pi^b$ that depended on the unobserved state. We will write $\mathbb{E}_{n,t}$ to denote a sample average of the $n$ data points collected at time $t$, e.g. $\mathbb{E}_{n,t}[f(S_t,A_t,S_{t+1})] := \frac{1}{n}\sum_{i=1}^n f(S_t^{(i)},A_t^{(i)},S_{t+1}^{(i)})$. Nominal (non-robust) FQI (Ernst et al. 2006, Le et al. 2019, Duan et al. 2021) successively forms approximations $\hat{Q}_t$ at each time step by minimizing the Bellman error. Since $Q_t(s,a) = \mathbb{E}[Y_t(Q_{t+1})|S_t=s,A_t=a]$, FQE solves a sequential loss minimization problem: $\hat{Q}_t \in \arg\min_{q_t\in Q}\mathbb{E}_{n,t}[(Y_t(\hat{Q}_{t+1}) - q_t(S_t,A_t))^2]$. FQE/I is an example of *pseudo-outcome* regression. Pseudo-outcome regression has recently been used in causal inference (Kennedy 2020, Semenova and Chernozhukov 2021). We present the fitted-Q-iteration algorithm

**Algorithm 1** Confounding-Robust Fitted-Q-Iteration

Estimate the marginal behavior policy $\pi_t^b(a|s)$. Compute $\{\alpha_t(S_t^{(i)}, A_t^{(i)})\}_{i=1}^n$ as in Equation (2). Initialize $\hat{\overline{Q}}_T = 0$.
**for** $t = T-1, \ldots, 1$ **do**
  Compute the nominal outcomes $\{Y_t^{(i)}(\hat{\overline{Q}}_{t+1})\}_{i=1}^n$ as in eq. (5).
  For $a \in \mathcal{A}$, where $A_t^{(i)} = a$, fit $\hat{Z}_t^{1-\tau}$ the $(1-\tau)$th conditional quantile of the outcomes $Y_t^{(i)}$.
  Compute pseudooutcomes $\{\tilde{Y}_t^{(i)}(\hat{Z}_t^{1-\tau}, \hat{\overline{Q}}_{t+1})\}_{i=1}^n$ as in eq. (7).
  For $a \in \mathcal{A}$, where $A_t^{(i)} = a$, fit $\hat{\overline{Q}}_t$ via least-squares regression of $\tilde{Y}_t^{(i)}$ against $(S_t^{(i)}, A_t^{(i)})$.
  Compute $\pi_t^*(s) \in \arg\max_a \hat{\overline{Q}}_t(s,a)$.
**end for**

in the main text for brevity: evaluation (Le et al. 2019) is analogous, replacing the maximum over next-timestep actions with evaluation under the evaluation policy.

In our robust version of FQI, we approximate the robust Bellman operator from eq. (4), using the closed-form in Proposition 5. Unlike in the usual FQI algorithm, we now have an additional nuisance function: the conditional quantile. This suggests a simple two-stage procedure. First, estimate $Z_t^{1-\tau}$, and then estimate the conditional expectation in eq. (6) via regression using the estimated $Z_t^{1-\tau}$.

### 3.4. Improving estimation: the orthogonal pseudo-outcome

The two-stage procedure depends on the conditional quantile function $Z_t^{1-\tau}$, a *nuisance function* that must be estimated but is not our substantive target of interest. To avoid transferring biased first-stage estimation error of $Z_t^{1-\tau}$ to the Q-function, we introduce orthogonalization. Orthogonalized estimators remove the first-order dependence of estimating the target on the error in nuisance functions (Kennedy 2022, Newey 1994, Chernozhukov et al. 2018, Laan and Robins 2003). (See Section 6 for more discussion). We focus on the following orthogonal moment condition:

$$\tilde{Y}_t(Z,Q) := \alpha_t Y_t(Q) + \tfrac{1-\alpha_t}{1-\tau}\Big(Y_t(Q)\mathbb{I}\left[Y_t(Q) \le Z\right] - Z \cdot \{\mathbb{I}\left[Y_t(Q) \le Z\right] - (1-\tau)\}\Big) \tag{7}$$

Note that $\mathbb{E}[\{\mathbb{I}\left[Y_t(Q) \le Z\right] - (1-\tau) \mid S_t, A_t] = 0$. It is Neyman-orthogonal with respect to error in $Z^{1-\tau}$. When the quantile functions are consistent, the orthogonalized pseudo-outcome enjoys quadratic, not linear dependence on the first-stage estimation error in the quantile functions. We describe in more detail in the next section on guarantees. Note that this orthogonal moment is equivalent to the CVaR minimization formula (Rockafellar et al. 2000). Though the CVaR minimization formula is well-known, its favorable orthogonal properties relative to other representations of CVaR did not appear in that early work.[6]

We can apply the same pseudo-outcome regression procedure, as appears in (Kennedy 2020, Semenova and Chernozhukov 2021) to this orthogonal target. Our final estimator sequentially solves this following squared-loss minimization problem:

$$\hat{\overline{Q}}_t \in \arg\min_{q_t} \mathbb{E}_{n,t}[(\tilde{Y}_t(\hat{Z}_t^{1-\tau}, \hat{\overline{Q}}_{t+1}) - q_t(S_t, A_t))^2]. \tag{8}$$

We summarize the algorithm in Algorithm 1. In the appendix, we discuss a sample-splitting version in more detail; we describe the approach, which is standard, in the main text for brevity.

*Partial orthogonality: Comparison to other related work* The primary goal of our orthogonal estimation is to introduce *some* orthogonality *without estimating behavior policy propensities at all*. This allows the method to remain close to methods that practitioners use, such as fitted-Q-evaluation which estimates outcome models alone (the $Q$ functions), at the cost of some statistical efficiency. Therefore we orthogonalize the robust $Q$ with respect to the conditional quantile only. In contrast, Jeong and Namkoong (2020) consider orthogonality of a related (marginal) CVaR-type functional, however they view the *quantile* function as fixed, and perturb with respect to the behavior policy $\pi_b$ alone. In some sense, we seek the opposite: perturb the quantile function but not the behavior policy, so that these perspectives are complementary. Refining efficient estimators to improve potential instability from behavior policy estimation is a promising direction for future work.

*Sample splitting.* Lastly, to ensure independent errors in nuisance estimation and the fitted-Q regression, for the theoretical results, we study a cross-time variant of the standard cross-fitting/sample-splitting scheme for orthogonalized estimation and machine learning. Interleaving between timesteps ensures downstream policy evaluation errors are independent of errors in nuisance evaluation at time $t$. Finally, we note that sample splitting can be avoided by posing Donsker-type assumptions on the function classes in the standard way. In the experiments (and algorithm description) in the interest of data-efficiency we do not data-split. Recent work of Chen et al. (2022) shows rigorously that sample-splitting may not be necessary under stability conditions; extending that analysis to this setting would be interesting future work.

[6] We thank a reviewer for noting this important connection.

*Estimating and parametrizing conditional quantiles and conditional expected shortfall.* A large literature discusses methods for quantile regression (Koenker and Hallock 2001, Meinshausen 2006, Belloni and Chernozhukov 2011), as well as conditional expected shortfall (Cai and Wang 2008, Kato 2012) and can guide the choice of function class for quantiles and $\overline{Q}$ appropriately. We can learn the conditional quantile functions by minimizing the *pinball loss* over a function class $\mathcal{Z}$: $Z_t^{1-\tau}(Y_t(Q) \mid S_t, A_t) \in \arg\min_{z\in\mathcal{Z}} \mathbb{E}[L_{1-\tau}(Y_t(Q), z(S_t, A_t))]$, where the pinball loss $L_\tau(y, \hat{y})$ is $(1-\tau)(\hat{y} - y)$if $y < \hat{y}$, else $\tau(y - \hat{y})$ if $y \geq \hat{y}$.

REMARK 1 (EXTENSION TO CONTINUOUS ACTIONS). In the main text, we discuss discrete actions although the method directly extends to continuous action spaces at the cost of sharpness. See Section 4 for more details.

## 4. Connecting the model to data, to instantiate the method.

In this section, we focus on connecting potential models of unobserved confounders to the observed data and Assumption 1. We go into more detail as to graphical characterizations, our suggested statistical testing procedure, and calibrating the ambiguity sets from data. We conclude with a "meta-algorithm" for practitioners to use these tools to calibrate the strength of assumptions about their data.

### 4.1. Connecting Assumption 1 to data and models

*Both Assumptions 1 and 2 are necessary to conclude a Marginal MDP from Observed Markov Marginals* The next proposition highlights how *both* Assumptions 1 and 2 are necessary to conclude the relevant Marginal MDP; i.e. our characterization is tight.

PROPOSITION 6. *Given Assumption 1, the following results hold:*

1. *Without Assumption 2, there may exist a policy $\pi \neq \pi^b$ that may or may not depend on $U_t$ such that the observed-state Markov property does not hold for $\pi$, and hence the online decision problem may be a POMDP.*
2. *Under Assumption 2, there may exist a policy $\pi \neq \pi^b$ that does depend on $U_t$ such that the observed-state Markov property does not hold for $\pi$.*
3. *Under Assumption 2, for all policies $\pi^e$ that do not depend on $U_t$, the observed-state Markov property holds for $\pi^e$.*

Although Assumptions 1 and 2 are not strong enough to ensure that all alternative policies depending on $U_t$ yield off-policy Markovian transitions, together they ensure that for all policies $\pi^e$ restricted to observed states, transitions are Markovian.

*Interpreting Assumption 2, Faithfulness* To illustrate the role of faithfulness, suppose that the observational distribution satisfies Assumption 1 and therefore the conditional independence that $S_{t+1} \perp\!\!\!\perp S_{t-1} \mid S_t, A_t$. Consider two underlying causal paths permitted or not by a faithfulness assumption:

$$S_{t-1} \to U_{t-1} \to S_t \to U_t \to S_{t+1} \qquad \text{allowed under faithfulness if } S_{t+1} \perp\!\!\!\perp S_{t-1} \mid S_t, A_t \tag{9}$$

$$S_{t-1} \to U_t \to S_{t+1} \qquad \text{not allowed under faithfulness} \tag{10}$$

In both paths (eqs. (9) and (10)), a change in $S_{t-1}$ causes a change in $U_t$, which then causes a change in $S_{t+1}$. In the allowed path eq. (9), consistent with memoryless UCs Definition 1, the observed state $S_t$ *fully mediates* the information flowing from $S_{t-1}$ to $S_{t+1}$. Therefore, after conditioning on $S_t$, $S_{t-1}$ becomes irrelevant for predicting $S_{t+1}$. However, in Equation (10), the causal influence of $S_{t-1}$ to $S_{t+1}$ through unobserved $U_t$ *bypasses* the observed state. Then, even when conditioning on $S_t$, there remains an open pathway for causal influence from $S_{t-1}$ to $S_{t+1}$. Faithfulness requires that causal dependence appears in the observational distribution as a probabilistic *conditional dependence* that contradicts our supposition (Assumption 1). In this way, faithfulness and Assumption 1 would rule out dependence of unobserved confounders across timesteps such as in eq. (10).

*Testing for the Observed-State Markov Property* Crucially, Assumption 1 is an assumption on the observational joint distribution and is therefore testable from data. Testing for the validity of Markovian state representations (vs. a higher-order state representation or general POMDP) is difficult, since conditional independence testing itself is a difficult statistical problem (Shah and Peters 2020). We introduce a practical approach based on sample-splitting that tests whether adding one additional timestep of history improves the (aggregated) mean-squared error of the next-state transition regressions[7]. We incrementally add lags to the state until we fail to reject the null hypothesis of conditional independence $S_{t+1} \perp\!\!\!\perp S_{t-k} \mid S_{t-k+1}, \ldots, S_t, A_t$. Such an approach is informative since if the observed marginals are higher-order Markovian (Markovian after adding additional timesteps of history), then the only UCs that remain must be higher-order memoryless UCs. We describe the procedure in full in Section 3.6.

If the observed-state Markov property plus faithfulness is essentially equivalent to memoryless confounding, then we might see our test reject the null hypothesis (of Markovianity) if we don't add any lags. This is exactly what we find in our case study on MIMIC-III in Section 6.3 — we only fail to reject the Markov property after three lags of the state variable are included. We give an example of a DAG satisfying the observed-state Markov property with lags in Section B.3.

[7] Recent work of (Shi et al. 2020) estimate the conditional characteristic function and develop a test using double machine learning. However, the implementation is not set up to handle generic multivariate settings and we faced computational scaling issues.

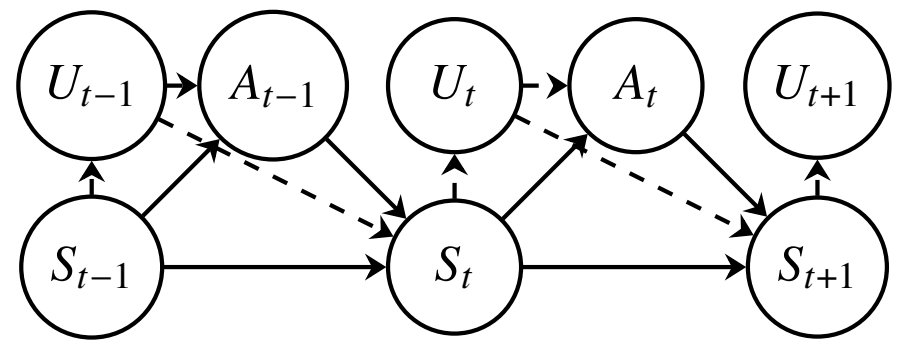

(a) DAG on latents and observables - emission model on unobserved confounders.

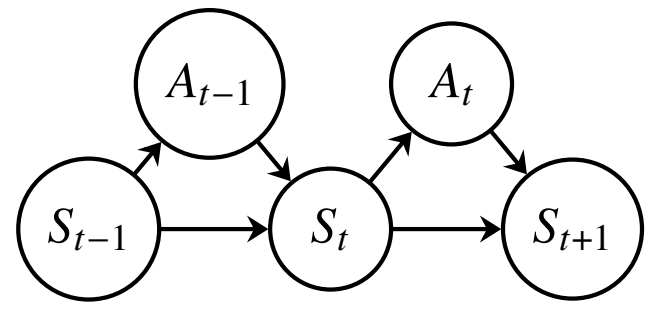

(b) Maximal Ancestral Graph (MAG) of Figure 6a. The marginals are Markovian.

**Figure 1 Underlying DAGs on time-homogenous and their latent projections to a maximal ancestral graph**

*Connecting other causal models to Assumption 1* Given an alternative causal model on unobserved confounders and states, how does one verify whether Assumption 1 will hold — i.e. whether its observational distribution, marginalizing out latents, is Markovian? *Maximal ancestral graphs* from the causal graph literature provide exactly this characterization (Richardson and Spirtes 2002). In fact, they describe something stronger than what we have sought previously, since they describe when the marginal observational distribution is Markovian for *all* policies, including those depending on $U_t$. A MAG represents a DAG after all latent variables have been marginalized out, and it preserves all entailed conditional independence relations among the measured variables which are true in the underlying DAG. MAGs are maximal in the sense that no additional edge may be added to the graph without changing the independence model. Therefore, when the MAG of any causal model on MDPs with unobserved confounders indicates marginal transitions are Markovian (under all policies, potentially depending on $U_t$), this certifies Assumption 1. Figure 6 highlights how adding an additional restriction to Definition 1, orienting the direction of causality that $S_t \to U_t$, implies Assumption 1 for all $(S_t, U_t)$ policies. This analysis also highlights how our path analysis of Proposition 6 is necessary, so that we can weaken our required conditions for $(S_t)$-dependent policies to just Assumptions 1 and 2.

### 4.2. Calibrating the causal ambiguity: How to choose $\Lambda$?

The first approach is to calibrate $\Lambda$ using corresponding values for observed variables (Hsu and Small 2013). In a setting where $s \in \mathbb{R}^d$, calibrating the marginal sensitivity model works as follow: for $k \in \{1, \ldots, d\}$, let $s_t^{-k}$ denote the state leaving out the $i$th component. We predict the action propensities using this leave-one-out state: $\pi_t^b(A_t \mid S_t^{-k}) := P_{\text{obs}}(A_t | S_t^{-k})$, and then compute the resulting odds ratio. One can then set $\Lambda$ based on these distributions of odds-ratios generated by observed confounders, e.g. $\hat{\Lambda} = \max_i \left\| \left( \frac{\pi_t^b(A_t|S_t)}{1-\pi_t^b(A_t|S_t)} \right) / \left( \frac{\pi_t^b(A_t|S_t^{-k})}{1-\pi_t^b(A_t|S_t^{-k})} \right) \right\|_\infty$ or based on another quantile of the odds-ratio distribution. For example, in a hypothetical clinical setting, let's say the patient's smoking status is the strongest observed driver of treatment and has an effective $\Lambda = 3$. A practitioner might say "I do not believe there exists an unobserved variable with more explanatory power than smoking" to justify a choice of $\Lambda = 3$.[8]

### 4.3. Meta-algorithm: justifying the method

In this section, we have introduced different tools for justifying key assumptions and choices ($\Lambda$) for the method. How would a practitioner use these tools to justify use of our method from data? We suggest the following high-level flow:

- Run the conditional independence test (Algorithm 2) to determine whether Assumption 1 (Observed Markov) holds. If not, try to satisfy Assumption 1 on a history-augmented state space: use Algorithm 3 to use the conditional independence test (Algorithm 2) to determine how many lags satisfy Assumption 1 and therefore the higher-order memoryless UC assumption Definition 2.[9] Balance bias-variance by checking overlap on the augmented state.
- Interpret potential choices of $\Lambda$ using the tools of Section 4.2. (Justifying the final choice requires domain knowledge of the potential informativity of relevant UCs).
- Run the confounding-robust fitted-Q-iteration (Algorithm 1).

## 5. Analysis and guarantees

We describe the orthogonalized estimation results, before the results about the full output of the robust fitted-Q-iteration. For this section, when another type of norm is not indicated, we let $\|f\| := \mathbb{E}[f^2]^{1/2}$ indicate the 2-norm. First, we require some regularity conditions for estimation. We assume nonnegative bounded rewards throughout.

Assumption 4 **(Boundedness)**. *Outcomes are nonnegative and bounded:* $0 \leq R_t \leq B_R, \forall t$. *The state space is bounded.*

We assume the transitions are continuously distributed, a common regularity condition for the analysis of quantiles.

Assumption 5 **(Bounded conditional density)**. *Assume that* $P_t(s_{t+1} \mid s_t, a) < M_P, \forall t, s_t, s_{t+1}$ *a.s.*

[8] Another approach is to plot the worst-case value of a candidate policy $\pi^e$ over a range of values for the sensitivity parameter $\Lambda$. If the value of the candidate policy is consistently better than baseline until e.g. $\Lambda = 10$, this would indicate the new policy is very robust — it will out-perform the baseline unless there exists an unobserved confounder with an enormous 10x impact on the odds-ratio for $A_t$. We illustrate this later in our sepsis case study.

[9] If it it requires all $K-1$ lags, the underlying data is non-Markovian and likely a POMDP, so use other proximal inference methods (Bennett and Kallus 2024).

We let $\hat{\mathbb{E}}_n$ indicate a function obtained by regression, on an appropriate data split independent of the nuisance estimation. Define

$$
\begin{aligned}
\hat{\overline{Q}}_t(s,a) &= \hat{\mathbb{E}}_n[\tilde{Y}_t(\hat{Z}_t, \hat{\overline{Q}}_{t+1}) \mid s,a] && \text{feasible regressed robust Q,} \\
\tilde{\overline{Q}}_t(s,a) &= \hat{\mathbb{E}}_n[\tilde{Y}_t(Z_t, \hat{\overline{Q}}_{t+1}) \mid s,a] && \text{oracle-nuisance regressed robust Q} \\
\overline{Q}_t(s,a) &= \mathbb{E}[\tilde{Y}_t(Z_t, \hat{\overline{Q}}_{t+1}) \mid s,a] && \text{oracle robust Q.}
\end{aligned}
$$

In the above, $\hat{\overline{Q}}_t(s,a) = \hat{\mathbb{E}}_n[\tilde{Y}_t(\hat{Z}_t, \hat{\overline{Q}}_{t+1}) \mid s,a]$ is the feasible *regressed* robust-Q-estimator with estimated nuisance $\hat{Z}$, while $\tilde{\overline{Q}}(s,a) = \hat{\mathbb{E}}_n[\tilde{Y}_t(Z_t, \hat{\overline{Q}}_{t+1}) \mid s,a]$ is the *regressed* robust-Q-estimator with *oracle* nuisance $Z$, and $\overline{Q}_t(s,a)$ is the true robust Q output at time $t$ (relative to the future $Q$ functions that are the output of the algorithm).

We assume the following regression stability assumption, which appears in Kennedy (2020). It is a generalization of stochastic equicontinuity and is satisfied, for example, by nonparametric linear smoothers.

Assumption 6 **(Regression stability)**. *Suppose $\mathcal{D}_1$ and $\mathcal{D}_2$ are independent training and test samples, respectively. Let: 1. $\widehat{f}(x) = \widehat{f}(x; \mathcal{D}_1)$ be an estimate of a function $f(x)$ using the training data $\mathcal{D}_1$, 2. $\widehat{b}(x) = \widehat{b}(x; \mathcal{D}_1) \equiv \mathbb{E}[\widehat{f}(x) - f(x) \mid \mathcal{D}_1, X = x]$ the conditional bias of the estimator $\widehat{f}$, 3. $\widehat{\mathbb{E}}_n[Y \mid X = x]$ denote a generic regression estimator that regresses outcomes on covariates in the test sample $\mathcal{D}_2$. Then the regression estimator $\widehat{\mathbb{E}}_n$ is defined as stable at $X = x$ (with respect to a distance metric $d$ ) if*

$$
\frac{\widehat{\mathbb{E}}_n[\widehat{f}(x)|X=x] - \widehat{\mathbb{E}}_n[f(x)|X=x] - \widehat{\mathbb{E}}_n[\widehat{b}(x)|X=x]}{\sqrt{\mathbb{E}\left(\left[\widehat{\mathbb{E}}_n[f(x)|X=x] - \mathbb{E}[f(x)|X=x]\right]^2\right)}} \xrightarrow{p} 0
$$

*whenever $d(\widehat{f}, f) \xrightarrow{p} 0$.*

Under these regularity conditions, we can show that the bias due to the first-stage estimation of the conditional quantiles is only quadratic in the estimation error of $\hat{Z}_t$.

Proposition 7 **(CVaR estimation error)**. *Assume Assumptions 4 to 6. For $a \in \mathcal{A}, t \in [T-1]$, if the conditional quantile estimation is $o_p(n^{-\frac{1}{4}})$ consistent, i.e. $\|\hat{Z}_t^{1-\tau} - Z_t^{1-\tau}\|_\infty = o_p(n^{-\frac{1}{4}})$, $\mathbb{E}[\|\hat{Z}_t^{1-\tau} - Z_t^{1-\tau}\|_2] = o_p(n^{-\frac{1}{4}})$, then*

$$
\|\hat{\overline{Q}}_t(S,a) - \overline{Q}_t(S,a)\|_2 \leq \|\tilde{\overline{Q}}_t(S,a) - \overline{Q}_t(S,a)\|_2 + o_p(n^{-\frac{1}{2}}).
$$

This implies we can maintain $o_p(n^{-\frac{1}{2}})$ consistent estimation of robust $\overline{Q}$ functions under weaker $o_p(n^{-\frac{1}{4}})$ consistency of the conditional quantile functions $Z$.

Next, we describe key assumptions for convergence of fitted-Q-iteration, concentrability which restricts the distribution shift in the sequential offline data vs. optimized policies, and approximate Bellman completeness which assumes the closedness of the regression function class under the Bellman operator. Both these assumptions are standard requirements for fitted-Q-iteration, but certainly not innocuous; they do impose restrictions.

ASSUMPTION 7 **(Concentrability)**. *Given a policy $\pi$, let $\rho_t^\pi$ denote the marginal distribution at time step $t$, starting from $s_0$ and following $\pi$, and $\mu_t$ denote the true marginal occupancy distribution under $\pi^b$. There exists a parameter $C$ such that*

$$\sup_{(s,a,t)\in\mathcal{S}\times\mathcal{A}\times[T-1]} \frac{\mathrm{d}\rho_t^\pi}{\mathrm{d}\mu_t}(s,a) \leq C \quad \textit{for any policy } \pi.$$

ASSUMPTION 8 **(Approximate Bellman completeness)**. *There exists $\epsilon > 0$ such that, for all $t \in [T-1]$, where $\epsilon$ is at most on the order of $O_p(n^{-\frac{1}{2}})$,*

$$\sup_{q_{t+1}\in\mathcal{Q}_{t+1}} \inf_{q_t\in\mathcal{Q}_t} \|q_t - \overline{\mathcal{T}}_t^\star q_{t+1}\|_{\mu_t}^2 \leq \epsilon.$$

Concentrability is analogous to sequential overlap or positivity, as it is called in single-timestep causal inference. It assumes a uniformly bounded density ratio between the true marginal occupancy distribution and those induced by arbitrary policies. Approximate Bellman completeness assumes that the function class $\mathcal{Q}$ is approximately closed under the robust Bellman operator. Assuming that $\epsilon$ is at most $O_p(n^{-\frac{1}{2}})$ is somewhat restrictive, but is consistent with frameworks for local model misspecification that consider local asymptotics with $O_p(n^{-\frac{1}{2}})$ vanishing bias.

Although we ultimately seek an optimal policy, approaches based on fitted-Q-evaluation and iteration instead optimize the squared loss, which is related to the Bellman error that is a surrogate for value suboptimality.

DEFINITION 4 (BELLMAN ERROR). Under data distribution $\mu_t$, define the Bellman error of function $q = (q_0, \ldots, q_{T-1})$ as: $\mathcal{E}(q) = \frac{1}{T}\sum_{t=0}^{T-1} \|q_t - \overline{\mathcal{T}}_t^* q_{t+1}\|_{\mu_t}^2$

The next lemma, which appears as Duan et al. (2021, Lemma 3.2) (finite horizon), Xie and Jiang (2020, Thm. 2) (infinite horizon), justifies this approach by relating the Bellman error to the value suboptimality.

LEMMA 1 **(Bellman error to value suboptimality)**. *Under Assumption 7, for any $q \in \mathcal{Q}$, we have that, for $\pi$ the policy that is greedy with respect to $q$, $V_1^*(s_1) - V_1^\pi(s_1) \leq 2T\sqrt{C \cdot \mathcal{E}(q^\pi)}$.*

We will describe convergence results based on generic results for loss minimization over a function class of restricted complexity. We use standard covering and bracketing numbers to quantify the functional complexity of infinite function classes.

Definition 5 (Covering numbers, e.g. (van de Vaart and Wellner 1996)). Let $(\mathcal{F}, \|\cdot\|)$ be an arbitrary semimetric space. Then the covering number $N(\epsilon, \mathcal{F}, \|\cdot\|)$ is the minimal number of balls of radius $\epsilon$ needed to cover $\mathcal{F}$.

Definition 6 (Bracketing numbers). Given two functions $l$ and $u$, the bracket $[l, u]$ is the set of all functions $f$ with $l \leq f \leq u$. An $\epsilon$-bracket is a bracket $[l, u]$ with $\|u - l\| < \epsilon$. The bracketing number $N_{[]}(\epsilon, \mathcal{F}, \|\cdot\|)$ is the minimum number of $\epsilon$-brackets needed to cover $\mathcal{F}$.

The covering and bracketing numbers for common function classes such as linear, polynomials, neural networks, etc. are well-established in standard references, e.g. Wainwright (2019), van de Vaart and Wellner (1996). We assume either that the function class for $Q, \mathcal{Z}$ is finite (but possibly exponentially large), or has well-behaved *covering* and *bracketing* numbers.

Assumption 9 **(Finite function classes.)**. *The Q-function class $Q$ and conditional quantile class $\mathcal{Z}$ are finite but can be exponentially large.*

Assumption 10 **(Infinite function classes with well-behaved covering number.)**. *The Q-function class $Q$, and conditional quantile class $\mathcal{Z}$ have covering numbers $N(\epsilon, Q, d)$, $N(\epsilon, \mathcal{Z}, d)$ (respectively).*

Theorem 2 **(Fitted Q Iteration guarantee)**. *Suppose Assumptions 4 to 8 and let $B_R$ be the bound on rewards. Recall that $\mathcal{E}(\hat{Q}) = \frac{1}{T}\sum_{t=0}^{T-1}\left\|\hat{Q}_t - \overline{\mathcal{T}}_t^\star \hat{Q}_{t+1}\right\|_{\mu_t}^2$. Then, with probability greater than $1-\delta$, under Assumption 9 (finite function class), we have that*

$$\mathcal{E}(\hat{Q}) \leq \epsilon_{Q,\mathcal{Z}} + \frac{56(T^2+1)B_R\log\{T|Q||\mathcal{Z}|/\delta\}}{3n} + \sqrt{\frac{32(T^2+1)B_R\log\{T|Q||\mathcal{Z}|/\delta\}}{n}\epsilon_{Q,\mathcal{Z}}} + o_p(n^{-1}),$$

*while under Assumption 10 (infinite function class), choosing the covering number approximation error $\epsilon = O(n^{-1})$ such that $\epsilon_{Q,\mathcal{Z}} = O(n^{-1})$, we have that*

$$\mathcal{E}(\hat{Q}) \leq \epsilon_{Q,\mathcal{Z}} + \frac{1}{T}\sum_{t=0}^{T-1}\left\{\frac{56(T-t-1)^2\log\{TN_{[]}(2\epsilon L_t, \mathcal{L}_{q_t(z'),z}, \|\cdot\|)/\delta\}}{3n}\right\} + o_p(n^{-1}).$$

*where $L_t = KB_r(T-t-1)\Lambda$ for an absolute constant $K$.*

Finally, putting the above together with Lemma 1, our sample complexity bound states that the policy suboptimality is on the order of $O(n^{-\frac{1}{2}})$. Note that this analysis omits estimation error in $\pi^b$ for simplicity. Note that Lemma 5 of the appendix gives that $N_{[]}\left(2\epsilon L, \mathcal{L}_{q(z'),z}, \|\cdot\|\right) \leq N(\epsilon, Q \times \mathcal{Z}, \|\cdot\|) \leq N(\epsilon, Q, \|\cdot\|) N(\epsilon, \mathcal{Z}, \|\cdot\|)$. Therefore ensuring some $\epsilon = cn^{-\frac{1}{2}}$ approximation error (for some arbitrary constant $c$) can be achieved by fixing $\epsilon' = \frac{\epsilon}{2L}$; i.e. we require finer approximation.

*Proof sketch.* As appears elsewhere in the analysis of FQI (Duan et al. 2021), we may obtain the following standard decomposition:

$$\|\hat{\overline{Q}}_{t,\hat{Z}_t} - \overline{\mathcal{T}}^*_{t,\hat{Z}_t}\hat{\overline{Q}}_{t+1}\|^2_{\mu_t}$$
$$= \mathbb{E}_\mu[(\tilde{Y}_t(\hat{Z}_t^{1-\tau}, \hat{\overline{Q}}_{t+1}) - \hat{\overline{Q}}_{t,\hat{Z}_t}(S_t, A_t))^2] - \mathbb{E}_\mu[(\tilde{Y}_t(Z_t^{1-\tau}, \hat{\overline{Q}}_{t+1}) - \overline{Q}^\dagger_{t,Z_t}(S_t, A_t))^2] + \|\overline{Q}^\dagger_{t,Z_t} - \overline{\mathcal{T}}^*_t \hat{\overline{Q}}_{t+1}\|^2_{\mu_t}$$

where $\overline{Q}^\dagger_{t,Z_t}$ is the oracle squared loss minimizer, relative to the $\hat{\overline{Q}}_{t+1}$ output from the algorithm. Assumption 8 (completeness) bounds the last term. Our analysis differs onwards with additional decomposition relative to estimated nuisances and applying orthogonality from Proposition 7.

Finally, we note that our analysis extends immediately to the infinite-horizon case, discussed in Section 5.2 of the appendix due to space constraints. Crucially, the (s,a)-rectangular uncertainty set admits a stationary worst-case distribution (Iyengar 2005).

### 5.1. Bias-variance tradeoff in selection of $\Lambda$

We can quantify the dependence of the sample complexity on constants related to problem structure. We consider an equivalent regression target which better illustrates this dependence.

Corollary 1. *Assume that the same function classes $Q, \mathcal{Z}$ are used for every timestep, and they are VC-subgraph with dimensions $v_q, v_z$. Assume that $\epsilon_{Q,\mathcal{Z}} = 0$. Then there exist absolute constants $K, k$ such that*

$$\mathcal{E}(\hat{Q}) \leq K\{\log(v_q+v_z)+2(v_q+v_z)+2((v_q+v_z)-1)(T-1)\left(\log\left(2KB_r\Lambda(T-1)n/\epsilon\right)-1\right)\}n^{-1}+o_p(n^{-1}).$$

Note that the width of confidence bounds on the robust $Q$ function scale logarithmically in $\Lambda$, which illustrates *robustness-variance-sharpness* tradeoffs. Namely, as we increase $\Lambda$, we estimate more extremal tail regions, which is more difficult. Sharper tail bounds on conditional expected shortfall estimation would also qualitatively yield similar insights.

### 5.2. Confounding with Infinite Data

While Theorem 2 analyses the difficulty of estimating the robust value function, here we analyze how the true robust value function differs from the nominal value function at the population-level for policy evaluation (not optimization). This gives a sense of how potentially conservative the method is, in case unconfoundedness held after all. We consider a simplified linear Gaussian setting.

Proposition 8. *Let $\mathcal{S} = \mathbb{R}$ and $\mathcal{A} = \{0,1\}$. Define parameters $\theta_P, \theta_R, \sigma_P \in \mathbb{R}$. Suppose in the observational distribution that $S_{t+1}|S_t, A_t \sim \mathcal{N}(\theta_P S_t, \sigma_P)$, $R(s,a,s') = \theta_R s'$, $\pi_t^e(1|S_t) = 0.5$, and consider some $\pi^b$ such that $\pi_t^b(A_t|S_t)$ does not vary with $S_t$. Finally, let $\beta_i := \theta_R \sum_{k=1}^{i} \theta_P^k$ and notice that the nominal, non-robust value functions are $V_{T-i}^{\pi^e}(s) = \beta_i s$ for $i \geq 1$. Then:*

$$|V_0^{\pi^e}(s) - \bar{V}_0^{\pi^e}(s)| \leq (16\theta_P)^{-1}(\textstyle\sum_{i=0}^{T-1}\beta_i)\sigma_P \log(\Lambda).$$

Note that the cost of robustness gets worse as the horizon $T$ increases, depending on the value of $\theta_P$. The parameter $\theta_P$ is the autoregressive coefficient how strongly last period's state impacts this period's state. In the framework of linear systems, $\theta_P$ determines stability, where whether the system is stable, marginally stable, or unstable gives different scalings in $T$ for the cost of robustness. For stable systems, unobserved confounding can at worst induce bias that is linear in horizon, but for unstable systems, the bias could increase exponentially.[10] This can be generalized to higher dimensions; the bias then depends on the spectrum of the transition matrix.

On the other hand, the scaling with the degree of confounding $\Lambda$ is independent of horizon, and has a modest $\log(\Lambda)$ rate. This is surprising: it suggests that the horizon of the problem presents more of a challenge than the strength of confounding at each time step, and that $T$ and $\Lambda$ do not interact at the population level — at least in a simple linear-Gaussian setting. Characterizing exactly when the scaling with $\Lambda$ is horizon-independent is a promising direction for future work.

## 6. Experiments

We first illustrate the benefits of our orthogonalized fitted-Q-iteration in a simulated example, where we know the ground-truth outcomes. Next, we illustrate how the robust fitted-Q-iteration allows robust evaluation of policies learned with methods similar to those used in the literature, and learning robust policies, revisiting the example of sepsis data from MIMIC-III since it has been widely studied in the literature.

### 6.1. Simulation: Orthogonality helps estimation

In this section, we validate the (estimation) performance of our method, including its scaling with the sensitivity parameter $\Lambda$ and the importance of orthogonalization. We perform simulation experiments in a mis-specified sparse linear setting with heteroskedastic conditional variance. Previous methods, Namkoong et al. (2020), Kallus and Zhou (2020a), Bruns-Smith (2021), cannot solve this continuous state setting with confounding at every time step. We use the following (marginal) data-generating process for the observational data:

$$\mathcal{S} \subset \mathbb{R}^d, \mathcal{A} = \{0,1\}, S_0 \sim \mathcal{N}(0, 0.01), \qquad \pi^b(1|S_t) = 0.5, \ \forall S_t$$
$$P_{\text{obs}}(S_{t+1}|S_t, A_t) = \mathcal{N}(\theta_\mu S_t + \theta_A a, \max\{\theta_\sigma S_t + \sigma, 0\}), \qquad R(S_t, A_t, S_{t+1}) = \theta_R^T S_{t+1}$$

with parameters $\theta_\mu, \theta_\sigma \in \mathbb{R}^{d \times d}, \theta_R, \theta_A \in \mathbb{R}^d, \sigma \in \mathbb{R}$ chosen such that $AS_t + \sigma > 0$ with probability vanishingly close to 1. The number of features $d = 25$ and $\theta_\mu$ and $\theta_\sigma$ are chosen to be column-wise sparse, with 5 and 20 non-zero columns respectively. We collect a dataset of size $n = 5000$ from a single trajectory. In the appendix we include results from a higher-dimensional setting with $d = 100$ and $n = 600$, findings are qualitatively similar.

[10] The term $(\sum_{i=0}^{T-1}\beta_i)/\theta_P$ is asymptotically linear in $T$ for $|\theta_P| < 1$; quadratic in $T$ for $|\theta_P| = 1$, and asymptotically $\theta_P^T$ for $|\theta_P| > 1$. In contrast, for the unconfounded problem, unstable systems are typically easier to estimate due to their better signal-to-noise ratio (Simchowitz et al. 2018).

| | $\Lambda = 1$ | $\Lambda = 2$ | $\Lambda = 5.25$ | $\Lambda = 8.5$ | $\Lambda = 11.75$ | $\Lambda = 15$ |
|---|---|---|---|---|---|---|
| Method | FQI | Non-orth/Orth | | | | |
| MSE($\bar{V}_0^*$) | 0.29 | 0.69 / 0.41 | 11 / 0.56 | 51 / 0.71 | 171 / 1.3 | 433 / 2.7 |
| $\ell_2$ Param. Error | 2.5 | 3.5 / 2.7 | 7.3 / 3.1 | 17 / 3.4 | 34 / 3.7 | 56 / 3.9 |
| % Wrong Action | 0% | 5e-5% / 0% | 0.39% / 0% | 2.5% / 4e-5% | 5.4% / 6e-4% | 8.2% / 4e-3% |

**Table 1 Simulation results with $d = 25$ and $n = 5000$, reporting the value function MSE, Q function parameter error, and the portion of the time a sub-optimal action is taken. Each cell shows Non-Orthogonal / Orthogonal results for each $\Lambda$.**

We estimate $\bar{V}_1^*(s)$ for $T = 4$ and several different values of $\Lambda$, using both the orthogonalized and non-orthogonalized robust losses. For function approximation of the conditional mean and conditional quantile, we use Lasso regression.[11] For details see Section 8.1 in the Appendix.

We report the mean-squared error (MSE) of the value function estimate over 100 trials, alongside the average $\ell_2$-norm parameter error and the percentage of the time a wrong action is taken. The MSE and percentage of mistakes compare the estimated value function/policy to an analytic ground truth and are evaluated on an independently drawn and identically distributed holdout sample of size $n = 200,000$ drawn from the initial state distribution. See the Appendix for details on the ground truth derivation.

[11] Note that while this is correctly specified in the non-robust setting, the CVaR is non-linear in the observed state due to the non-linear conditional standard deviation of $\theta_R^T S_{t+1}$, and therefore the Lasso is a misspecified model for the quantile and robust value functions.

**Table 2 Cumulative reward $E[V^{\pi^*}(S_0)]$ for healthcare simulation with autoregressive unobserved confounders, with $U_t \to U_{t+1}$. Reward accumulated over 20 timesteps, $n = 2000$ evaluation trajectories, averaged over 75 Monte Carlo trials**

| Lag | Confounded FQI | Orthogonal Robust FQI $\Lambda = 0.5$ | Orthogonal Robust FQI $\Lambda = 0.75$ | Orthogonal Robust FQI $\Lambda = 1.0$ | State-Only Optimal (skyline) |
|---|---|---|---|---|---|
| Lag = 0 | 188.4 ± 2.9 | 207.4 ± 2.0 | 207.5 ± 2.0 | 204.8 ± 1.9 | 211.5 ± 25.8 |
| Lag = 1 | 179.9 ± 3.5 | 221.5 ± 1.7 | 220.1 ± 2.2 | 211.0 ± 2.3 | 219.8 ± 20.2 |

The low-dimensional results in Table 1 illustrate two important phenomena. First, the MSE increases with $\Lambda$. Estimated lower bounds become less reliable as $\Lambda$ increases. Second, the non-orthogonal algorithm suffers from substantially worse mean-squared error and as a result selects a sub-optimal action more often, especially at high levels of $\Lambda$. Orthogonalization has a very large impact not just in theory, but in practice.

### 6.2. Policy learning simulation: healthcare-inspired

We next develop a simulation testbed to highlight the benefits of our method for policy learning where ground-truth evaluation is possible. We develop a contextual extension of a healthcare-inspired simple Markov decision process on 6 tabular states, $L \in \{0, 1, \ldots, 5\}$ that first appeared in (Goyal and Grand-Clement 2022), with the last state being an absorbing state, representing mortality. These states might indicate severity of health condition, for example. There are three actions, high drug, low drug, or do nothing. At any timestep, patients can improve ($L_{t+1} = L_t + 1$), stay the same, or worsen ($L_{t+1} = L_t - 1$), with some tabular transition probabilities. The key tradeoff is that taking drug actions can worsen short-term rewards, while stabilizing patient health over the long-term. We add several components to extend this to a contextual, rich function-approximation setting with unobserved confounders, detailed in Appendix 8.2. We consider a hybrid setup where we assume final state observations $S_t \in \mathbb{R}^4$ are contextual views of the *latent* discrete state $L_t$; each value of $L_t = l$ is associated with a "center" vector $c_l \in \mathbb{R}^4$ and the continuous contextual state $S_t$ evolves as a mixture of an autoregressive process and the latent state, $S_{t+1} = \alpha S_t + (1 - \alpha)c_{L_t}, \alpha \in (0, 1)$; similarly, differences in observed state from the latent state, $S_t - c_{L_t}$, affect the transition probabilities, introducing contextual dynamics. We introduce unobserved confounders that are *state-dependent* and also have some autoregressive component, $U_t \longrightarrow U_{t+1}$. This illustrates robustness of our approach. The unobserved confounder $U_t \in \{-1, 1\}$ is such that high drug is more immediately costly and has a high risk of mortality, but when $U_t = 1$ it improves health state, so a policy with access to $U_t$ generally chooses high drug when $U_t = 1$. The confounded dataset is generated by mixing a policy trained to be optimal with unobserved confounders with additional bias towards taking the high drug. This introduces a spurious correlation between high drug actions and health improvement, which naive FQI picks up on. In Table 2 we compare naive FQI with XGBoost as the regressor with versions of our orthogonal robust method (XGBoost for robust Q regression and gradient-boosted regression for the quantile regression) for $\Lambda = 0.25, 0.5, 1$. As an unattainable "skyline" for comparison, we also include the observed-state-optimal policy that is learned on a separate *unconfounded* dataset obtained via uniform random exploration. Finally, since we introduce some direct dependence on $U_{t-1} \to U_t$ so that $U_t$ is Markovian rather than 1-memoryless, we illustrate how our method can handle *higher-order memoryless UCs* by comparing performance with the observed state only (Lag = 0) vs. including a lagged state, (Lag = 1). We see that even without including the lag, orthogonal robust FQI improves upon naive confounded FQI (9.5% improvement), illustrating robustness of our approach to moderate violations of the memoryless assumption. However, including the lag indeed improves performance, leading to increased 23.1% improvement upon confounded FQI. Notably, for modest values of $\Lambda$, performance of orthogonal robust FQI from *confounded* data nears that of *unattainable* state-only optimal FQI learned on uniformly random exploration data. As $\Lambda$ increases further, the robust approach becomes more conservative and prefers to avoid mortality events, which increases usage of the costly high-drug action and attenuates improvements.

### 6.3. Complex real-world healthcare data

In the next computational experiments, we show how our method extends to more complex real-world healthcare data via a case study around the use of MIMIC-III data for off-policy evaluation of learned policies for the management of sepsis in the ICU with fluids and vasopressors (Larkin 2023a). Sepsis is an umbrella term for an extreme response to infection and is a leading cause of mortality, healthcare costs, and readmission. The management of sepsis is complex, dynamic i.e. tracking the patient's state over time, and has substantial uncertainty about clinical guidelines (Evans et al. 2021). For example, giving IV fluids is expected to be beneficial at the very beginning, but there are also expected risks from too much (Larkin 2023b).The pioneering efforts in releasing the MIMIC-III database enabled the development model-based or offline reinforcement learning methods (Liu et al. 2020, Raghu et al. 2017a, 2018, Lu et al. 2020, Rosenstrom et al. 2022). However, a crucial challenge is *off-policy evaluation* for credible, data-driven estimates of the benefits of these learned policies, that are less vulnerable to model assumptions.

Crucial assumptions such as *unconfoundedness* are likely violated in this setting: treatment decisions probably included additional information not recorded in the database. (Indeed, the clinical literature certainly discusses other aspects of patient state and potential actions not included in the data). On the other hand, the comprehensive electronic health record (EHR) contains the most important factors in clinical decision-making such as patient vitals. So, our methods that develop *robust bounds* can highlight the sensitivity of current learned policies to potential violations of sequential unconfoundedness. Since many research works used fitted-Q-iteration, we compare confounding-robust policies vs. naive policies for prescriptive insights.

We now describe the specific MDP data primitives. We follow the data preprocessing of Killian et al. (2020). The data covers 72 hours past the onset of sepsis. Observed actions, administration of fluids or vaso-pressors, were categorized by volume and segmented into quantiles per each action type based on observational frequency. This leads to 25 possible discrete actions. Demographic and contextual features include age, gender, weight, ventilation and re-admission status. Other time-varying features include patient information such as blood pressure, heart rate, INR, various blood cell counts, respiratory rate, and different measures of oxygen levels (see Killian et al. (2020, Table 2) for exact description). The reward function takes on three values: $R = \{-1, 0, +1\}$ where $-1$ indicates patient death, $+1$ indicates leaving the hospital; and 0 for all other events.

#### 6.3.1. Instantiating our framework

*Determining the number of lags and model selection* We run our conditional independence testing algorithm (Algorithm 2) to determine the number of lags needed to establish Markovian marginal transition probabilities. The number of lags introduces a bias-variance trade-off: including not enough lags can lead to some misspecification of the uncertainty set's robustness, while including too many lags greatly harms estimation due to exponentially decreasing "effective sample complexity" for 1) history-dependent behavior policy overlap and 2) learning history dependent nuisances. This is a fundamental issue in history-dependent estimation (Zhang and Jiang 2024). Our preferred specification is including **two lagged states**, where 11/17 (64.71%) conditional independence tests done at the time-step level fail to reject the null hypothesis of conditional independence. We find that this captures much of the dependencies, still allowing for one-stage-lagged-Markovian unobserved confounders, while retaining good estimation properties[12]. We leave a complete model selection approach for future work.

*Calibrating the robustness parameter* $\Lambda$ See Figure 15 in the appendix to see a calibration plot of the distributions of odds-ratios obtained by dropping each covariate. (Note that the preprocessing results in data representations, so the dimensions are not directly interpretable). The 90% quantile of the lower bound on $\Lambda$ is given by $\Lambda = 1.42$, and the 99% quantile is given by $\Lambda = 2.48$.

*Regression estimatiions* For this case study, we perform flexible non-parametric regression using gradient-boosted trees in place of the simple linear models in our earlier simulations (Friedman 2001, Hastie et al. 2009). Features include the full state vector and indicators for each action.

Implementing the robust estimator for MSM parameter $\Lambda$ requires only a few simple modifications of nominal FQI with off-the-shelf tools. First, we estimate the behavior policy $\pi^b$ using a gradient-boosted classifier. We estimate a conditional quantile model using gradient-boosted regression with the quantile loss, natively available in the `scikit-learn` package. We fit the robust Q orthogonal pseudo-outcome regression with gradient-boosted regression. We compute the value functions and optimal policies for a time horizon up to $T = 15$.

**6.3.2. MIMIC Results** This case study is not meant to be a medical analysis, but concretely illustrates why caution is needed for interpreting offline RL applied to healthcare settings.

Finally, in Figure 2a we summarize how the robust optimal actions change as the sensitivity parameter $\Lambda$ is increased. We coarsen the $5 \times 5$ action space into four groups: no/less treatment (low action indices for both fluids/vasopressors), only IV fluid (high fluid action index), only vasopressors, and both fluid and vasopressors (high action indices for both fluids/vasopressors). At the far left, we have $\Lambda = 1$, which corresponds to the nominal policy, where there is an even mix of treatments. As $\Lambda$ increases, the number of untreated or those receiving only fluid drops. We see substantial substitution to vasopressors and some increase in both treatment, resulting in overall an increase in vasopressor usage. The historical data reflects much more intensive usage of fluids, which the nominal policy also picks up on. Overall the robust policies move away from fluid-only actions towards vasopressors. This is in line with meta-analyses and studies in the clinical literature that suggest that conservative management (especially if concerned about mortality risk

[12] Our conditional independence testing approach tests the data from each timestep separately; we arbitrarily set a threshold for 90% of timesteps where we fail to reject the null hypothesis of conditional independence to fix a number of lags. Including one lagged state results in 3/18 (16.67%) of timesteps reporting independent, two lagged states result in 11/17 (64.71%), while three lagged states result in 13/16 (81.25%) and four lagged states result in full independence. In the interest of transparency, we report that with three lagged states, there are extreme overlap violations. Two lags balances bias-variance issues in estimating the bounds.

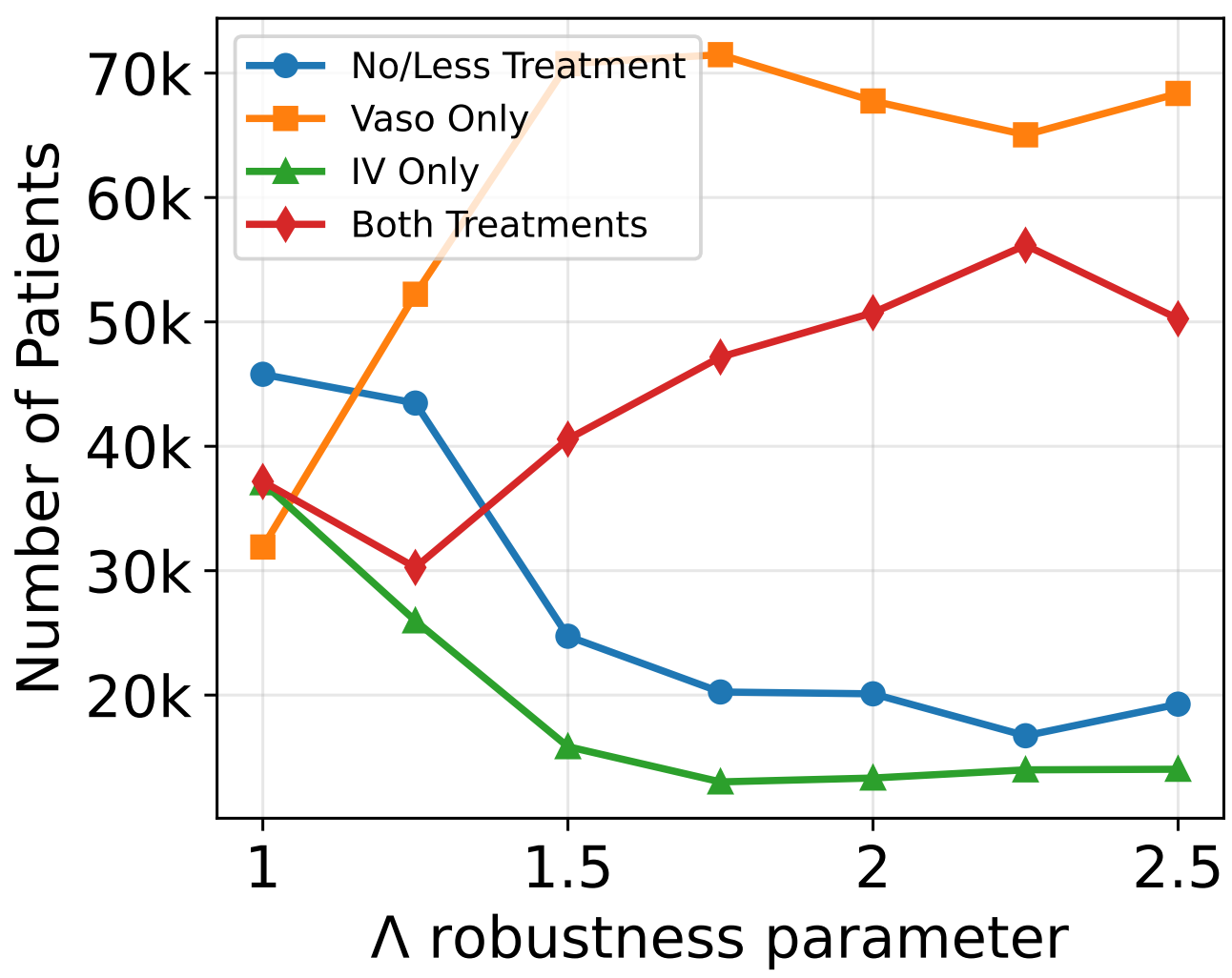

(a) Counts of actions taken by the robust optimal policy vs. the sensitivity parameter Λ. To simplify visualization we coarsen action into four groups: no/less treatment, only IV fluid, only vasopressors, and both fluid and vasopressors.

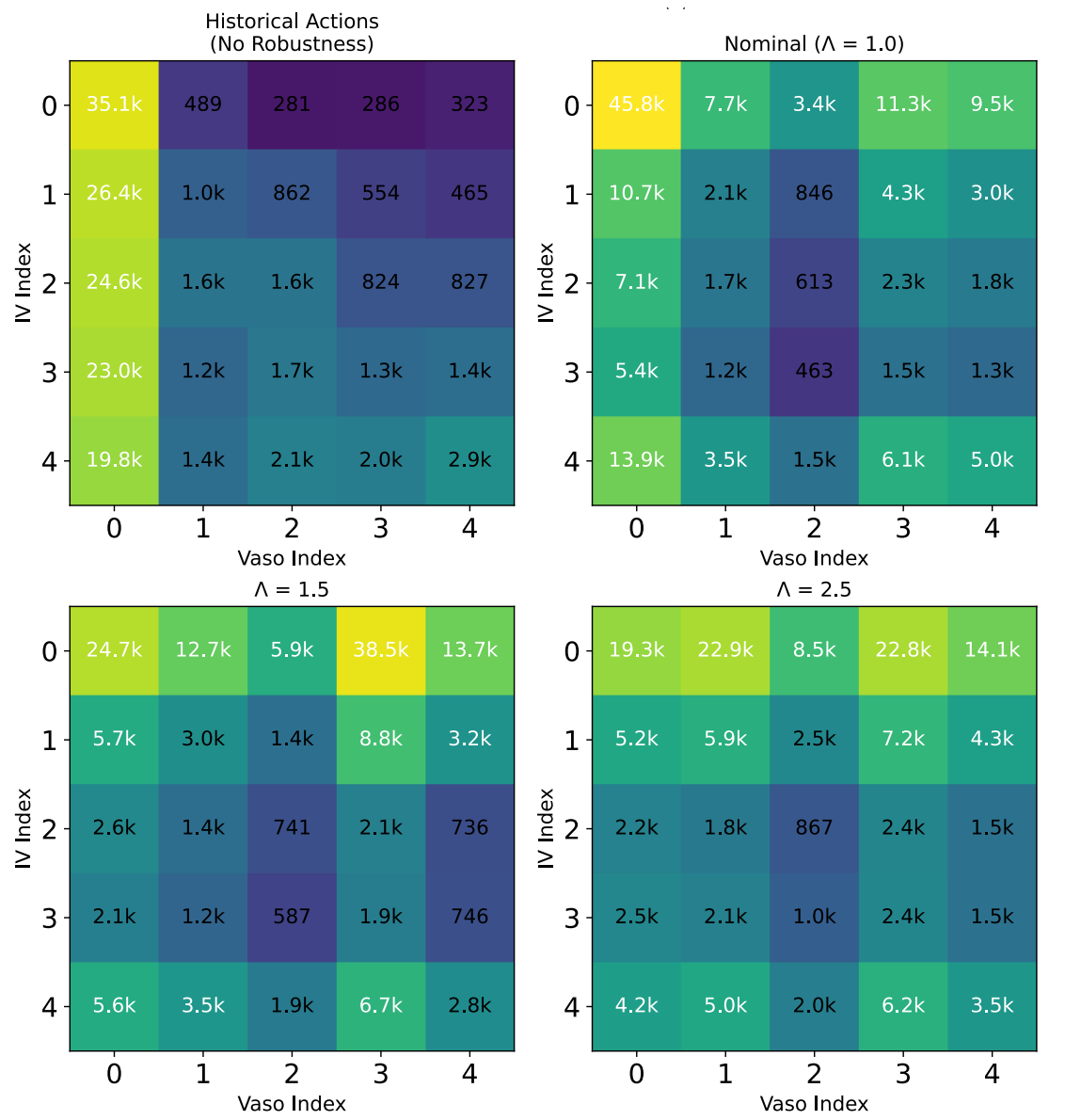

(b) Heatmaps of log(action counts) + 1. Yellow is higher, blue is lower count. IV action index is on the y-axis, vasopressor is on the x-axis; dosage increases in action index. Top left heatmap is historical actions, top right is nominal $\Lambda = 1$, and bottom $\Lambda = 1.5, 2.5$.

**Figure 2 Summary and heatmaps of optimal actions as $\Lambda$ increases.**

(Silversides et al. 2017)) is aligned with preferring vasopressors to IV fluids, where excessive usage might pose risks (Marik and Bellomo 2016, Semler et al. 2020).

## 7. Extension: offline-online RL

In the previous sections, we discussed obtaining robust bounds from offline data for robust-optimal policy learning, via fitted-Q-iteration. Our bounds can guide future randomized experimentation, even if the robust policy based on historical data alone is not deployed. We illustrate this via the following extension to warmstarting online RL.

Our ability to warmstate online learning with our offline bounds relies crucially on our early marginal MDP characterization. In the online setting, under our assumptions (Assumptions 1 and 2 or Definition 1 (memorylessness)), policies that don't use unobserved confounders generate (unconfounded) Markov decision processes. This is a crucial difference from handling general unobserved confounders in a general POMDP, where standard online RL algorithms don't apply.

In this section, we show how robust bounds can be used to warmstart a state-of-the-art reinforcement learning algorithm under linear function approximation, LSVI-UCB (Jin et al. 2020), a well-studied variant of least-squares value iteration (LSVI) (Bradtke and Barto 1996, Osband et al. 2016) using linear function approximation. By contrast, naively (non-robustly) warmstarting LSVI-UCB by using confounded offline data severely degrades online performance.

This extension is most closely related to recent papers that warmstart reinforcement learning from offline data with unobserved confounding, although these have been restricted to tabular settings. We provide a more extensive discussion in Section 6.2.

### 7.1. LSVI-UCB

We first introduce the basic setup of linear MDPs and LSVI-UCB (Jin et al. 2020). Our prior problem setup implies that the online counterpart over observed states is a Markov decision process. Further, we add functional form restrictions that the Q functions in the induced MDPs are linear and satisfy completeness. Let $\phi(s,a) : \mathcal{S} \times A \to \mathbb{R}^d$ be a feature map, and consider the function class $\mathcal{F}_{\text{lin}} := \{f(s,a) = \langle \theta, \phi(s,a) \rangle : \theta \in \mathbb{R}^d\}$.

ASSUMPTION 11 **(Linearity and Completeness).** *For any policy $\pi^e$ that is only a function of the observed state, the Q function is linear, $Q_t^{\pi^e} \in \mathcal{F}_{lin}, \forall t$. Furthermore, for all $f \in \mathcal{F}$, we have the completeness condition:*

$$g(s,a) = \mathbb{E}_{S_{t+1}}[R_t + \max_{A'} f(S_{t+1}, A') | S_t = s, A_t = a] \in \mathcal{F}_{lin}, \forall t.$$

Under these assumptions, the online LSVI-UCB procedure of Jin et al. (2020) has $\sqrt{T}$ total regret. But if $\pi^b$ does depend on the unobserved state[13], then the observed state transition probabilities will be biased in the offline dataset. Our confounding-robust bounds enable use of the offline dataset to warmstart LSVI-UCB, improving performance.

### 7.2. Warm-started LSVI-UCB

Here we outline the full algorithm for warm-starting LSVI-UCB presented in Algorithm 4. (Warm-starting other optimistic algorithms is essentially similar). The intuition is that the key step of LSVI-UCB, and other algorithms based on the principle of optimism under uncertainty, is planning according to the optimistic estimates of the value function, i.e. so that the estimated value function $\widehat{V}_t^n(s)$ satisfies that $\widehat{V}_t^n(s) \geq V_t^\star(s), \forall t, n, s, a$. This, in turn, bounds the per-episode regret by the difference between *optimistic* value function and true value function, $V_0^\star(s_0) - V_0^{\pi^n}(s_0) \leq \widehat{V}_0^n(s_0) - V_0^{\pi^n}(s_0)$. In the beginning, this difference is large due to sample uncertainty; but collecting more data over time shrinks the optimistic bonus and tends towards exploitation. Using the observational data, we can obtain *valid* robust bounds which can be used as a form of strong prior knowledge on the value function. That is, a basic idea is to truncate optimistic bounds by optimistic upper bounds over the confounded observational dataset. (Zhang and Bareinboim (2019) consider a similar approach but for tabular data). Truncating the optimistic bounds by prior knowledge 1) remains optimistic under valid bounds and 2) reduces the contribution of optimism to regret.

[13] If the offline policy $\pi^b$ is independent of the unobserved state $u$, then the online and offline MDPs are identical, and the setting reduces to one similar to Xie et al. (2021b).

We now describe the basic algorithm in more detail, while a full description is in Section 4.2. We run online LSVI-UCB, as in Jin et al. (2020) — each iteration we update our Q model and then collect a trajectory by taking actions that are optimal with respect to that Q model. The standard optimism bonus is $\xi\phi^T\Sigma_t^{-1}\phi$, where $\Sigma_t$ is the sample Gram matrix and $\xi$ is the width of the confidence interval; its value is derived theoretically but in practice it is often a hyperparameter. The key difference with standard LSVI-UCB is that at the start of each iteration, we run our robust FQE algorithm on the offline data to get robust upper bounds[14] on the Q function for the current policy, $\hat{\overline{Q}}$.

Thus, in each iteration we have two valid upper bounds on the Q function: the upper bound from the standard optimism bonus, and the upper bound from robust FQE on the offline data. For our warm-started LSVI-UCB, we choose whichever one is <u>smaller</u>. As a result, we retain the theoretical guarantees from optimism as proven in Jin et al. (2020), while possibly improving performance when $\hat{\overline{Q}}$ is sharper than the online upper confidence bound. To handle potential small-sample instabilities, we simply set the offline bonus to 0 when $\hat{\overline{Q}}_t(\cdot,\cdot) < \theta_t^\top \boldsymbol{\phi}(\cdot,\cdot)$.

Finally, note that in practice, we can compute the robust <u>optimal</u> Q parameters once at the start using robust FQI, before the online procedure begins. See Section 4.2 for more details.

### 7.3. Simulation Experiments with Warm-starting

We provide preliminary experiments to demonstrate two key points. First, warmstarting LSVI-UCB from our valid robust bounds can result in substantial performance gains compared to the purely online algorithm. Second, naively warm-starting LSVI-UCB (without robustness) from confounded offline data performs much <u>worse</u> compared to the purely online algorithm.

For offline-online simulations, we consider a linear-gaussian MDP with an unobserved confounder $U_t$, heteroskedastic rewards, where $\pi^b(A_t \mid S_t, U_t) = 1/2$ if $A_t = 3 - U_t, 1/6$ otherwise

[14] Note that since we want upper bounds instead of lower bounds, we compute the $\tau = \Lambda/(1+\Lambda)$ conditional quantile instead of the $1-\tau$ conditional quantile.

$\pi^b(A_t \mid S_t) = 1/4$. Then the smallest valid value of the MSM parameter is $\Lambda = 3$. (See Section 4.2 for more details).

Using this setup, we run the following three experiments: (1) standard LSVI-UCB without warm-starting, (2) warm-started LSVI-UCB using our robust bounds as in Algorithm 4, and (3) naive LSVI-UCB treating the offline data as unconfounded and continuing online.The third experiment is a (non-Bayesian) version of Algorithm 1 in Wang et al. (2021); but due to the unobserved confounders $U_t$, naive confidence intervals on offline data are invalid.

For all experiments, we use horizon $T = 4$, number of trajectories $K = 250$, and LSVI-UCB parameters $\xi = 0.07$ and $\lambda = 10^{-6}$.[15] See the Appendix for a discussion of results with different hyperparameters. We compare performance in terms of the cumulative regret: $\sum_{k=1}^{K}[V_0^*(s_0^k) - V_0^{\pi^k}(s_0^k)]$, where $V_t^*$ is the optimal value function.

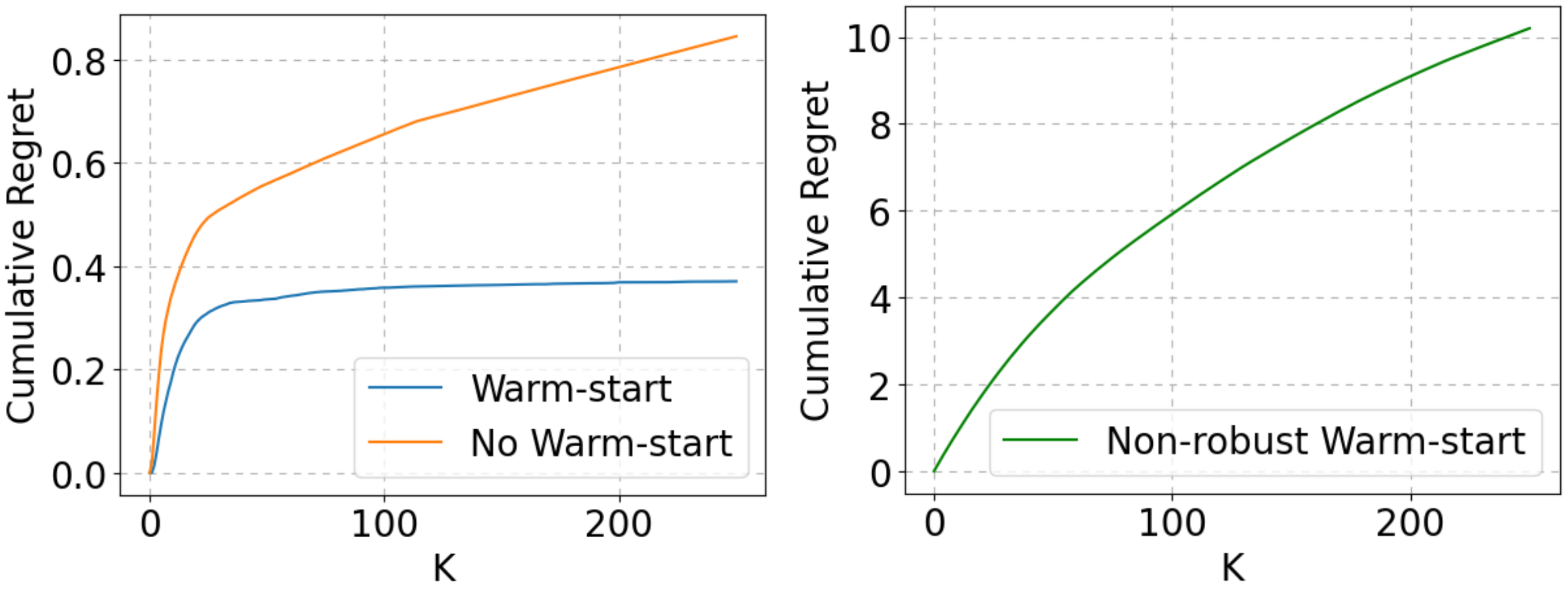

(a) (b)

**Figure 3 Simulation results for online LSVI-UCB. Cumulative regret is an average of over 200 trials. Panel (a) plots the cumulative regret of LSVI-UCB without warm-starting, and with robust warm-starting following Algorithm 4. Panel (b) plots the cumulative regret of LSVI-UCB where the offline data is naively treated as if had been collected online.**

We plot the results in Figure 3. The y-axis displays the cumulative regret averaged over 200 repeats of each algorithm. In Figure 3a, we compare the cumulative regret of LSVI-UCB without warm-starting and LSVI-UCB using our robust warm-starting algorithm. Our warm-started algorithm enjoys less than half the cumulative regret of standard LSVI-UCB after 250 online trajectories. In

[15] Note that $\xi$ has to be set sufficiently large for standard LSVI-UCB to have a valid upper confidence interval, whereas our warm-starting bounds will result in a valid interval regardless of $\xi$, providing some additional robustness to hyperparameter tuning.

Figure 3b, we show results for naive warm-starting from offline data.The cumulative regret after 250 trajectories is > 10 times higher than standard LSVI-UCB and > 20 times higher than robust warm-starting. The offline data misleads non-robust warm-starting to confidently choose the wrong action, and it takes a substantial amount of online data collection to correct this.

## 8. Conclusion

We developed a robust fitted-Q-iteration algorithm under memoryless unobserved confounders, leveraging function approximation, conditional quantiles, and orthogonalization. We derived sample complexity guarantees, demonstrated the effectiveness of our algorithm and the benefits of orthogonality in simulation experiments, and then provided a case-study with complex real-world healthcare data. Interesting directions for future work include falsifiability-based analyses to draw on competing identification proposals.

## References

Ali RA, Richardson TS, Spirtes P (2009) Markov equivalence for ancestral graphs .

Anand TV, Ribeiro AH, Tian J, Bareinboim E (2023) Causal effect identification in cluster dags. Proceedings of the AAAI Conference on Artificial Intelligence, volume 37, 12172–12179.

Aronow PM, Lee DK (2013) Interval estimation of population means under unknown but bounded probabilities of sample selection. Biometrika 100(1):235–240.

Auer P, Jaksch T, Ortner R (2008) Near-optimal regret bounds for reinforcement learning. Advances in neural information processing systems 21.

Behzadian B, Petrik M, Ho CP (2021) Fast algorithms for $l\infty$-constrained s-rectangular robust mdps. Advances in Neural Information Processing Systems 34:25982–25992.

Belloni A, Chernozhukov V (2011) l1-penalized quantile regression in high-dimensional sparse models. The Annals of Statistics 39(1):82–130.

Bennett A, Kallus N (2019) Policy evaluation with latent confounders via optimal balance. Advances in Neural Information Processing Systems, 4827–4837.

Bennett A, Kallus N (2024) Proximal reinforcement learning: Efficient off-policy evaluation in partially observed markov decision processes. Operations Research 72(3):1071–1086.

Bennett A, Kallus N, Li L, Mousavi A (2021) Off-policy evaluation in infinite-horizon reinforcement learning with latent confounders. International Conference on Artificial Intelligence and Statistics, 1999–2007 (PMLR).

Bennett A, Kallus N, Oprescu M, Sun W, Wang K (2024) Efficient and sharp off-policy evaluation in robust markov decision processes. Advances in Neural Information Processing Systems 37:112962–113000.

Bibaut AF, Malenica I, Vlassis N, van der Laan MJ (2019) More efficient off-policy evaluation through regularized targeted learning. arXiv preprint arXiv:1912.06292 .

Bonvini M, Kennedy E, Ventura V, Wasserman L (2022) Sensitivity analysis for marginal structural models. arXiv preprint arXiv:2210.04681 .

Bonvini M, Kennedy EH (2021) Sensitivity analysis via the proportion of unmeasured confounding. Journal of the American Statistical Association 1–11.

Bradtke SJ, Barto AG (1996) Linear least-squares algorithms for temporal difference learning. Machine learning 22:33–57.

Bruns-Smith DA (2021) Model-free and model-based policy evaluation when causality is uncertain. International Conference on Machine Learning, 1116–1126 (PMLR).

Cai Z, Wang X (2008) Nonparametric estimation of conditional var and expected shortfall. Journal of Econometrics 147(1):120–130.

Caro F, Gallien J, Díaz M, García J, Corredoira JM, Montes M, Ramos JA, Correa J (2010) Zara uses operations research to reengineer its global distribution process. Interfaces 40(1):71–84.

Chen J, Jiang N (2019) Information-theoretic considerations in batch reinforcement learning. International Conference on Machine Learning, 1042–1051 (PMLR).

Chen Q, Syrgkanis V, Austern M (2022) Debiased machine learning without sample-splitting for stable estimators. arXiv preprint arXiv:2206.01825 .

Chen S, Zhang B (2021) Estimating and improving dynamic treatment regimes with a time-varying instrumental variable. arXiv preprint arXiv:2104.07822 .

Chernozhukov V, Chetverikov D, Demirer M, Duflo E, Hansen C, Newey W, Robins J (2018) Double/debiased machine learning for treatment and structural parameters.

Chernozhukov V, Cinelli C, Newey W, Sharma A, Syrgkanis V (2022) Long story short: Omitted variable bias in causal machine learning. Technical report, National Bureau of Economic Research.

Chow Y, Tamar A, Mannor S, Pavone M (2015) Risk-sensitive and robust decision-making: a cvar optimization approach. Advances in Neural Information Processing Systems, 1522–1530.

Clarke WL, Anderson S, Kovatchev B (2008) Evaluating clinical accuracy of continuous glucose monitoring systems: continuous glucose-error grid analysis (cg-ega). Current diabetes reviews 4(3):193–199.

Delage E, Iancu DA (2015) Robust multistage decision making. The operations research revolution, 20–46 (INFORMS).

Denton BT (2018) Optimization of sequential decision making for chronic diseases: From data to decisions. Recent Advances in Optimization and Modeling of Contemporary Problems, 316–348 (INFORMS).

Dorn J, Guo K (2022) Sharp sensitivity analysis for inverse propensity weighting via quantile balancing. Journal of the American Statistical Association (just-accepted):1–28.

Dorn J, Guo K, Kallus N (2021) Doubly-valid/doubly-sharp sensitivity analysis for causal inference with unmeasured confounding. arXiv preprint arXiv:2112.11449 .

Duan Y, Jin C, Li Z (2021) Risk bounds and rademacher complexity in batch reinforcement learning. International Conference on Machine Learning, 2892–2902 (PMLR).

Ernst D, Stan GB, Goncalves J, Wehenkel L (2006) Clinical data based optimal sti strategies for hiv: a reinforcement learning approach. Proceedings of the 45th IEEE Conference on Decision and Control, 667–672 (IEEE).

Evans L, Rhodes A, Alhazzani W, Antonelli M, Coopersmith CM, French C, Machado FR, Mcintyre L, Ostermann M, Prescott HC, et al. (2021) Surviving sepsis campaign: international guidelines for management of sepsis and septic shock 2021. Intensive care medicine 47(11):1181–1247.

FDA (2021) Aiml_samd_action_plan. `https://www.fda.gov/media/145022/download`, (Accessed on 09/06/2023).

Feng T, Zhang Y (2017) Modeling strategic behavior in the competitive newsvendor problem: An experimental investigation. Production and Operations Management 26(7):1383–1398.

Foster DJ, Syrgkanis V (2019) Orthogonal statistical learning. arXiv preprint arXiv:1901.09036 .

Friedman JH (2001) Greedy function approximation: a gradient boosting machine. Annals of statistics 1189–1232.

Fu J, Norouzi M, Nachum O, Tucker G, Wang Z, Novikov A, Yang M, Zhang MR, Chen Y, Kumar A, et al. (2021) Benchmarks for deep off-policy evaluation. arXiv preprint arXiv:2103.16596 .

Fu Z, Qi Z, Wang Z, Yang Z, Xu Y, Kosorok MR (2022) Offline reinforcement learning with instrumental variables in confounded markov decision processes. arXiv preprint arXiv:2209.08666 .

Gottesman O, Johansson F, Komorowski M, Faisal A, Sontag D, Doshi-Velez F, Celi LA (2019) Guidelines for reinforcement learning in healthcare. Nature medicine 25(1):16–18.

Goyal V, Grand-Clement J (2022) Robust markov decision processes: Beyond rectangularity. Mathematics of Operations Research .

Han S (2022) Optimal dynamic treatment regimes and partial welfare ordering. Journal of American Statistical Association (just-accepted) .

Hastie T, Tibshirani R, Friedman JH, Friedman JH (2009) The elements of statistical learning: data mining, inference, and prediction, volume 2 (Springer).

Ho CP, Petrik M, Wiesemann W (2021) Partial policy iteration for l1-robust markov decision processes. J. Mach. Learn. Res. 22:275–1.

Howard RA (1960) Dynamic programming and markov processes. .

Hsu JY, Small DS (2013) Calibrating sensitivity analyses to observed covariates in observational studies. Biometrics 69(4):803–811.

Imani E, Graves E, White M (2018) An off-policy policy gradient theorem using emphatic weightings. Advances in Neural Information Processing Systems 31.

Iyengar GN (2005) Robust dynamic programming. Mathematics of Operations Research 30(2):257–280.

Jeong S, Namkoong H (2020) Assessing external validity over worst-case subpopulations. arXiv preprint arXiv:2007.02411 .

Jesson A, Douglas A, Manshausen P, Meinshausen N, Stier P, Gal Y, Shalit U (2022) Scalable sensitivity and uncertainty analysis for causal-effect estimates of continuous-valued interventions. arXiv preprint arXiv:2204.10022 .

Jiang N, Li L (2016) Doubly robust off-policy value evaluation for reinforcement learning. Proceedings of the 33rd International Conference on Machine Learning .

Jin C, Yang Z, Wang Z, Jordan MI (2020) Provably efficient reinforcement learning with linear function approximation. Conference on Learning Theory, 2137–2143 (PMLR).

Jin Y, Yang Z, Wang Z (2021) Is pessimism provably efficient for offline rl? International Conference on Machine Learning, 5084–5096 (PMLR).

Jordan MI, Wang Y, Zhou A (2022) Empirical gateaux derivatives for causal inference. arXiv preprint arXiv:2208.13701 .

Kallus N, Mao X, Zhou A (2018) Interval estimation of individual-level causal effects under unobserved confounding. arXiv preprint arXiv:1810.02894 .

Kallus N, Uehara M (2020a) Double reinforcement learning for efficient off-policy evaluation in Markov decision processes. Journal of Machine Learning Research 21(167):1–63.

Kallus N, Uehara M (2020b) Statistically efficient off-policy policy gradients. International Conference on Machine Learning, 5089–5100 (PMLR).

Kallus N, Zhou A (2020a) Confounding-robust policy evaluation in infinite-horizon reinforcement learning. arXiv preprint arXiv:2002.04518 .

Kallus N, Zhou A (2020b) Minimax-optimal policy learning under unobserved confounding. Management Science .

Kallus N, Zhou A (2022) Stateful offline contextual policy evaluation and learning. International Conference on Artificial Intelligence and Statistics, 11169–11194 (PMLR).

Kato K (2012) Weighted nadaraya–watson estimation of conditional expected shortfall. Journal of Financial Econometrics 10(2):265–291.

Kausik C, Lu Y, Tan K, Makar M, Wang Y, Tewari A (2024) Offline policy evaluation and optimization under confounding. International Conference on Artificial Intelligence and Statistics, 1459–1467 (PMLR).

Kennedy EH (2020) Optimal doubly robust estimation of heterogeneous causal effects. arXiv preprint arXiv:2004.14497 .

Kennedy EH (2022) Semiparametric doubly robust targeted double machine learning: a review. arXiv preprint arXiv:2203.06469 .

Killian TW, Zhang H, Subramanian J, Fatemi M, Ghassemi M (2020) An empirical study of representation learning for reinforcement learning in healthcare. arXiv preprint arXiv:2011.11235 .

Koenker R, Hallock KF (2001) Quantile regression. Journal of economic perspectives 15(4):143–156.

Komorowski M, Celi LA, Badawi O, Gordon AC, Faisal AA (2018) The artificial intelligence clinician learns optimal treatment strategies for sepsis in intensive care. Nature medicine 24(11):1716–1720.

Kremer M, Minner S, Van Wassenhove LN (2010) Do random errors explain newsvendor behavior? Manufacturing & Service Operations Management 12(4):673–681.

Kumar M, Dev S, Khalid MU, Siddenthi SM, Noman M, John C, Akubuiro C, Haider A, Rani R, Kashif M, et al. (2023) The bidirectional link between diabetes and kidney disease: mechanisms and management. Cureus 15(9).

Laan MJ, Robins JM (2003) Unified methods for censored longitudinal data and causality (Springer).

Larkin H (2023a) Vasopressors or high-volume iv fluids both effective for sepsis. JAMA 329(7):532–532.

Larkin M (2023b) Finding the Optimal Fluid Strategies for Sepsis — medscape.com. `https://www.medscape.com/viewarticle/993925?form=fpf#vp_2`, [Accessed 27-08-2023].

Le H, Voloshin C, Yue Y (2019) Batch policy learning under constraints. International Conference on Machine Learning, 3703–3712 (PMLR).

Lewis G, Syrgkanis V (2020) Double/debiased machine learning for dynamic treatment effects via g-estimation. arXiv preprint arXiv:2002.07285 .

Liao L, Fu Z, Yang Z, Wang Y, Kolar M, Wang Z (2021) Instrumental variable value iteration for causal offline reinforcement learning. arXiv preprint arXiv:2102.09907 .

Liu Q, Li L, Tang Z, Zhou D (2018) Breaking the curse of horizon: Infinite-horizon off-policy estimation. Advances in Neural Information Processing Systems, 5356–5366.

Liu S, See KC, Ngiam KY, Celi LA, Sun X, Feng M (2020) Reinforcement learning for clinical decision support in critical care: comprehensive review. Journal of medical Internet research 22(7):e18477.

Lobo EA, Ghavamzadeh M, Petrik M (2020) Soft-robust algorithms for batch reinforcement learning. arXiv preprint arXiv:2011.14495 .

Lu M, Shahn Z, Sow D, Doshi-Velez F, Li-wei HL (2020) Is deep reinforcement learning ready for practical applications in healthcare? a sensitivity analysis of duel-ddqn for hemodynamic management in sepsis patients. AMIA Annual Symposium Proceedings, volume 2020, 773 (American Medical Informatics Association).

Ma X, Liang Z, Xia L, Zhang J, Blanchet J, Liu M, Zhao Q, Zhou Z (2022) Distributionally robust offline reinforcement learning with linear function approximation. arXiv preprint arXiv:2209.06620 .

Malinsky D, Spirtes P (2016) Estimating causal effects with ancestral graph markov models. Conference on Probabilistic Graphical Models, 299–309 (PMLR).

Marik P, Bellomo R (2016) A rational approach to fluid therapy in sepsis. BJA: British Journal of Anaesthesia 116(3):339–349.

Meinshausen N (2006) Quantile regression forests. Journal of Machine Learning Research 7:983–999.

Miao R, Qi Z, Zhang X (2022) Off-policy evaluation for episodic partially observable markov decision processes under non-parametric models. arXiv preprint arXiv:2209.10064 .

Miratrix LW, Wager S, Zubizarreta JR (2018) Shape-constrained partial identification of a population mean under unknown probabilities of sample selection. Biometrika 105(1):103–114.

Namkoong H, Keramati R, Yadlowsky S, Brunskill E (2020) Off-policy policy evaluation for sequential decisions under unobserved confounding. arXiv preprint arXiv:2003.05623 .

Navaneethan SD, Zoungas S, Caramori ML, Chan JC, Heerspink HJ, Hurst C, Liew A, Michos ED, Olowu WA, Sadusky T, et al. (2023) Diabetes management in chronic kidney disease: synopsis of the kdigo 2022 clinical practice guideline update. Annals of internal medicine 176(3):381–387.

Newey WK (1994) The asymptotic variance of semiparametric estimators. Econometrica: Journal of the Econometric Society 1349–1382.

Newey WK, McFadden D (1994) Large sample estimation and hypothesis testing. Handbook of Econometrics 4:2111–2245.

Nilim A, El Ghaoui L (2005) Robust control of markov decision processes with uncertain transition matrices. Operations Research 53(5):780–798.

Norton M, Khokhlov V, Uryasev S (2021) Calculating cvar and bpoe for common probability distributions with application to portfolio optimization and density estimation. Annals of Operations Research 299(1):1281–1315.

Oberst M, Sontag D (2019) Counterfactual off-policy evaluation with gumbel-max structural causal models. International Conference on Machine Learning, 4881–4890 (PMLR).

Olma T (2021) Nonparametric estimation of truncated conditional expectation functions. arXiv preprint arXiv:2109.06150 .

Orellana L, Rotnitzky A, Robins JM (2010) Dynamic regime marginal structural mean models for estimation of optimal dynamic treatment regimes, part i: main content. The international journal of biostatistics 6(2).

Osband I, Van Roy B, Wen Z (2016) Generalization and exploration via randomized value functions. International Conference on Machine Learning, 2377–2386 (PMLR).

Panaganti K, Xu Z, Kalathil D, Ghavamzadeh M (2022) Robust reinforcement learning using offline data. arXiv preprint arXiv:2208.05129 .

Pearl J (2009) Causality (Cambridge university press).

Pearl J, Glymour M, Jewell NP (2016) Causal inference in statistics: A primer (John Wiley & Sons).

Raghu A, Komorowski M, Ahmed I, Celi L, Szolovits P, Ghassemi M (2017a) Deep reinforcement learning for sepsis treatment. arXiv preprint arXiv:1711.09602 .

Raghu A, Komorowski M, Celi LA, Szolovits P, Ghassemi M (2017b) Continuous state-space models for optimal sepsis treatment: a deep reinforcement learning approach. Machine Learning for Healthcare Conference, 147–163 (PMLR).

Raghu A, Komorowski M, Singh S (2018) Model-based reinforcement learning for sepsis treatment. arXiv preprint arXiv:1811.09602 .

Rashidinejad P, Zhu B, Ma C, Jiao J, Russell S (2021) Bridging offline reinforcement learning and imitation learning: A tale of pessimism. Advances in Neural Information Processing Systems 34:11702–11716.

Richardson T, Spirtes P (2002) Ancestral graph markov models. The Annals of Statistics 30(4):962–1030.

Robins JM, Rotnitzky A, Scharfstein DO (2000) Sensitivity analysis for selection bias and unmeasured confounding in missing data and causal inference models. Statistical models in epidemiology, the environment, and clinical trials, 1–94 (Springer).

Rockafellar RT, Uryasev S, et al. (2000) Optimization of conditional value-at-risk. Journal of risk 2:21–42.

Rosenbaum PR (2004) Design sensitivity in observational studies. Biometrika 91(1):153–164.

Rosenstrom E, Meshkinfam S, Ivy JS, Goodarzi SH, Capan M, Huddleston J, Romero-Brufau S (2022) Optimizing the first response to sepsis: An electronic health record-based markov decision process model. Decision Analysis 19(4):265–296.

Saghafian S (2021) Ambiguous dynamic treatment regimes: A reinforcement learning approach. arXiv preprint arXiv:2112.04571 .

Scharfstein D, McDermott A, Diaz I, Carone M, Lunardon N, Turkoz I (2018) Global sensitivity analysis for repeated measures studies with informative drop-out: A semi-parametric approach. Biometrics 74(1):207–219.

Scharfstein DO, Nabi R, Kennedy EH, Huang MY, Bonvini M, Smid M (2021) Semiparametric sensitivity analysis: Unmeasured confounding in observational studies. arXiv preprint arXiv:2104.08300 .

Semenova V (2017) Debiased machine learning of set-identified linear models. arXiv preprint arXiv:1712.10024 .

Semenova V (2023) Debiased machine learning of set-identified linear models. Journal of Econometrics .

Semenova V, Chernozhukov V (2021) Debiased machine learning of conditional average treatment effects and other causal functions. The Econometrics Journal 24(2):264–289.

Semler MW, Janz DR, Casey JD, Self WH, Rice TW (2020) Conservative fluid management after sepsis resuscitation: a pilot randomized trial. Journal of intensive care medicine 35(12):1374–1382.

Shah RD, Peters J (2020) The hardness of conditional independence testing and the generalised covariance measure .

Shi C, Uehara M, Huang J, Jiang N (2022a) A minimax learning approach to off-policy evaluation in confounded partially observable markov decision processes. International Conference on Machine Learning, 20057–20094 (PMLR).

Shi C, Wan R, Song R, Lu W, Leng L (2020) Does the markov decision process fit the data: Testing for the markov property in sequential decision making. International Conference on Machine Learning, 8807–8817 (PMLR).

Shi C, Zhu J, Shen Y, Luo S, Zhu H, Song R (2024) Off-policy confidence interval estimation with confounded markov decision process. Journal of the American Statistical Association 119(545):273–284.

Shi C, Zhu J, Ye S, Luo S, Zhu H, Song R (2022b) Off-policy confidence interval estimation with confounded markov decision process. Journal of the American Statistical Association 1–12.

Silversides JA, Major E, Ferguson AJ, Mann EE, McAuley DF, Marshall JC, Blackwood B, Fan E (2017) Conservative fluid management or deresuscitation for patients with sepsis or acute respiratory distress syndrome following the resuscitation phase of critical illness: a systematic review and meta-analysis. Intensive care medicine 43(2):155–170.

Simchowitz M, Mania H, Tu S, Jordan MI, Recht B (2018) Learning without mixing: Towards a sharp analysis of linear system identification. Conference On Learning Theory, 439–473 (PMLR).

Sinclair SR, Frujeri FV, Cheng CA, Marshall L, Barbalho HDO, Li J, Neville J, Menache I, Swaminathan A (2023) Hindsight learning for mdps with exogenous inputs. International Conference on Machine Learning, 31877–31914 (PMLR).

Singh R, Syrgkanis V (2022) Automatic debiased machine learning for dynamic treatment effects. arXiv preprint arXiv:2203.13887 .

Song Y, Zhou Y, Sekhari A, Bagnell JA, Krishnamurthy A, Sun W (2022) Hybrid rl: Using both offline and online data can make rl efficient. arXiv preprint arXiv:2210.06718 .

Spirtes P, Glymour CN, Scheines R (2000) Causation, prediction, and search (MIT press).

Su X (2008) Bounded rationality in newsvendor models. Manufacturing & Service Operations Management 10(4):566–589.

Tan Z (2012) A distributional approach for causal inference using propensity scores. Journal of the American Statistical Associatioon .

Tang Z, Feng Y, Li L, Zhou D, Liu Q (2019) Doubly robust bias reduction in infinite horizon off-policy estimation. arXiv preprint arXiv:1910.07186 .

Tennenholtz G, Mannor S, Shalit U (2019) Off-policy evaluation in partially observable environments. arXiv preprint arXiv:1909.03739 .

Tennenholtz G, Shalit U, Mannor S, Efroni Y (2021) Bandits with partially observable confounded data. Uncertainty in Artificial Intelligence, 430–439 (PMLR).

Thomas P, Theocharous G, Ghavamzadeh M (2015) High confidence policy improvement. International Conference on Machine Learning, 2380–2388.

Uber Privacy Center (????) URL `https://www.uber.com/global/en/privacy/overview/?uclick_id=1272f334-5389-40cf-8dbe-bd0f3c8a0695`.

Uehara M, Kiyohara H, Bennett A, Chernozhukov V, Jiang N, Kallus N, Shi C, Sun W (2022) Future-dependent value-based off-policy evaluation in pomdps. arXiv preprint arXiv:2207.13081 .

Uhler C, Raskutti G, Bühlmann P, Yu B (2013) Geometry of the faithfulness assumption in causal inference. The Annals of Statistics 436–463.

van de Vaart A, Wellner J (1996) Weak Convergence and Empirical Processes: With Applications to Statistics (Springer Mathematics).

VanderWeele TJ, Ding P (2017) Sensitivity analysis in observational research: introducing the e-value. Annals of internal medicine 167(4):268–274.

Voloshin C, Le HM, Jiang N, Yue Y (2019) Empirical study of off-policy policy evaluation for reinforcement learning. arXiv preprint arXiv:1911.06854 .

Wachtel RE, Dexter F (2010) Review of behavioral operations experimental studies of newsvendor problems for operating room management. Anesthesia & Analgesia 110(6):1698–1710.

Wainwright MJ (2019) High-dimensional statistics: A non-asymptotic viewpoint, volume 48 (Cambridge university press).

Wang J, Qi Z, Shi C (2022) Blessing from human-ai interaction: Super reinforcement learning in confounded environments. arXiv preprint arXiv:2209.15448 .

Wang L, Yang Z, Wang Z (2021) Provably efficient causal reinforcement learning with confounded observational data. Advances in Neural Information Processing Systems 34:21164–21175.

Wang S, Si N, Blanchet J, Zhou Z (2023a) A finite sample complexity bound for distributionally robust q-learning. International Conference on Artificial Intelligence and Statistics, 3370–3398 (PMLR).

Wang S, Si N, Blanchet J, Zhou Z (2023b) On the foundation of distributionally robust reinforcement learning. arXiv preprint arXiv:2311.09018 .

Wang S, Si N, Blanchet J, Zhou Z (2024) Sample complexity of variance-reduced distributionally robust q-learning. Journal of Machine Learning Research 25(341):1–77.

Xie T, Cheng CA, Jiang N, Mineiro P, Agarwal A (2021a) Bellman-consistent pessimism for offline reinforcement learning. Advances in neural information processing systems 34:6683–6694.

Xie T, Jiang N (2020) Q* approximation schemes for batch reinforcement learning: A theoretical comparison. Conference on Uncertainty in Artificial Intelligence, 550–559 (PMLR).

Xie T, Jiang N, Wang H, Xiong C, Bai Y (2021b) Policy finetuning: Bridging sample-efficient offline and online reinforcement learning. Advances in neural information processing systems 34:27395–27407.

Xu W, Ma Y, Xu K, Bastani H, Bastani O (2023a) Uniformly conservative exploration in reinforcement learning. International Conference on Artificial Intelligence and Statistics, 10856–10870 (PMLR).

Xu Y, Zhu J, Shi C, Luo S, Song R (2023b) An instrumental variable approach to confounded off-policy evaluation. International Conference on Machine Learning, 38848–38880 (PMLR).

Yadlowsky S, Namkoong H, Basu S, Duchi J, Tian L (2018) Bounds on the conditional and average treatment effect in the presence of unobserved confounders. arXiv preprint arXiv:1808.09521 .

Yang S, Lok JJ (2018) Sensitivity analysis for unmeasured confounding in coarse structural nested mean models. Statistica Sinica 28(4):1703.

Yang W, Zhang L, Zhang Z (2022) Toward theoretical understandings of robust markov decision processes: Sample complexity and asymptotics. The Annals of Statistics 50(6):3223–3248.

Zhang J, Bareinboim E (2019) Near-optimal reinforcement learning in dynamic treatment regimes. Advances in Neural Information Processing Systems, 13401–13411.

Zhang Y, Jiang N (2024) On the curses of future and history in future-dependent value functions for off-policy evaluation. Advances in Neural Information Processing Systems 37:124756–124790.

Zhang Y, McCoy RG, Mason JE, Smith SA, Shah ND, Denton BT (2014) Second-line agents for glycemic control for type 2 diabetes: are newer agents better? Diabetes Care 37(5):1338–1345.

Zhao Q, Small DS, Bhattacharya BB (2019) Sensitivity analysis for inverse probability weighting estimators via the percentile bootstrap. Journal of the Royal Statistical Society: Series B (Statistical Methodology) 81(4):735–761.

Zhao YQ, Zeng D, Laber EB, Kosorok MR (2015) New statistical learning methods for estimating optimal dynamic treatment regimes. Journal of the American Statistical Association 110(510):583–598.

Zhou Z, Zhou Z, Bai Q, Qiu L, Blanchet J, Glynn P (2021) Finite-sample regret bound for distributionally robust offline tabular reinforcement learning. International Conference on Artificial Intelligence and Statistics, 3331–3339 (PMLR).

*Table of contents*

- Section A includes discussions of graphical causal inference to characterize the marginal transition probabilities.
- Section B discusses the characterizations in Section 2, vis-a-vis Assumptions 1 and 2.
- Section 3 includes proofs of results of the marginal MDP.
- Section 4 includes additional variants of the main algorithm (cross-fitting, infinite-horizon), and warmstarting
- Section 6 includes additional discussion.
- Section 7 includes proofs of analysis of robust FQI.
- Section 8 contains additional detail on computational experiments.

## Appendix A: Graphical Causal Inference and Faithfulness: Preliminaries

We state some preliminaries on graphical causal inference that are unnecessary for the development of the main methodology, but underlie the identification analysis. The connection between nonparametric structural causal models and (offline) reinforcement learning has been well-established, and we do not provide full details here. See Tennenholtz et al. (2019, Appendix C), (Oberst and Sontag 2019).

We introduce concepts from directed acyclical graphical models (DAGs). Our discussion of preliminaries will use standard notation in graphical causal inference, potentially unrelated to the rest of the paper. A graph $\mathcal{G} = (V, E)$ over a set of nodes $V$ contains at most one directed edge between any pair of nodes. In a directed graph, an edge from node $X$ to node $Y$ is denoted $X \to Y$.

Definition 7 (Path). Given a node set $V$ and an edge set $E$, we define a path $p_{XY}$ from node $X$ to node $Y$ with $X, Y \in V, X \neq Y$, as a sequence of edges $p_{XY} = (e_1, \dots, e_\ell)$ such that $e_k \in E$ for all $1 \leq k \leq \ell, e_1$ starts with node $X, e_\ell$ ends with node $Y$, consecutive edges are connected, and nodes on the path do not repeat (other than as start- and endpoint of consecutive edges).

A directed path from $X$ to $Y$ is then a path where all edges point toward $Y$. Any node connected by a directed path from $X$ is a descendant of $X$, any node connected by a directed path to $X$ is an ancestor of $X$. Parents and children of a node $X$ are the direct causes and effects, respectively, of $X$ in $\mathcal{G}$. A directed acyclic graph (DAG) is a directed graph in which there is no pair of distinct nodes $(X, Y)$ such that there is a directed path from $X$ to $Y$ and an edge $Y \to X$. We say that a node $X$ is a collider on a path if its adjacent edges point into $X$ (e.g., $\to X \leftarrow$). A noncollider on a path is a node $X$ that is either a mediator (e.g., $\to X \to$) or a common cause (e.g., $\leftarrow X \to$).

Definition 8 ($d$-separation (as exposited in (Pearl et al. 2016))). A path $p$ is blocked by a set of nodes $\mathbf{Z}$ if and only if

- $p$ contains a chain of nodes $A \to B \to C$ or a fork $A \leftarrow B \to C$ such that the middle node $B$ is in $\mathbf{Z}$ (i.e., $B$ is conditioned on), or
- $p$ contains a collider $A \to B \leftarrow C$ such that the collision node $B$ is not in $\mathbf{Z}$, and no descendant of $B$ is in $\mathbf{Z}$.

If $\mathbf{Z}$ blocks every path between two nodes $X$ and $Y$, then $X$ and $Y$ are $d$-separated, conditional on $\mathbf{Z}$, and thus are independent conditional on $\mathbf{Z}$.

The d-separation condition informs of what variables in $\mathcal{G}$ must be independent, conditional on which other variables. We say that if $X \neq Y$, and $X$ and $Y$ are not in $\mathbf{W}$, then $X$ and $Y$ are d-connected given set $\mathbf{W}$ if and only if they are not d-separated given $\mathbf{W}$.

The criterion extends naturally to d-separating *sets* of vertices. If $\mathbf{U}, \mathbf{V}$, and $\mathbf{W}$ are disjoint sets of vertices in $\mathcal{G}$ and $\mathbf{U}$ and $\mathbf{V}$ are not empty then we say that $\mathbf{U}$ and $\mathbf{V}$ are d-separated given $\mathbf{W}$ if and only if every pair $\langle U, V \rangle$ in the cartesian product, $\mathbf{U} \times \mathbf{V}$, is d-separated given $\mathbf{W}$. If $\mathbf{U}, \mathbf{V}$ and $\mathbf{W}$ are disjoint sets of vertices in $\mathcal{G}$ and $\mathbf{U}$ and $\mathbf{V}$ are not empty then we say that $\mathbf{U}$ and $\mathbf{V}$ are d-connected given $\mathbf{W}$ if and only if $\mathbf{U}$ and $\mathbf{V}$ are not d-separated given $\mathbf{W}$. (Anand et al. 2023) builds on this "cluster" characterization. Our later analysis also leverages d-separation between *sets* of vertices, namely observed states and unobserved confounders at different timesteps.

Two key assumptions relate d-separation in the graph with probabilistic conditional independences. The first is often referred to as the causal Markov condition.

DEFINITION 9 (MARKOV CONDITION). If node $X$ is d-separated from node $Y$ given conditioning set $\mathbf{C}$ in graph $\mathcal{G} = (V, E)$ with $X, Y \in V$ and $\mathbf{C} \subseteq V \backslash \{X, Y\}$, then $X$ is probabilistically independent of $Y$ given $\mathbf{C}$ in the distribution over the graph $P_{\mathcal{G}}(V)$ :

$$X \perp Y \mid \mathbf{C} \text{ in } \mathcal{G} \quad \Longrightarrow \quad X \perp\!\!\!\perp Y \mid \mathbf{C} \text{ in } P_{\mathcal{G}}(V)$$

The Causal Markov condition holds by definition of the probability distribution represented by the causal graph. It asserts that a d-separation in the graph corresponds to a conditional independence in the resulting probability distribution.

In order to make conclusions about the underlying causal graphical structure, given observed conditional independences in the data, we need an additional assumption, called *faithfulness*. Faithfulness is a converse of the Markov property. It ensures that observed conditional independences in the observational distribution correspond to d-separations in the underlying causal graph, rather than circumstantial path or edge cancellations.

ASSUMPTION 12 (**Faithfulness**). *If variable $X$ is probabilistically independent of variable $Y$ given conditioning set $\mathbf{C}$ in the distribution over the graph $P_{\mathcal{G}}(V)$, then $X$ is d-separated from $Y$ given $\mathbf{C}$ in graph $\mathcal{G} = (V, E)$ :*

$$X \perp\!\!\!\perp Y \mid \mathbf{C} \textit{ in } P_{\mathcal{G}}(V) \quad \Longrightarrow \quad X \perp Y \mid \mathbf{C} \textit{ in } \mathcal{G}.$$

Causal Markov condition and faithfulness together imply that a (conditional) d-separation exists in the DAG if and only if a corresponding (conditional) independence exists in the probability distribution. Faithfulness is an assumption that is essentially required for causal discovery (Spirtes et al. 2000), or the related subtask of ascertaining graphical structure in the DGP based on observed conditional independences in data. It rules out edge or path cancellations that can be viewed as circumstantial edge cases. However, like many other assumptions in causal inference, it is untestable; and prior analysis shows that these happenstance path cancellations nonetheless occupy nontrivial volume in Lebesgue measure (Uhler et al. 2013).

### A.1. Maximal Ancestral Graphs

*Maximal ancestral graphs* (Richardson and Spirtes 2002, MAGs) characterize the conditional independences within an observational distribution that is marginalized over latent variables - exactly the case of the marginalized transition probability (see further discussions in (Malinsky and Spirtes 2016, Ali et al. 2009)). This provides a very straightforward way to check, given an underlying DAG on the full-information state space, whether or not the observed-state distribution will be Markovian. We now give a quick overview of MAGs along with an illustrative example. A MAG (Maximal Ancestral Graph) can have directed and bidirected edges: $\rightarrow$ and $\leftrightarrow$. (The original formulation allows undirected edges to indicate selection, but we will not use them here). Bidirected edges ultimately indicate observed dependence between variables that could be due to unobserved confounders. A MAG represents a DAG after all latent variables have been marginalized out, and it preserves all entailed conditional independence relations among the measured variables which are true in the underlying DAG. In a MAG $\mathcal{M}$, a tail mark at $X_i$ (e.g., $X_i \rightarrow X_j$ ) means that $X_i$ is an ancestor of $X_j$ in all DAGs represented by $\mathcal{M}$. An arrowhead at $X_i$ (e.g., $X_i \leftarrow X_j$ or $X_i \leftrightarrow X_j$ ) means that $X_i$ is not an ancestor of $X_j$ in all DAGs represented by $\mathcal{M}$. A $\leftrightarrow$ edge between two variables indicates that neither variable is an ancestor of the other (though they are probabilistically dependent). Maximal ancestral graphs are *maximal* in the sense that no additional edge may be added to the graph without changing the independence model. An inducing path $\pi$ relative to a set $\mathbf{L}$, between vertices $X$ and $Y$ in an ancestral graph $\mathcal{G}$, is a path on which every nonendpoint vertex not in $\mathbf{L}$ is both a collider on $\pi$ and an ancestor of at least one of the endpoints, $X$ and $Y$. One can construct the MAG $\mathcal{M}$ from the DAG $\mathcal{G} = \mathcal{G}(V, \mathbf{L})$ by the following procedure:

- For each pair $X, Y \in V$, $X$ and $Y$ are adjacent in $\mathcal{M}$ iff there is an inducing path between them relative to $\mathbf{L}$ in $\mathcal{G}$.
- For each adjacent pair $(X, Y)$ in $\mathcal{M}$, orient $X \to Y$ in $\mathcal{M}$ if $X \in \mathrm{an}_{\mathcal{G}}(Y)$; orient $Y \to X$ in $\mathcal{M}$ if $Y \in \mathrm{an}_{\mathcal{G}}(X)$; orient $X \leftrightarrow Y$ otherwise.

An inducing path $\pi$ relative to a set $\boldsymbol{L}$, between vertices $X$ and $Y$ in an ancestral graph $\mathcal{G}$, is a path on which every nonendpoint vertex not in $L$ is both a collider on $\pi$ and an ancestor of at least one of the endpoints, $X$ and $Y$. One can construct the MAG $\mathcal{M}$ from the DAG $\mathcal{G} = \mathcal{G}(V, L)$ by the following procedure:

- For each pair $a, b \in V, a$ and $b$ are adjacent in $\mathcal{M}$ iff there is an inducing path between them relative to $L$ in $\mathcal{G}$.
- For each adjacent pair $(a, b)$ in $\mathcal{M}$, orient $a \to b$ in $\mathcal{M}$ if $a \in$ an $_{\mathcal{G}}(b)$; orient $b \to a$ in $\mathcal{M}$ if $b \in \mathrm{an}_{\mathcal{G}}(a)$; orient $a \leftrightarrow b$ otherwise.

### A.2. Example: the DAG for full-information MDP

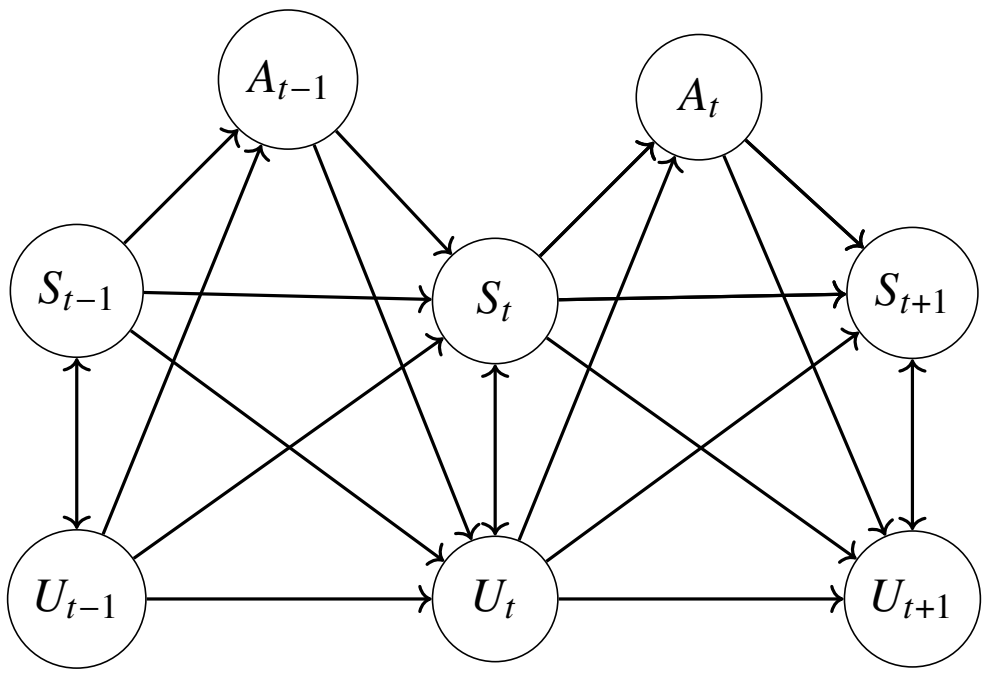

**Figure 4 Full Information MDP**

In what follows, we represent the observed state as a cluster state, readily obtained by concatenating different state dimensions. See (Anand et al. 2023) for discussion of such "cluster ADMGs".

### A.3. What is a faithfulness violation? Path Cancellation

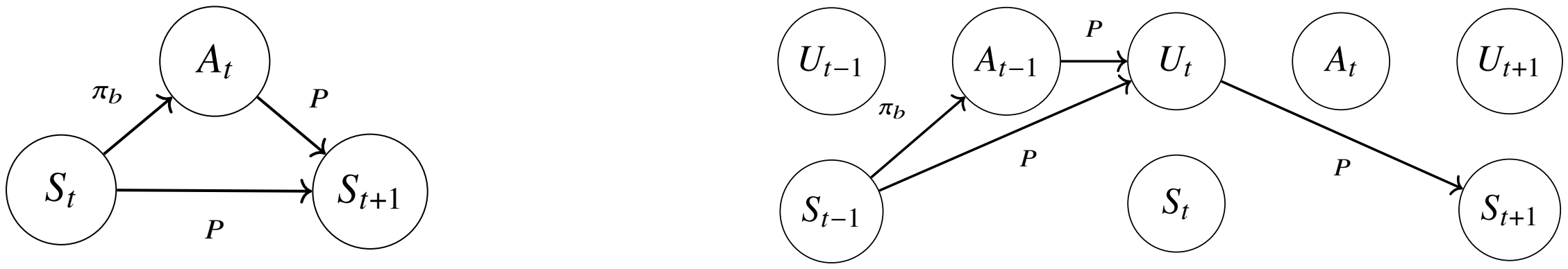

(a) Simple path cancellation (b) Path cancellation with $U_t$

**Figure 5 Causal graphical models for MDPs with path cancellation. We label the edges to emphasize which relationships are a function of the transition probabilities $P$ and which are a function of the policy $\pi_b$.**

To illustrate the basic concept, in the simplified example in Figure 5a, there is a direct effect of $A_t$ on $S_{t+1}$ and of $S_t$ on $S_{t+1}$. However, the policy $\pi_b$ can be chosen to make $S_t \perp\!\!\!\perp S_{t+1}$. If we used a different policy in the same MDP, we could have $S_t \not\perp\!\!\!\perp S_{t+1}$. The same idea applies to Figure 5b, which is more relevant to the observed-state Markov property: $\pi_b$ could be chosen to make $U_t \perp\!\!\!\perp S_{t-1}|S_t, A_t$ and therefore $S_{t+1} \perp\!\!\!\perp S_{t-1}|S_t, A_t$. However, under a different policy, we could have $S_{t+1} \not\perp\!\!\!\perp S_{t-1}|S_t, A_t$. Typically, the $\pi_b$ that would cause path cancellation have Lebesgue measure zero (Spirtes et al. 2000), and so these can be thought of as adversarially chosen policies.

*An example use of MAGs* We give a brief example of the use of MAGs to verify whether or not an underlying full-information MDP satisfies the observed-state Markov property. We will consider the particularly tricky question of the direction of the arrow between $U_t$ and $S_t$. Figure 6 displays the MAG corresponding to two models: the memoryless setting where $S_t \to U_t$ and $U_t$ is an unobserved confounder ($U_t \to S_{t+1}, U_t \to A_t$) in Figure 6a the alternative where instead $U_t \to S_t$. The memoryless model is the only underlying model on latents (when the graph structure is homogenous in time) that admits Markovian marginals, when policies are allowed to depend on $U_t$. (Recall that in our earlier characterization, we required assurance that transition probabilities were Markovian for all policies depending on $S_t$ alone). Flipping one edge ($U_t \to S_t$) introduces potential dependencies across timesteps and is a much more difficult POMDP regime. These dependencies are represented as adjacencies in the MAG (Figure 6d) between $(S_{t-1}, A_{t-1})$ and $S_{t+1}$, the red edges that indicate potential dependencies that result in non-Markovian observed marginals. First, consider the full-information causal DAG depicted in Figure 6a, where $U_t \to S_t, U_t \to A_t$, and $U_t \to S_{t+1}$, but $U_t \not\to U_{t+1}$, and this structure repeats at every timestep, i.e. is time-homogeneous. Then Assumption 1 holds and $P_{\pi^b}(S_{t+1} \mid S_t, A_t, H_t) = P_{\pi^b}(S_{t+1} \mid S_t, A_t)$, i.e. the observational distribution is Markovian. If all of the above holds but instead $S_t \to U_t$, then the observational distribution is not Markovian.

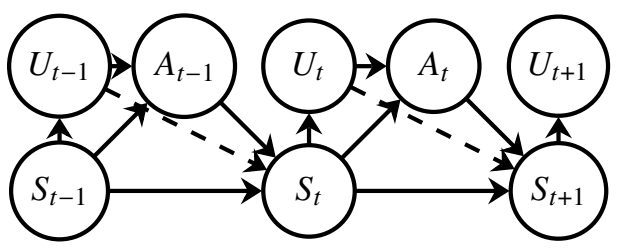

(a) DAG on latents and observables - emission model on unobserved confounders.

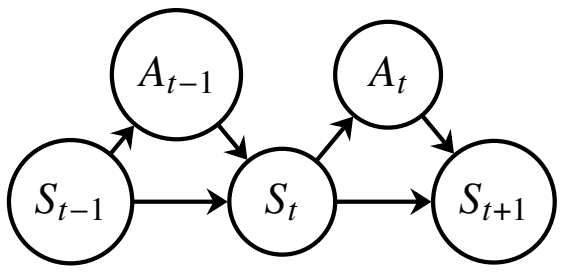

(b) Maximal Ancestral Graph (MAG) of Figure 6a. The marginals are Markovian.

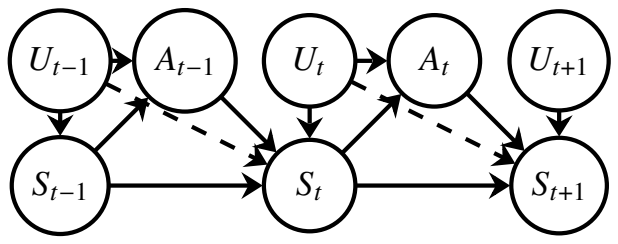

(c) DAG on latents and observables: general POMDP where $U_t \to S_t$.

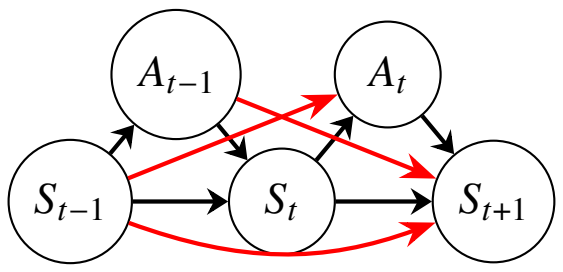

(d) MAG of model where $U_t \to S_t$ and is an UC.

**Figure 6 Underlying DAGs on time-homogenous and their latent projections to a maximal ancestral graph**

It is quite important that $S_t \to U_t$ but not the other way around. (The other way around means that observed variables become colliders on inducing paths). For example, in this model: There are no inducing paths between $A_t, A_{t'}$ because inducing paths must go through $S_t$, which is not a collider on the inducing path (because of the forward arrows under the MDP system, and that $S_t \to U_t$ but not vice versa. Similarly, there are no inducing paths between $A_t, S_{t'}$ for $t' > t$. There are no inducing paths between $S_t, S_{t'}$ for $|t - t'| \geq 2$ (state variables at least two timesteps apart) because any such inducing path needs to go through observed variable $S_t$, which is not a collider. The temporal ordering orients the edges of the MAG.

## Appendix B: Memoryless Confounders vs Our General Setting

*Memoryless Implies Observed-State Markov* We begin by showing that under memoryless confounding, Assumption 1 always holds. This is the case in both the time-homogeneous or time-inhomogeneous settings. The argument is straightforward. Consider any policy $\pi$:

$$\begin{aligned}
P_\pi(A_t|S_t) &= \int \pi_t(A_t|S_t,U_t)P_\pi(U_t|S_t)dU_t \\
&= \int \pi_t(A_t|S_t,U_t)P_t(U_t|S_t,S_{t-1},U_{t-1},A_{t-1})dU_t \\
&= \int \pi_t(A_t|S_t,U_t,H_t)P_t(U_t|S_t,S_{t-1},U_{t-1},A_{t-1},H_{t-1})dU_t \\
&= P_\pi(A_t|S_t,H_t).
\end{aligned}$$

The second line follows by the memoryless confounding condition, and the third line follows because the policy $\pi_t$ and the transitions $P_t$ are independent of the past. The last line simply reverses the argument but now conditional on $H_t$. Similarly, we have:

$$\begin{aligned}
P_\pi(S_{t+1}|S_t,A_t) &= \int P_t(S_{t+1}|S_t,U_t,A_t)P_\pi(U_t|S_t)dU_t \\
&= \int P_t(S_{t+1}|S_t,U_t,A_t)\frac{\pi_t(A_t|S_t,U_t)}{P_\pi(A_t|S_t)}P_\pi(U_t|S_t)dU_t \\
&= \int P_t(S_{t+1}|S_t,U_t,A_t)\frac{\pi_t(A_t|S_t,U_t)}{P_\pi(A_t|S_t)}P_t(U_t|S_t,S_{t-1},U_{t-1},A_{t-1})dU_t \\
&= \int P_t(S_{t+1}|S_t,U_t,A_t,H_t)\frac{\pi_t(A_t|S_t,U_t,H_t)}{P_\pi(A_t|S_t,H_t)}P_t(U_t|S_t,S_{t-1},U_{t-1},A_{t-1},H_{t-1})dU_t \\
&= P_\pi(S_{t+1}|S_t,A_t,H_t).
\end{aligned}$$

The second line follows from Bayes rule. The rest of the argument is the same as above, except in the fourth line we additionally use the fact that we already established $P_\pi(A_t|S_t) = P_\pi(A_t|S_t,H_t)$. Note that memoryless confounding doesn't imply Assumption 2, because the memoryless condition is a fact about conditional independence not about an underlying DAG. One could theoretically have an unfaithful graph (with path cancellation for example) that satisfies memoryless confounding. However, faithfulness is an assumption aimed at tying observed conditional independences to the underlying structure, and memoryless confounders is a condition on unobserved variables, so it would be unusual to discuss whether or not such a condition was faithful.

### B.1. Time-Homogeneous Setting

Now we consider what happens when we assume Assumptions 1 and 2, and whether or not the resulting system has to have memoryless confounders. In the time-homogeneous case, we show that these assumptions are nearly equivalent to memoryless confounding. In the time-inhomogeneous case the situation is more complicated. We begin by proving that in the time-homogeneous case under Assumptions 1 and 2, either the confounders are memoryless or there is no confounding. All backdoor paths between $S_{t+1}$ and $S_{t-1}$ that are not blocked by $S_t$ and $A_t$ must pass through $U_t$, so we split the analysis into two halves.

*First, assume there is no unblocked backdoor path from $U_t$ to $S_{t+1}$.* In all such DAGs there is no edge $U_t \to S_{t+1}$. The only other possible paths are $U_t \to U_{t+1} \to S_{t+1}$ and $U_t \to U_{t+1} \leftarrow S_{t+1}$. These are blocked in the following cases: (i) there is no edge $U_t \to U_{t+1}$, no edge $S_t \to U_{t+1}$ or $A_t \to U_{t+1}$, and there may be an edge $S_t \to U_t$ (but not $U_t \to S_t$); (ii) there is no edge $U_t \to U_{t+1}$, no edge $U_t \to A_t$, and there may be an edge between $S_t$ and $U_t$ in either direction; (iii) there is an edge $U_t \to U_{t+1}$, but no edge $S_t \to U_{t+1}$ or $U_t \to A_t$ or $A_t \to U_{t+1}$, and no edge between $S_t$ and $U_t$ in either direction.

We illustrate these cases:

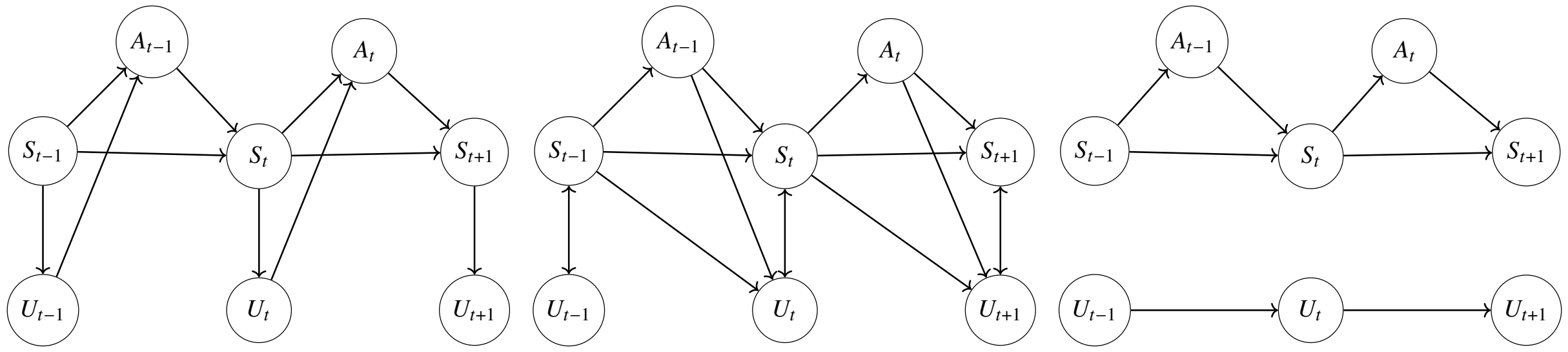

(a) Case 1 (b) Case 2 (c) Case 3

In Case 1, we cannot have an edge from $U_t \to S_t$ without inducing a collider. Conditioning on $S_t$ would generate a correlation between $S_{t-1}$ and $U_t$ (shared parents or $S_t$), inducing an unblocked backdoor path. In Case 3, note that we cannot have the edge from $S_t \to U_{t+1}$ because that would create the unblocked backdoor path $S_{t-1} \to U_t \to U_{t+1} \to U_{t+2} \leftarrow S_{t+1}$ (we leave $U_{t+2}$ out of the figure for space purposes). Note that Case 1 satisfies the memoryless condition. Cases 2 and 3 by contrast violate memorylessness. In Case 2, the observed state and action effect the next unobserved state, while in Case 3, the unobserved state effects the next unobserved state. However Cases 2 and 3 are trivial cases where there is no confounding. In Case 2, neither the policy nor the next state depend on $U_t$ so the problem is unconfounded. In Case 3, $U_t$ is completely causally disconnected from $S_t$ and $A_t$. **Second, assume there is no unblocked backdoor path from $S_{t-1}$ or $A_{t-1}$ to $U_t$.** There can be no edge directly from $S_{t-1} \to U_t$ or $A_{t-1} \to U_t$. All other such paths run through $U_{t-1} \to U_t$, so we have similar cases to above.

*Proof of Proposition 2* 1. There is no edge from $U_{t-1} \to U_t$. There may be an edge from $S_t \to U_t$ but not from $U_t \to S_t$.

2. There is an edge from $U_{t-1} \to U_t$ but there can be no edge from $S_{t-1} \to U_{t-1}$ or $U_{t-1} \to A_{t-1}$ or $U_t \to U_{t+1}$.

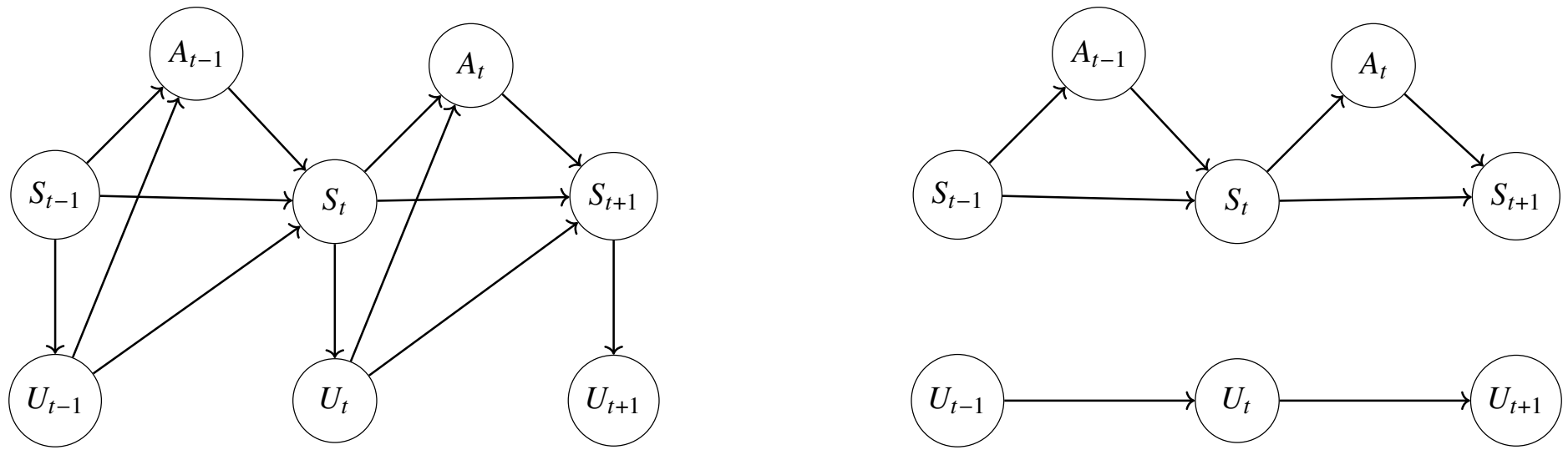

(a) Case 1 (b) Case 2

In Case 2, we cannot have an arrow from $U_t \to S_{t+1}$ because of the path: $S_{t-1} \leftarrow U_{t-2} \to U_{t-1} \to U_t \to S_{t+1}$, although we leave $U_{t-2}$ off the figure for space reasons.

Just as before, Case 1 satisfies the memoryless condition, but Case 2 has no causal relationship between $U$ and $S$ or $A$. Note that Case 1 in this figure is a generalization of Case 1 in the previous figure (it includes strictly more edges). To summarize, if Assumptions 1 and 2 hold in the time-homogeneous setting, either we have memoryless confounders, or we have a trivial case with no confounding at all.

### B.2. Time-Inhomogeneous Setting

In the time-inhomogeneous setting, the possible ways to break backdoor paths are more or less the same as described above. The key difference is that because the transitions can differ each period, the cases can be combined asymmetrically. I.e. if there is no unblocked backdoor path from $U_t$ to $S_{t+1}$ then there are no additional constraints at all on the edge from $t-1$ to $t$ — the path to $S_{t+1}$ is already blocked. Similarly if there is no unblocked path from $S_{t-1}$ or $A_{t-1}$ to $U_t$ then the edges from $U_t$ to $t+1$ are unconstrained. There are four essential cases illustrated in Figure 9 that can be combined in various ways. These follow essentially identical arguments to those in the previous section.

Importantly, unlike in the Time-Homogeneous setting, these can be combined in many different ways. We provide just a few illustrative examples with time horizon 6. Notice that usually each $U_t$ is either not a confounder or it is memoryless, but that these can be combined in various quite heterogeneous structures. Note that $U_2$ in Figure 11, and $U_2, U_4$ in Figure 12 are both confounders and not memoryless. However, the $U$ that they depend on in the past are otherwise disconnected from the rest of the DAG and so these are still effectively if not literally memoryless. So we might say that under Assumption 1 and Assumption 2 each time period in the time-inhomogeneous setting is either unconfounded or effectively memoryless. Of course, this will substantially complicate the proofs of our main results compared to just assuming memoryless confounders directly. Finally note that in these examples, sometimes the causal direction flows from $U_t \rightarrow S_t$ (so the unobserved state causes the observed state), whereas at other timesteps within the same system, $S_t \rightarrow U_t$ (so the observed state causes the unobserved state). Such behavior is very difficult to interpret, but cannot be ruled out by our general setting using Assumptions 1 and 2. However, this extra generality is worthwhile because Assumption 1 can be tested from observables.

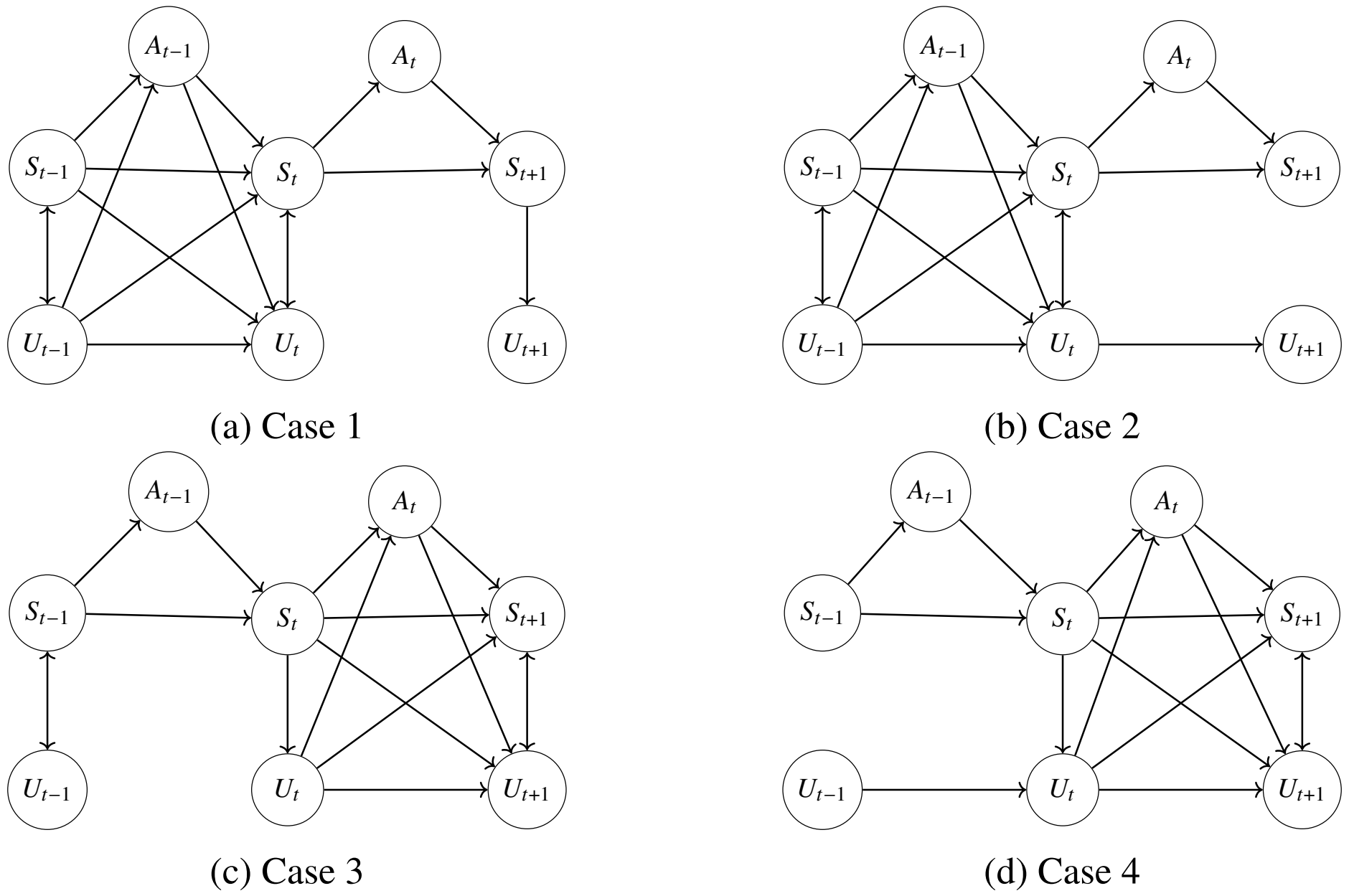

**Figure 9 Possible Cases in the Time-Inhomogeneous Setting**

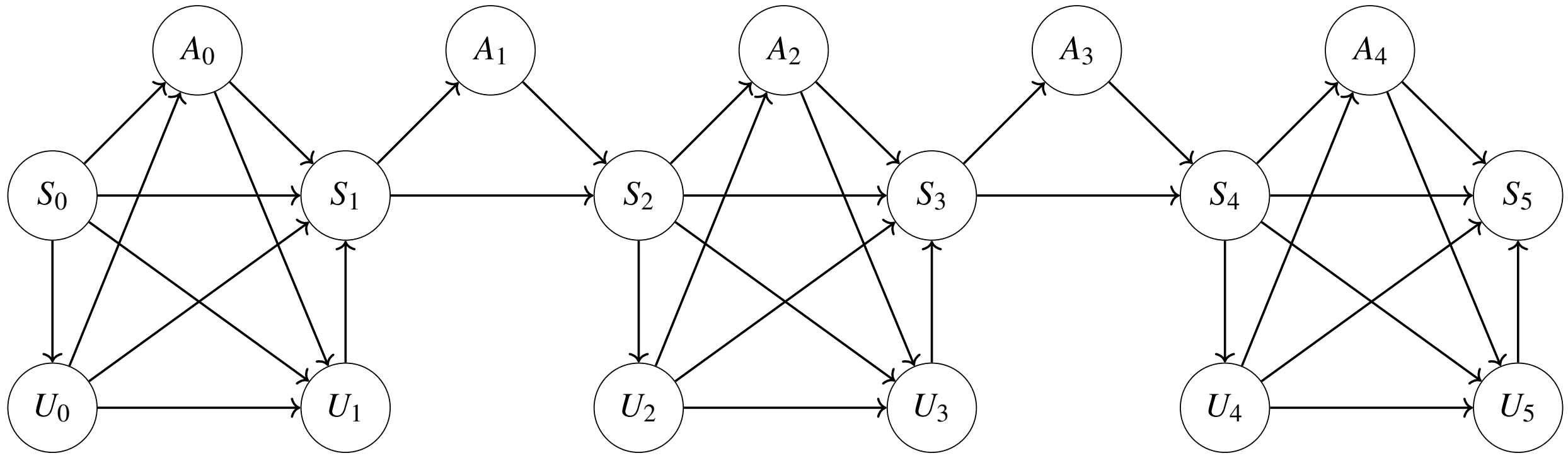

**Figure 10** **Example 1**

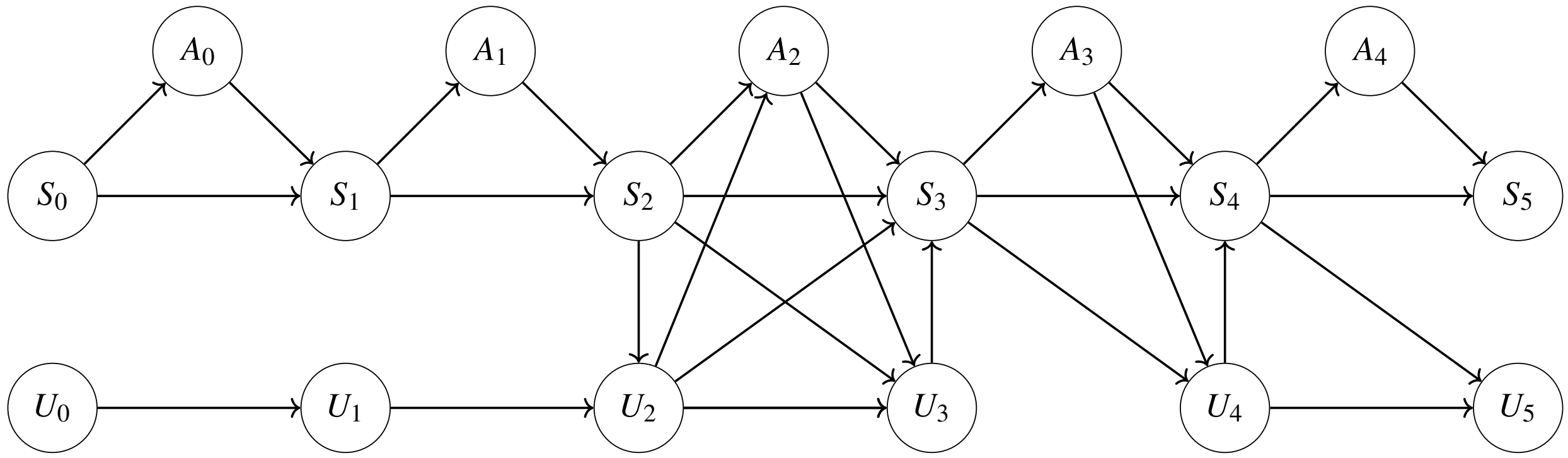

**Figure 11** **Example 2**

### B.3. Observed-State Markov with Lags

## Appendix 3: Proofs for Section 2

### 3.1. Proof of Proposition 6

**Part 1:** Let the observed states $s$, the actions $a$, and the unobserved states $u$ all be binary. We will consider three time steps, $t \in \{0,1,2\}$. We consider a simple construction. The initial probabilities $\chi(s,u) = 0.25, \forall s,u$. We generate the transition probabilities randomly by drawing from Unif$(0,1)$ and then normalizing appropriately. With with measure 1 wrt the Uniform distribution, $P(S_{t+1}, U_{t+1} | S_t, U_t, A_t)$ is not independent of $S_t, U_t, A_t$. Therefore, if we use the uniform policy $\pi^u(a) = 0.5, \forall a$, then we will have $P_{\pi^u}(S_2 = s_2 | S_1 = s_1, A_1 = a_1) \neq P_{\pi^u}(S_2 = s_2 | S_1 = s_1, A_1 = a_1, S_0 = s_0, A_0 = a_0)$. Note that for any $\pi^b$:

$$P_{\pi^b}(S_2|S_1, A_1) = \int_{\mathcal{U}} P_{\pi^b}(S_2|S_1, A_1, U_1) P_{\pi^b}(U_1|S_1, A_1)$$
$$P_{\pi^b}(S_2|S_1, A_1, S_0, A_0) = \int_{\mathcal{U}} P_{\pi^b}(S_2|S_1, A_1, U_1) P_{\pi^b}(U_1|S_1, A_1, S_0, A_0)$$

and so if $P_{\pi^b}(U_1|S_1, A_1) = P_{\pi^b}(U_1|S_1, A_1, S_0, A_0)$, then Assumption 1 would hold. We can achieve this by choosing $\pi^b(A_t|S_t, U_t) = 1$ if $A_t = U_t$ and 0 otherwise. Conditioning on $A_1$ then pins down $U_1$, so conditioning on $S_0, A_0$ has no effect. This completes the proof of part 1. Under this $\pi^b$, Assumption 1 holds, but even with the uniform policy $\pi^u$ that uses neither $s$ nor $u$, we would not end up with Markovian observed-state transitions. **Part 2:** We provide an example in Figure 14. Note that in the given graph, all paths from $S_2$ to $S_0, A_0$ are blocked by $S_1$ and $A_1$. Therefore, by faithfulness, $P_{\pi^b}(S_2|S_1, A_1) = P_{\pi^b}(S_2|S_1, A_1, S_0, A_0)$. However, if we replace $\pi^b$ with another policy $\pi$ that has an edge from $U_0$ to $A_0$ then by faithfulness we must have $P_\pi(S_2|S_1, A_1) \neq P_\pi(S_2|S_1, A_1, S_0, A_0)$ because there would exist an unblocked backdoor path from $S_2$ to $A_0$, in particular $A_0 \leftarrow U_0 \rightarrow U_1 \rightarrow S_2$.

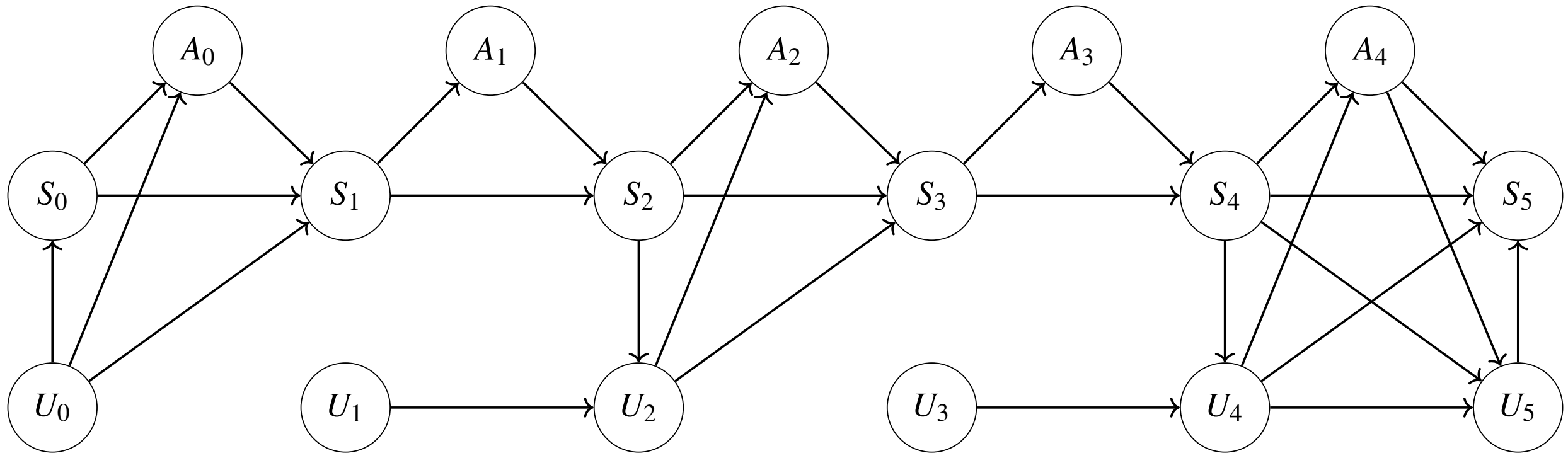

**Figure 12** **Example 3**

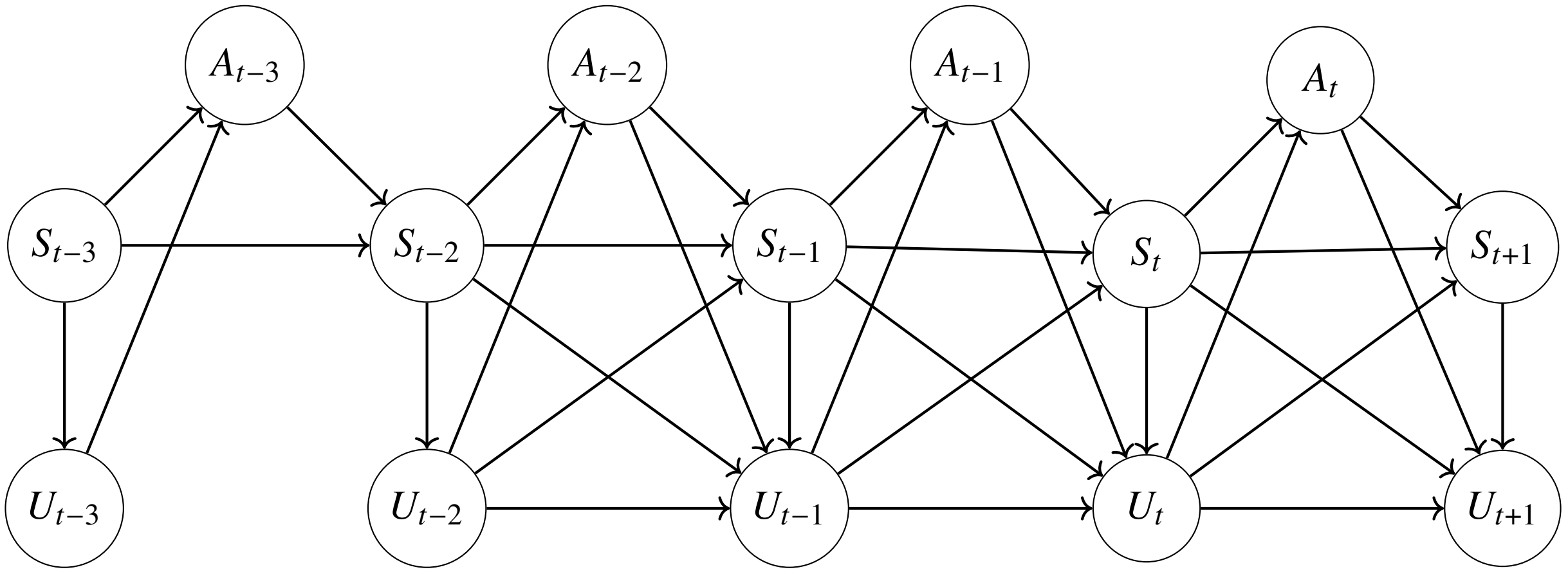

**Figure 13** **Observed-State Markov with 2 Lags**

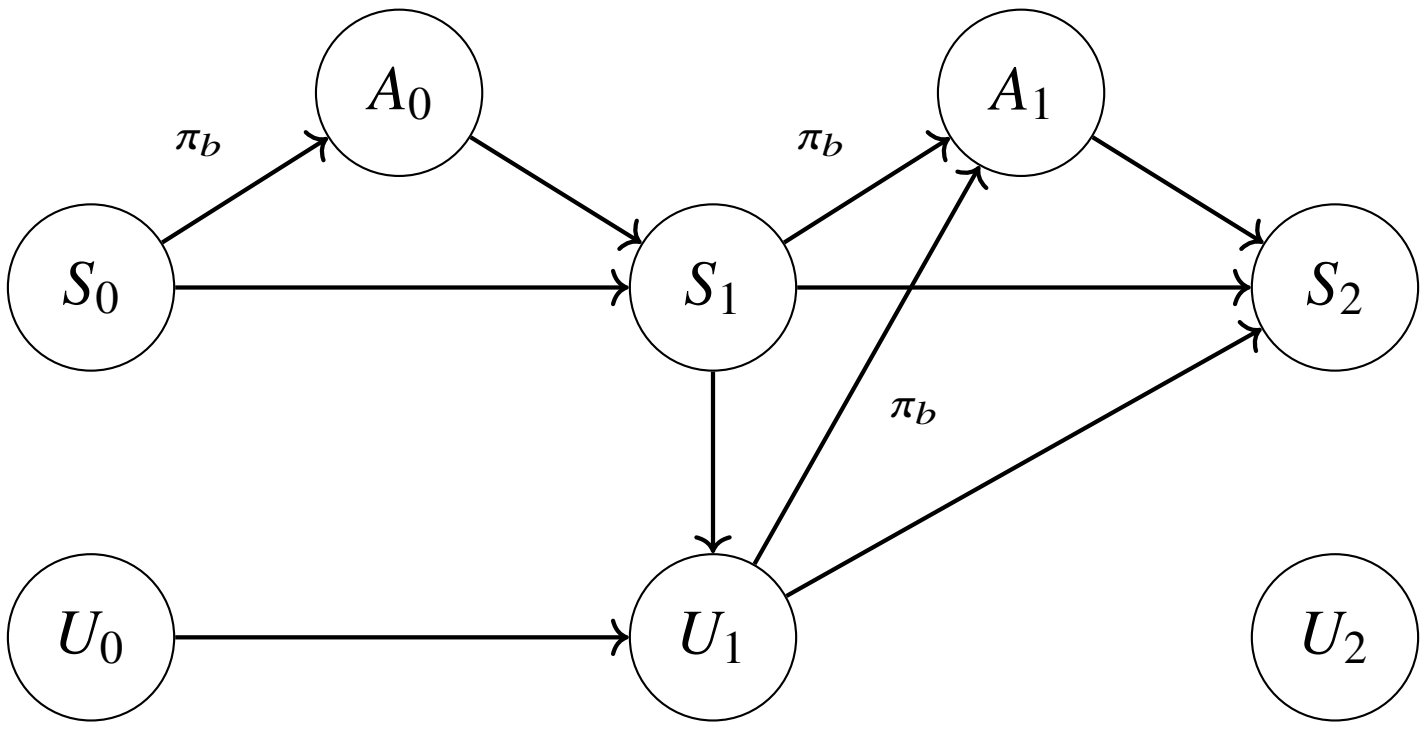

**Figure 14 Example for Part 2**

**Part 3:** Consider any time $t$. Under Assumption 1 and Assumption 2, while running $\pi^b$ given $S_t, A_t$, there are no unblocked paths from $S_{t+1}, ..., S_T$ to $S_{t-1}, ..., S_0, A_{t-1}, ..., A_0$. Similarly given $S_t$, there are no unblocked paths from $A_t, ..., A_T$ to $S_{t-1}, ..., S_0, A_{t-1}, ..., A_0$. The new policy $\pi^e$ could only induce a new unblocked path if it can add edges that didn't exist under $\pi^b$. This is only possible in the case where $\pi^b$ was missing an edge from some $S_{t'}$ to $A_{t'}$ for any $t'$. Finally, note that we do not have to consider new colliders, because $A$ cannot have more than one parent under $\pi^e$. We split the analysis into three cases. The new policy $\pi^e$ could add an edge could be added from $S$ to $A$ before time $t$, at time $t$, or after time $t$. Let's say the edge was added from $S_{t'}$ to $A_{t'}$ at some $t'$ before time $t$. There were no unblocked paths from $S_{t+1}$ to $S_{t'}$ or $A_{t'}$ given $S_t, A_t$ or from $A_t$ to $S_{t'}$ or $A_{t'}$ given $S_t$, so adding this edge cannot generate a new unblocked path. Let's say the edge was added from $S_t$ to $A_t$. $S_t$ is blocked so this can never generate a new unblocked path. Finally let's say the edge was added from $S_{t'}$ to $A_{t'}$ at some $t'$ after time $t$. There were no unblocked paths from $S_{t-1}, ..., S_0, A_{t-1}, ..., A_0$ to $S_{t'}$ or $A_{t'}$. Therefore, even if there is a path from $A_t$ or $S_{t+1}$ to $S_{t'}, A_{t'}$ adding this edge cannot generate a new path from $S_{t+1}$ to $S_{t-1}, ..., S_0, A_{t-1}, ..., A_0$. This completes the proof.

### 3.2. Confounding for Regression

We now prove the following proposition, from which Proposition 1 and Theorem 1 will follow.

PROPOSITION 9. *Let $f : \mathcal{S} \times \mathcal{A} \times \mathcal{S} \to \mathbb{R}$ be any function. Let $\pi$ be any policy such that:*

$$P_\pi(S_{t+1}|S_t = s, A_t = a, H_t = h) = P_\pi(S_{t+1}|S_t = s, A_t = a),$$
$$P_\pi(A_t|S_t = s, H_t = h) = P_\pi(A_t|S_t = s).$$

*For any $\pi^e$ that does not depend on $U_t$, given Assumption 2, for all $s, a, t$,*

$$\mathbb{E}_{\pi^e}\left[f(S_t, A_t, S_{t+1})|S_t = s, A_t = a\right] = \mathbb{E}_\pi\left[\frac{\pi_t(A_t|S_t)}{\pi_t(A_t|S_t, U_t)} f(S_t, A_t, S_{t+1})\middle| S_t = s, A_t = a\right].$$

Proof of Proposition 9: Fix any time $t$. First notice that:

$$\mathbb{E}_{\pi^e}\left[f(S_t, A_t, S_{t+1})|S_t, A_t\right] = \int f(S_t, A_t, S_{t+1}) P_{\pi^e}(S_{t+1}|S_t, A_t) dS_{t+1}$$

and similarly

$$\begin{aligned}\mathbb{E}_\pi\left[\frac{\pi_t(A_t|S_t)}{\pi_t(A_t|S_t, U_t)} f(S_t, A_t, S_{t+1})\middle| S_t, A_t\right] &= \int \frac{\pi_t(A_t|S_t)}{\pi_t(A_t|S_t, U_t)} f(S_t, A_t, S_{t+1}) P_\pi(S_{t+1}|S_t, A_t) dS_{t+1} \\ &= \int f(S_t, A_t, S_{t+1}) \int \frac{\pi_t(A_t|S_t)}{\pi_t(A_t|S_t, U_t)} P_\pi(S_{t+1}, U_t|S_t, A_t) dU_t dS_{t+1}.\end{aligned}$$

So it suffices to show that:

$$P_{\pi^e}(S_{t+1}|S_t, A_t) = \int \frac{\pi_t(A_t|S_t)}{\pi_t(A_t|S_t, U_t)} P_\pi(S_{t+1}, U_t|S_t, A_t) dU_t.$$

By the structure of the MDP, any backdoor path from $S_{t+1}$ to $S_{t-1}, A_{t-1}$ that is not blocked by $S_t, A_t$ must pass through $U_t$. Therefore we can proceed in cases: **Case 1:** consider the case where there is no unblocked backdoor path from $U_t$ to $S_{t+1}$. Then by faithfulness, there can be no edge directly from $U_t$ to $S_{t+1}$, which immediately implies that for all $s, u, a, t$ and for any $\pi'$:

$$P_t(S_{t+1}|S_t = s, U_t = u, A_t = a) = P_{\pi'}(S_{t+1}|S_t = s, A_t = a).$$

In this case:

$$\begin{aligned}\int_{\mathcal{U}} \frac{\pi(A_t|S_t)}{\pi(A_t|S_t, U_t)} P_\pi(S_{t+1}, U_t|S_t, A_t) dU_t &= \int_{\mathcal{U}} \frac{\pi(A_t|S_t)}{\pi(A_t|S_t, U_t)} P_\pi(U_t|S_t, A_t) P_t(S_{t+1}|U_t, S_t, A_t) dU_t \\ &= P_{\pi_e}(S_{t+1}|S_t, A_t) \int_{\mathcal{U}} \frac{\pi(A_t|S_t)}{\pi(A_t|S_t, U_t)} P_\pi(U_t|S_t, A_t) dU_t \\ &= P_{\pi_e}(S_{t+1}|S_t, A_t) \mathbb{E}_\pi\left[\frac{\pi(A_t|S_t)}{\pi(A_t|S_t, U_t)}\middle| S_t, A_t\right],\end{aligned}$$

where the second line follows by applying the fact that $P_t(S_{t+1}|S_t, U_t, A_t) = P_{\pi'}(S_{t+1}|S_t, A_t)$ for all $\pi'$ to the case of $\pi_e$. The proposition follows by recognizing:

$$\mathbb{E}_\pi \left[ \frac{\pi(A_t|S_t)}{\pi(A_t|S_t, U_t)} \middle| S_t, A_t \right] = 1.$$

**Case 2:** If there is an unblocked backdoor path from $U_t$ to $S_{t+1}$, then there cannot be an unblocked backdoor path from $S_{t-1}$ to $U_t$ or from $A_{t-1}$ to $U_t$. For Case 2, we additionally assume that there is no edge $U_{t-1} \to U_t$. Note that:

$$\begin{aligned}
P_{\pi_e}(S_{t+1}|S_t, A_t) &= P_{\pi_e}(S_{t+1}|S_t, A_t, H_{t-1}) \\
&= \int_{\mathcal{U}} P_{\pi_e}(S_{t+1}|S_t, A_t, U_t, H_{t-1}) P_{\pi_e}(U_t|S_t, A_t, H_{t-1}) dU_t \\
&= \int_{\mathcal{U}} P_t(S_{t+1}|S_t, A_t, U_t) P_{\pi_e}(U_t|S_t, H_{t-1}) dU_t,
\end{aligned}$$

and similarly for $\pi$ (but with our desired ratio). Therefore, it suffices to show that:

$$P_{\pi_e}(U_t|S_t, H_{t-1}) = P_\pi(U_t|S_t, H_{t-1}).$$

Since there is no edge from $U_{t-1}$ to $U_t$, for any policy $\pi'$, we have:

$$P_{\pi'}(U_t|S_t, H_{t-1}) = P_t(U_t|S_t, U_{t-1}, S_{t-1}, A_{t-1}).$$

**Case 3:** In the final case, just as in Case 2 there is an unblocked backdoor path from $U_t$ to $S_{t+1}$, so there cannot be an unblocked backdoor path from $S_{t-1}$ to $U_t$ or from $A_{t-1}$ to $U_t$. However, unlike Case 2, in Case 3, there does exist an edge $U_{t-1} \to U_t$. In this final case, we use an inductive argument. As in Case 2, it suffices to show:

$$P_{\pi_e}(U_t|S_t, H_{t-1}) = P_\pi(U_t|S_t, H_{t-1}). \tag{11}$$

Note that:

$$\begin{aligned}
P_\pi(U_t|S_t, H_{t-1}) &= \int_{\mathcal{U}} P_\pi(U_t|S_t, U_{t-1}, H_{t-1}) P_\pi(U_{t-1}|S_t, H_{t-1}) dU_{t-1} \\
&= \int_{\mathcal{U}} P_\pi(U_t|S_t, U_{t-1}, H_{t-1}) \frac{P_\pi(S_t|U_{t-1}, H_{t-1})}{P_\pi(S_t|H_{t-1})} P_\pi(U_{t-1}|H_{t-1}) dU_{t-1} \\
&= \frac{\int_{\mathcal{U}} P_t(S_t, U_t|S_{t-1}, U_{t-1}, A_{t-1}) P_\pi(U_{t-1}|H_{t-1}) dU_{t-1}}{P_\pi(S_t|H_{t-1})} \\
&= \frac{\int_{\mathcal{U}} P_t(S_t, U_t|S_{t-1}, U_{t-1}, A_{t-1}) P_\pi(U_{t-1}|H_{t-1}) dU_{t-1}}{\int_{\mathcal{U}} P_t(S_t|S_{t-1}, U_{t-1}, A_{t-1}) P_\pi(U_{t-1}|H_{t-1})}
\end{aligned}$$

and similarly for $\pi_e$. Only one term in this last expression can vary with the policy: $P_\pi(U_{t-1}|H_{t-1})$.

As we're in the case where an edge exists from $U_t$ to $S_{t+1}$ and an edge exists from $U_{t-1}$ to $U_t$, then by faithfulness there cannot be an edge from $U_{t-1}$ to $A_{t-1}$. Therefore, we have:

$$\begin{aligned} P_\pi(U_{t-1}|H_{t-1}) &= P_\pi(U_{t-1}|S_{t-1}, A_{t-1}, H_{t-2}) \\ &= P_\pi(U_{t-1}|S_{t-1}, H_{t-2}), \end{aligned}$$

and similarly for $\pi_e$. So we only need to show $P_\pi(U_{t-1}|S_{t-1}, H_{t-2}) = P_{\pi_e}(U_{t-1}|S_{t-1}, H_{t-2})$ — exactly the same as Equation (11) but for the previous time step. We can repeat the same argument by induction noting that we have assumed edges between $U_{t-1}$ to $U_t$ and from $U_t$ to $S_{t+1}$, and therefore the restrictions on $U_{t-1}$ w.r.t. $U_{t-2}, A_{t-2}$, and $S_{t-2}$ are the same as for $U_t$.

The base case is simple:

$$P_{\pi_e}(U_0|S_0) = P_\pi(U_0|S_0) = \chi(U_0|S_0),$$

because the initial state distribution is the same for all policies.

### 3.3. Proof of Proposition 1

We use this auxiliary lemma:

LEMMA 2. *Under Assumptions 1 and 2, for any two policies $\pi^{e_1}$ and $\pi^{e_2}$ that don't depend on $U_t$, we have:*

$$P_{\pi^{e_1}}(S_{t+1}|S_t, A_t) = P_{\pi^{e_2}}(S_{t+1}|S_t, A_t).$$

Proof:

Consider $\pi^{e_1}$. By Proposition 6.3, since $\pi^{e_1}$ doesn't depend on $U_t$, we have that $P_{\pi^{e_1}}(S_{t+1}|S_t, A_t) = P_{\pi^{e_1}}(S_{t+1}|S_t, A_t, H_t)$. Thus we can apply Proposition 9 with $\pi = \pi^{e_1}$ and $\pi^e = \pi^{e_2}$ and $f(S_t, A_t, S_{t+1}) = \mathbb{I}\{S_{t+1} = s_{t+1}\}$ to get for all $s_{t+1}, s, a, t$

$$\begin{aligned} P_{\pi^{e_2}}(S_{t+1} = s_{t+1}|S_t = s, A_t = a) &= \mathbb{E}_{\pi^{e_2}}\left[\mathbf{1}\{S_{t+1} = s_{t+1}\}|S_t = s, A_t = a\right] \\ &= \mathbb{E}_{\pi^{e_1}}\left[\frac{\pi^{e_1}(A_t|S_t)}{\pi^{e_1}(A_t|S_t, U_t)}\mathbf{1}\{S_{t+1} = s_{t+1}\}\middle|S_t = s, A_t = a\right] \\ &= \mathbb{E}_{\pi^{e_1}}\left[\frac{\pi^{e_1}(A_t|S_t)}{\pi^{e_1}(A_t|S_t)}\mathbf{1}\{S_{t+1} = s_{t+1}\}\middle|S_t = s, A_t = a\right] \\ &= \mathbb{E}_{\pi^{e_1}}\left[\mathbf{1}\{S_{t+1} = s_{t+1}\}|S_t = s, A_t = a\right] \\ &= P_{\pi^{e_1}}(S_{t+1} = s_{t+1}|S_t = s, A_t = a). \end{aligned}$$

The proof of Proposition 1 then goes as follows: choose any $\pi^{e_1}$ that doesn't depend on $U_t$. Let $P_t^m(S_{t+1}|S_t, A_t) := P_{\pi^{e_1}}(S_{t+1}|S_t, A_t)$. Consider any $\pi^e$. We have:

$$\begin{aligned}
P_{\pi^e}(S_0, A_0, ..., S_T, A_T) &= P_{\pi^e}(S_0)P_{\pi^e}(A_0|S_0)P_{\pi^e}(S_1|S_0, A_0)P_{\pi^e}(A_1, S_2, A_2, ..., S_T, A_T|S_1, S_0, A_0) \\
&= P_{\pi^e}(S_0)P_{\pi^e}(A_0|S_0)\prod_{t=1}^{T} P_{\pi^e}(A_t|S_t)P_{\pi^e}(S_t|S_{t-1}, A_{t-1}) \\
&= P_{\pi^e}(S_0)P_{\pi^e}(A_0|S_0)\prod_{t=1}^{T} P_{\pi^e}(A_t|S_t)P_{\pi^{e_1}}(S_t|S_{t-1}, A_{t-1}) \\
&= \chi^m(S_0)\pi^e(A_0|S_0)\prod_{t=1}^{T} \pi^e(A_t|S_t)P_t^m(S_t|S_{t-1}, A_{t-1})
\end{aligned}$$

The second line follows because $P_{\pi^e}(A_t|S_t) = P_{\pi^e}(A_t|S_t, U_t)$ which is independent of the past, and because by Proposition 6.3, $P_{\pi^e}(S_{t+1}|S_t, A_t) = P_{\pi^e}(S_{t+1}|S_t, A_t, H_t)$. The third line follows by Lemma 2. The fourth line follows from the definitions of $\chi^m$ and $P_t^m$. The final expression is exactly the standard MDP probability factorization for the marginal MDP, which completes the proof.

### 3.4. Proof of Theorem 1

From Proposition 1, we have that for all $\pi^e$, $P_t^m(S_{t+1}|S_t, A_t) = P_{\pi^e}(S_{t+1}|S_t, A_t)$ — this is true by construction for the marginal MDP, and because the marginal MDP is equivalent to the underlying under $\pi^e$, it is also true for the underlying MDP. Then we apply Proposition 9 with $\pi = \pi^b$ and any $\pi^e$.

### 3.5. Confounded Rewards

In the main text, for simplicity we assume that at each time step, the reward $R_t = r_t(S_t, A_t, S_{t+1})$, a deterministic function of $S_t, A_t, S_{t+1}$. Even in this simplified setting, $R_t$ is confounded due to its dependence on $S_{t+1}$ (which is confounded). It turns out that for our results the simplified setting is essentially without loss of generality, as we establish in this section. In the general case, we define the rewards and state transitions jointly. The transitions at time $t$ are:

$$P_t(S_{t+1}, U_{t+1}, R_t|S_t, U_t, A_t).$$

This allows $R_t$ to depend on $S_t, U_t, A_t, S_{t+1}, U_{t+1}$ as well as have additional independent auxiliary randomness. In this setting, we need to expand Assumption 1 to apply to the rewards as well.

Assumption 13 **(Observed-State Markov Property with Rewards).** *Let $H_t := (S_{t-1}, A_{t-1}, R_{t-1}, ..., S_0, A_0, R_0)$ be the history of observed variables before time $t$. Then for all $s, a, h, t$:*

$$P_{obs}(S_{t+1}, R_t | S_t = s, A_t = a, H_t = h) = P_{obs}(S_{t+1}, R_t | S_t = s, A_t = a),$$
$$P_{obs}(A_t | S_t = s, H_t = h) = P_{obs}(A_t | S_t = s).$$

This condition is still testable from observables.

Under Assumptions 13 and 2, all of the results in the paper follow from virtually identical arguments. For example, here is a version of Proposition 9 with rewards:

Proposition 10. *Let $f$ be any function that is measurable with repsect to $S_t, A_t, S_{t+1}, R_t$. Let $\pi$ be any policy such that:*

$$P_\pi(S_{t+1}, R_t | S_t = s, A_t = a, H_t = h) = P_\pi(S_{t+1}, R_t | S_t = s, A_t = a),$$
$$P_\pi(A_t | S_t = s, H_t = h) = P_\pi(A_t | S_t = s).$$

*For any $\pi^e$ that does not depend on $U_t$, given Assumption 2, for all $s, a, t$,*

$$\mathbb{E}_{\pi^e}\left[f(S_t, A_t, S_{t+1}, R_t) | S_t = s, A_t = a\right] = \mathbb{E}_\pi\left[\frac{\pi_t(A_t|S_t)}{\pi_t(A_t|S_t, U_t)} f(S_t, A_t, S_{t+1}, R_t) \middle| S_t = s, A_t = a\right].$$

The proof is virtually unchanged from Proposition 9. As in the proof of Proposition 9 it suffices to show that:

$$P_{\pi^e}(S_{t+1}, R_t | S_t, A_t) = \int \frac{\pi_t(A_t|S_t)}{\pi_t(A_t|S_t, U_t)} P_\pi(S_{t+1}, R_t, U_t | S_t, A_t) dU_t.$$

All backdoor paths from $S_{t+1}, R_t$ to $S_{t-1}, A_{t-1}, R_{t-1}$ unblocked by $S_t$ and $A_t$ must pass through $U_t$. Thus as before we have the same three cases: **Case 1:** $S_{t+1}$ and $R_t$ are both independent of $U_t$ given $S_t, A_t$. The proof is identical to Case 1 of Proposition 9. **Case 2:** There is no edge from $U_{t-1} \to U_t$. The proof is identical to Case 2 of Proposition 9.

**Case 3:** There is an edge from $U_{t-1} \to U_t$. The proof is identical to Case 3 of Proposition 9.

### 3.6. Testing for Markovian trajectories

*Methodology* We propose a simple regression-based test of conditional independence, using sample-splitting to estimate the impact of additional prior history information on improving estimation of $S_t$. We build on this test to sequentially test for the largest lag $k$ that is still predictive of $S_{t+1}$, assuming that the order of the process is homogeneous over time.

We test the conditional independence statement of whether the next state is independent of the $k$th lagged state, given the current state and action $(S_t, A_t)$ and $C_{t,k}$, the auxiliary history strictly in between $S_t$ and $S_{t-k}$:

$$S_{t+1} \perp S_{t-k} \mid S_t, A_t, C_{t,k}, k \geq 1 \text{ where } C_{t,k} = \begin{cases} \emptyset & \text{if } k = 1 \\ \{S_{t-k'}\}_{k'=1}^{k'=k-1} & \text{if } k > 1 \end{cases}$$

. For the case of lag $k = 1$, the statement is $S_{t+1} \perp S_{t-1} \mid \{S_t, A_t\}$. The test compares the predictive performance of two regression models:

- **Model 1 (Null Hypothesis $H_0$):** This model predicts the next state $S_{t+1}$ using only the information in the conditioning set.

$$S_{t+1} = f_1(S_t, A_t, C_{t,k}) + \epsilon_1$$

- **Model 2 (Alternative Hypothesis $H_1$):** This model includes as additional predictors the lagged states from $S_{t-k}$ up to and including $S_t$.

$$S_{t+1} = f_2(S_t, A_t, C_{t,k}, S_{t-k}) + \epsilon_2$$

If Model 2 does not provide a statistically significant improvement in prediction accuracy over Model 1, we fail to reject the null hypothesis of conditional independence. We use Mean Squared Error (MSE) as the performance metric.

REMARK 2. Overall this approach is slightly weaker than current conditional independence tests like generalized covariance measure of Shah and Peters (2020), which would attempt to test for full distributional independence of the residuals, but whose multi-variate extension is more complicated.

**Algorithm 2** Testing for lag $k$

1: **Input:** Dataset $\mathcal{D}_{t,obs} = \{S_t^{(i)}, A_t^{(i)}, S_{t+1}^{(}i)\}$, timestep $t$, lag $k$, significance level $\alpha$, independence threshold $\theta$.
2: Train two models via K-fold CV to predict $S_{t+1}$:
3: Model 1 conditions on history up to lag $k-1$: $(S_t, A_t, \ldots, S_{t-k+1})$.
4: Model 2 additionally conditions on the state at lag $k$: $(S_t, A_t, \ldots, S_{t-k+1}, S_{t-k})$.
5: For each fold $j = 1, \ldots, K$, train both models on the training data and calculate the MSE difference on the validation data: $\Delta_j \leftarrow \text{MSE}_{1,j} - \text{MSE}_{2,j}$.
6: Perform a one-sided t-test on the mean of $K$ differences, $\sum_{j=1}^{K} \Delta_j$, against a null hypothesis of $H_0 : \mathbb{E}[\Delta] \leq 0$.
7: Return: Conclude the data is not Markovian for any order up to $K-1$.

**Algorithm 3** Multi-Lag Conditional Independence Test

1: **Input:** Dataset $\mathcal{D}_{obs}$, max lag $K$, significance level $\alpha$, independence threshold $\theta$.
2: **for** lag $k = 1, \ldots, K$ **do**
3: Use Algorithm 2 to test each valid timestep $t$'s data and return the ratio

$$\rho_t = \frac{\text{number of times failed to reject the null hypothesis of conditional independence}}{\text{number of valid timesteps}}$$

4: Check for $(k-1)$-order Markov property: If $\rho_k \geq \theta$, conclude the data is $(k-1)$-Markov and terminate.
5: **end for**
6: Return: Conclude the data is not Markovian for any order up to $K-1$.

*Algorithm*

## Appendix 4: Additions on method

### 4.1. Extension to continuous actions

Although the manuscript focuses on binary or categorical actions, the method can directly be extended to continuous action spaces, at the expense of sharpness results and interpretability of the robust set. Jesson et al. (2022) proposes a continuous-action sensitivity model which instead directly bounds the density ratio (rather than the odds ratio):

$$\frac{1}{\Lambda} \leq \frac{\pi_t^b(a \mid s)}{\pi_t^b(a \mid s, u)} \leq \Lambda \tag{12}$$

In the continuous setting, densities could be greater than 1, which would violate conditions on the odds ratio. One way to interpret this sensitivity parameter is via implications for the KL-divergence of nominal and complete propensity scores. We can readily apply this to our problem by changing the uncertainty set on $W$ to that implied by the above. Namely, solve the same linear program but enforce that $W_t = \frac{\pi_t^b(a|s)}{\pi_t^b(a|x,u)}$ satisfy the constraints of eq. (12) rather than Assumption 3:

$$(\bar{\mathcal{T}}_t^* Q)(s,a) = \min_{W_t} \left\{ \mathbb{E}_{\text{obs}}\left[W_t Y_t(Q) | S_t = s, A_t = a\right] : \mathbb{E}_{\text{obs}}\left[W_t | S_t = s, A_t = a\right] = 1, \ \Lambda^{-1} \leq W_t \leq \Lambda, \text{a.e.} \right\}.$$

That is, the characterization of Proposition 5 holds, replacing the $(\alpha_t, \beta_t)$ bounds arising from the MSM with $(\Lambda^{-1}, \Lambda)$. The pointwise solution of the $(s,a)$-conditional optimization problem is structurally the same, i.e. a conditional quantile characterization at a different level. The only difference algorithmically is in the conditional quantile estimation; in the continuous action setting, we would appeal to function approximation and minimize the (orthogonalized) pinball loss with the action as a covariate. In the infinite-data, nonparametric limit, this would be well-specified; in practice, there will be some additional approximation error. Given those conditional quantiles, the rest of the method, (orthogonalization, etc.) proceeds analogously as discussed previously.

### 4.2. Warmstarting

*Parametrization for the simulation.* For offline-online simulations, we consider a linear-gaussian MDP with an unobserved confounder $U_t$ using the following parameterization:

$$\mathcal{S} \subset \mathbb{R}^8, \mathcal{A} = \{0,1,2,3\}, S_0 \sim \mathcal{N}(0, 0.1), \qquad \mathcal{U} = \{0,1,2,3\}, P_t(U_t|S_t) = 1/4$$
$$\pi^b(A_t|S_t, U_t) = 1/2 \text{ if } A_t = 3 - U_t, \ 1/6 \text{ otherwise} \implies \pi^b(A_t|S_t) = 1/4,$$
$$P_t(S_{t+1}|S_t, A_t, U_t) = \mathcal{N}(\theta_{\mu,s} S_t + \theta_{\mu,a} A_t + \theta_{\mu,u} U_t, \max\{\theta_{\sigma,s} S_t + \theta_{\sigma,a} A_t + 0.2, 0\}),$$
$$R_t = \mathcal{N}(\theta_{R,s}^T S_{t+1}, 10^{-8} + \mathbb{I}[U_t = 3]\,\mathbb{I}[A_t = 0]\,\sigma_R)$$

where the parameters $\theta_{\mu,s}, \theta_{\sigma,s} \in \mathbb{R}^{d\times d}$ and $\theta_{\mu,a}, \theta_{\mu,s}, \theta_{\sigma,a}, \theta_{R,s} \in \mathbb{R}^d$ are dense. Note that we've added some additional variability to the reward through the parameter $\sigma_R \in \mathbb{R}$; this is incorporated into our CVaR-based bounds without alteration because the variability is captured by the conditional quantile function. Finally, note that the smallest valid value for the MSM parameter is $\Lambda = 3$, as can be computed directly from $\pi^b(A_t|S_t, U_t)$ and $\pi^b(A_t|S_t)$.

*Amortizing calculations of robust bounds* We can compute the robust optimal Q parameters once at the start using robust FQI, before the online procedure begins.

By the definition of the robust optimal policy, the robust optimal Q function is always larger than the robust Q function of the online policy — thus using the robust optimal Q function is still a valid upper bound for the purposes of optimism. Formally, by saddlepoint properties, the policies evaluated by LSVI-UCB, $\hat{\pi}_k$, are feasible but suboptimal for the optimization problem that the robust $Q$ function solves: since $(\bar{\pi}^*, \bar{P}_t^*) \in \arg\max_\pi \inf_{\bar{P}_t \in \mathcal{P}_t} \mathbb{E}_{\bar{P}_t}\left[R_t + g\left(S_{t+1}, \pi_{t+1}^e\right) \mid s, a\right]$, we have that $\hat{\overline{Q}}_t \geq \hat{\overline{Q}}_t^{\hat{\pi}_k}$ (i.e. evaluating the latter at $\bar{P}_t^*$). This lets us perform offline robust FQI only once (instead of $K$ times), which saves substantial computational cost at the expense of slightly looser upper bounds.

**Algorithm 4** Warm-Started LSVI-UCB

Estimate the marginal behavior policy, $\pi_t^b(a|s)$, in the offline data.
**for** episode $k = 1, \ldots, K$ **do**
  Initialize $\theta_T, \hat{\overline{Q}}_T = 0$
  **for** timestep $t = T-1, \ldots, 0$ **do**
    Estimate $\hat{\overline{Q}}_t$, robust $Q$ function from observational dataset $\mathcal{D}_{obs}$, via robust policy eval for $\pi_t(\cdot) := \text{argmax}_a Q_{t+1}(\cdot, a)$, using the offline data as in Steps 4-6 of Algorithm 1
    $\Sigma_t \leftarrow \sum_{k'=1}^{k-1} \boldsymbol{\phi}\left(s_t^{k'}, a_t^{k'}\right) \boldsymbol{\phi}\left(s_t^{k'}, a_t^{k'}\right)^\top + \lambda \cdot \mathbf{I}$
    $\theta_t \leftarrow \Sigma_t^{-1} \sum_{k'=1}^{k-1} \boldsymbol{\phi}\left(s_t^{k'}, a_t^{k'}\right) \left[r_t^{k'} + \max_a Q_{t+1}\left(s_{t+1}^{k'}, a\right)\right]$
    $Q_t(\cdot,\cdot) \leftarrow \min \Big\{\theta_t^\top \boldsymbol{\phi}(\cdot,\cdot) + \xi \left[\boldsymbol{\phi}(\cdot,\cdot)^\top \Sigma_t^{-1} \boldsymbol{\phi}(\cdot,\cdot)\right]^{1/2},$
      $\max\{\theta_t^\top \boldsymbol{\phi}(\cdot,\cdot), \hat{\overline{Q}}_t(\cdot,\cdot)\},$
      $T\Big\}$
  **end for**
  **for** step $t = 0, \ldots, T-1$ **do**
    Take action $a_t^k \leftarrow \pi_t^k(s_t^k) := \text{argmax}_{a \in \mathcal{A}} Q_h\left(s_t^k, a\right)$, and observe $r_t^k$ and $s_{t+1}^k$
  **end for**
**end for**

## Appendix 5: Analysis and Guarantees

We first describe the estimation benefits we receive from orthogonalization before discussing analysis of robust fitted-Q-evaluation and iteration, and insights. (All proofs are in the appendix).

### 5.1. Algorithm variants - with cross-fitting

**Algorithm 5** Confounding-Robust Fitted-Q-Iteration

1: Initialize $\hat{\overline{Q}}_T = 0$. Obtain index sets of cross-fitted folds, $\{\mathcal{I}_{k(i,t)}\}_{i\in[K],t\in[T]}$

2: **for** $t = T-1, \ldots, 1$ **do**

3: Using data $\{\left(S_t^i, A_t^i, R_t^i, S_{t+1}^i\right) : k(i,t) = k'\}$:

Estimate the marginal behavior policy $\pi_t^b(a|s)$ and evaluate bounds $\alpha_t(s_t, a_t), \beta_t(s_t, a_t)$ as in Equation (2).

Compute nominal outcomes $\{Y_t^{(i)}(\hat{\overline{Q}}_{t+1}^{-k'})\}_{i=1}^n$ as in eq. (5).

For all $a \in \mathcal{A}$, fit $\hat{Z}_t^{1-\tau,k'}(s,a)$ the $(1-\tau)$th conditional quantile of the outcomes $Y_t^{(i)}$.

4: Using data $\{\left(S_t^i, A_t^i, R_t^i, S_{t+1}^i\right) : k(i,t) = -k'\}$:

Compute pseudo-outcomes $\{\tilde{Y}_t^{(i)}(\hat{Z}_t^{1-\tau,k'}, \hat{\overline{Q}}_{t+1}^{-k'})\}_{i=1}^n$ as in eq. (7).

Fit $\hat{\overline{Q}}_t^{-k'}$ via least-squares regression of $\tilde{Y}_t^{(i)}$ against $(s_t^{(i)}, a_t^{(i)})$.

5: Obtain the robust Q-function by averaging across folds: $\hat{\overline{Q}}_t = \sum_{k'=1}^K \hat{\overline{Q}}_t^{(k)}$

6: Compute $\pi_t^*(s) \in \arg\max_a \hat{\overline{Q}}_t(s,a)$.

7: **end for**

In the main text, we described sample splitting but omitted it from the algorithmic description for a simpler presentation. In Algorithm 5 we discuss the cross-fitting in detail. We use cross-time fitting and introduce folds that partition trajectories and timesteps, where $k(i,t) \in [K]$ designates the fold. For $K = 2$ we consider timesteps interleaved by parity (e.g. odd/even timesteps in the same fold). We let $-k(i,t)$ denote that nuisance $\hat{\mu}^{-k(i,t)}$ is learned from $\{S_{t'}^{(i)}, A_{t'}^{(i)}, S_{t'+1}^{(i)}\}_{i\in\mathcal{I}_{-k(i)}}$, where $t'$ and $t$ have the same parity, e.g. from the $-k(i)$ trajectories and from timesteps of the same evenness or oddness but is only used for evaluation in the other fold.

### 5.2. Infinite-horizon results

Results for the infinite-horizon setting follow readily from our analysis of the finite-horizon setting and characterization of the uncertainty set. For completeness we state results here, succinctly. First, the algorithm is analogous except with $K$ iterations (restated in Algorithm 6).

**Algorithm 6** Confounding-Robust Fitted-Q-Iteration (Infinite Horizon)

Estimate the marginal behavior policy $\pi^b(a|s)$.
Compute $\{\alpha_k(s^{(i)}, a^{(i)})\}_{i=1}^n$ as in Equation (2).
Initialize $\hat{\overline{Q}}_k = 0$.
**for** $k = 1, \ldots, K$ **do**
    Compute the nominal outcomes $\{Y_k^{(i)}(\hat{\overline{Q}}_{k-1})\}_{i=1}^n$ as in Equation (5).
    Fit $\hat{Z}_k^{1-\tau}(s,a)$ the $(1-\tau)$th conditional quantile of the outcomes $Y_k^{(i)}$.
    Compute pseudooutcomes $\{\tilde{Y}_k^{(i)}(\hat{Z}_k^{1-\tau}, \hat{\overline{Q}}_{k-1})\}_{i=1}^n$ as in Equation (7).
    Fit $\hat{\overline{Q}}_k$ via least-squares regression of $\tilde{Y}_k^{(i)}$ against $(s^{(i)}, a^{(i)})$.
    Compute $\pi_k^*(s) \in \arg\max_a \hat{\overline{Q}}_k(s,a)$.
**end for**

In the infinite-horizon setting, we assume the data is generated from the distribution $\mu \in \Delta(\mathcal{S} \times \mathcal{A})$. We instead assume concentrability with respect to stationary distributions.

Assumption 14 **(Infinite-Horizon concentrability coefficient ).** *We assume that there exists $C < \infty$ s.t. for any admissible $\rho$, the stationary distribution induced under an evaluation policy,*

$$\forall (s,a) \in \mathcal{S} \times \mathcal{A}, \frac{\rho(s,a)}{\mu(s,a)} \leq C$$

We first list some helpful lemmas (i.e. infinite-horizon counterparts of the finite-horizon versions).

Our analysis as in Theorem 2 can also be applied to the infinite-horizon case via alternative lemmas standard in the infinite-horizon setting; below we use results from (Chen and Jiang 2019). We introduce a discount factor, $\gamma < 1$.

Theorem 3 **(Infinite-horizon FQI convergence).** *Suppose Assumptions 4, 5, 8 and 14 and let $\bar{V}_{\max} = \frac{1}{1-\gamma} B_R$ be the upper bound on $\bar{V}$. Then, with probability $> 1-\delta$, under Assumption 10, we have that*

$$\left\| \hat{\overline{Q}}_k - \overline{Q}^\star \right\|_{2,\nu} \leq \frac{1-\gamma^k}{1-\gamma} \sqrt{C\left(\epsilon_1 + \epsilon_{Q,\mathcal{Z}}\right)} + \gamma^k \bar{V}_{\max} + o_p(\gamma^k n^{-\frac{1}{2}}).$$

*where*

$$\epsilon_1 = \frac{56\bar{V}_{\max}^2 \log\{N(\epsilon, \mathcal{Q}, \|\cdot\|) N(\epsilon, \mathcal{Z}, \|\cdot\|)/\delta\}}{3n} + \sqrt{\frac{32\bar{V}_{\max}^2 \log\{N(\epsilon, \mathcal{Q}, \|\cdot\|) N(\epsilon, \mathcal{Z}, \|\cdot\|)/\delta\}}{n} \epsilon_{Q,\mathcal{Z}}}.$$

### 5.3. Confidence intervals for unobserved confounding from finite observational datasets

For simplicity, so far we have described warm-starting with bounds obtained from a large observational dataset without finite-sample uncertainty in estimating bounds. We provide an asymptotic confidence interval under linear function approximation that readily extends our warmstarting approach to a finite observational study.

Let $\theta_t, \overline{\theta}_t$ be the parameter for the nominal and robust Q-function, respectively. We consider state-feature vectors, denoted as $\phi_{t,a} = \phi(S_t, a)$, i.e. they take a product form over actions for simplicity. We first require regularity conditions on the feature covariances.

ASSUMPTION 15 **(Identification)**. *Let* $\Sigma := \mathbb{E}[\phi(s,a)^\top \phi(s,a)]$ *denote population covariance matrix of state-action features. Assume that there exist* $0 < K_{\min} < K_{\max} < \infty$ *that do not depend on* $d$ *s.t.* $K_{\min} \leqslant \min \operatorname{eig}(\Sigma) \leqslant \max \operatorname{eig}(\Sigma) \leqslant K_{\max}$ *for all* $d$.

ASSUMPTION 16 **(Error of second moments)**. *Let* $\epsilon = \tilde{Y}_t(Z_t, \hat{\overline{Q}}_{t+1}) - Q_t(s,a)$. *Assume lower and upper bounds on its second moments:* $0 < \underline{\sigma}^2 := \sup_{(s,a)\in(\mathcal{S}\times\mathcal{A})} \mathbb{E}\left[\epsilon^2 \mid s,a\right]$, *and* $\bar{\sigma}^2 := \sup_{(s,a)\in(\mathcal{S}\times\mathcal{A})} \mathbb{E}\left[\epsilon^2 \mid s,a\right] < 0$.

We show that orthogonality and cross-fitting yield asymptotic normality. Because of the backward recursive structure in estimation, our final asymptotic variance is that of estimation with generated regressors (i.e. the next-time-step $Q$ function), which we analyze via the asymptotic variance of the generalized method of moments (GMM) (Newey and McFadden 1994). Let $\zeta$ denote the parameter for the linear conditional quantile. We overload notation and let $\tilde{Y}_{t,a}(\zeta_t^\top, \overline{\theta}_{t+1})$ denote the $(a)$-conditional pseudo-outcome with linear conditional quantile $Z_t = \zeta_t^\top \phi_t$ and robust $Q$ function $\overline{Q}_t(s,a) = \overline{\theta}_t^\top \phi_t$, i.e. with $a'$ the maximizing action or drawn with respect to the policy distribution.

THEOREM 4 **(Asymptotic normality for linear FQE )**. *Under Assumptions 4 to 8 and 15, the asymptotic covariance is defined via* $\bar{\theta}$ *satisfying the following moment equations: let*

$$g_{t,a}(\zeta^*, \bar{\theta}) = \left[\left\{\tilde{Y}_{t,a}(\zeta_t^*, \overline{\theta}_{t+1}) - \overline{\theta}_{t,a}^\top \phi_{t,a}\right\} \phi_{t,a}^\top\right] \mathbb{I}\left[A_t = a\right] / p(a), \tag{13}$$

*then* $\overline{\theta}$ *satisfies the stacked moment equation* $\{0 = \mathbb{E}[g_{t,a}(\zeta^*, \bar{\theta})]\}_{a\in\mathcal{A}, t=0,\ldots,T-1}$.

$$\sqrt{n}(\hat{\bar{\theta}} - \bar{\theta}^*) \xrightarrow{d} -\left(G^\top G\right)^{-1} G^\top \tilde{I}, \text{ where } \tilde{I} \sim N(0, I)$$

*The matrix* $G = \partial g(\zeta^*, \bar{\theta}) / \partial\bar{\theta}$ *is an upper triangular matrix. The entries of G are as follows:*

$$\frac{\partial g_{t,a}(\zeta^*, \overline{\theta})}{\partial \overline{\theta}_{t,a}} = \mathbb{E}[\phi_{t,a}\phi_{t,a}^\top]$$

$$\frac{\partial g_{t,a}(\zeta^*, \overline{\theta})}{\partial \overline{\theta}_{t+1,a'}} = \mathbb{E}\left[\alpha_{t,a}(\phi_{t+1,a'}\phi_{t,a}^\top) + (1-\alpha_{t,a})(Z^\phi_{a',t,a}\phi_{t,a}^\top)\right]$$

$$\text{where } Z^\phi_{a_{t+1}}(S_t, a) = \mathbb{E}[\phi(S_{t+1}, a_{t+1}) \mid Y_{t+1} \le \zeta_{t,a}^\top \phi_{t,a}, S_t, A_t = a].$$

Based on the asymptotic variance characterization, we can add an appropriate confidence interval to $\hat{\bar{Q}}$ in Step 7 of Algorithm 4 to maintain a confidence upper bound on the Q function.

## Appendix 6: Additional discussion

### 6.1. Related Work

**Connections to pessimism in offline RL.** Pessimism is an important algorithmic design principle for offline RL in the *absence* of unobserved confounders (Xie et al. 2021a, Rashidinejad et al. 2021, Jin et al. 2021). Therefore, robust FQI with lower-confidence-bound-sized $\Lambda$ gracefully degrades to a pessimistic offline RL method if unobserved confounders were, contrary to our method's use case, not actually present in the data. Conversely, pessimistic offline RL with *state-wise* lower confidence bounds confers some robustness against unobserved confounders. But state-wise LCBs are viewed as overly conservative relative to a profiled lower bound on the average value (Xie et al. 2021a).

### 6.2. Related Work for Warmstarting

Zhang and Bareinboim (2019) warm-start a variant of UCRL (Auer et al. 2008) for *tabular* dynamic treatment regimes with bounds from confounded data. Wang et al. (2021) does consider offline data with confounding and a similar warm-starting procedure. However, they also assume point-identifiability via backdoor adjustment or frontdoor adjustment. We will demonstrate that when this assumption fails, their procedure can have worse regret than not using the offline data at all. Other recent works, without unobserved confounders, study finer-grained hybrid offline-online RL (Xie et al. 2021b, Song et al. 2022). (Tennenholtz et al. 2021) consider linear contextual bandits constrained by moment conditions from the offline data. Xu et al. (2023a) studies restricted exploration for outperforming a conservative policy. We focus instead on demonstrating 1) how robust bounds from offline data can augment expensive online data and 2) how assuming memoryless unobserved confounders admits a marginal Markov decision process online counterpart, enabling warm-starting, unlike modeling unobserved confounders with POMDPs. We leave a full characterization for future work.

### 6.3. Derivation of the Closed-Form for the Robust Bellman Operator

*Proof of Proposition 5* Dorn et al. (2021) show that the linear program has a closed-form solution corresponding to adversarial weights:

$$\tilde{Y}^{-}_{f,t}(s,a) = \mathbb{E}_{\pi^b}\left[W_t^* Y_t | S_t = s, A_t = a\right] \text{ where } W_t^* = \alpha_t \mathbb{I}\left[Y_t > Z_t^{1-\tau}\right] + \beta_t \mathbb{I}\left[Y_t \leq Z_t^{1-\tau}\right].$$

We can derive the form in Proposition 5 with a few additional transformations. Define:

$$\mu_t(s,a) := \mathbb{E}_{\pi^b}[Y_t|S_t=s, A_t=a],\;\; \mathrm{CVaR}_t^{1-\tau}(s,a) := \frac{1}{1-\tau}\mathbb{E}_{\pi^b}\left[Y_t\mathbb{I}\left[Y_t < Z_t^{1-\tau}\right]|S_t=s, A_t=a\right].$$

We use the following identity for any random variables $Y$ and $X$:

$$\mathbb{E}[Y|X] = \mathbb{E}[Y\mathbb{I}\left[Y > Z^{1-\tau}(Y|X)\right]|X] + \mathbb{E}[Y\mathbb{I}\left[Y \le Z^{1-\tau}(Y|X)\right]|X]$$

to deduce that

$$\tilde{Y}_{f,t}^{-}(s,a) = \alpha_t\mu_t(s,a) + (\beta_t-\alpha_t)(1-\tau)\mathrm{CVaR}_t^{1-\tau}(s,a),$$

which gives the desired convex combination by noticing that $(\beta_t-\alpha_t)(1-\tau) = (1-\alpha_t)$.

# Appendix 7: Proofs for Robust FQE/FQI

## 7.1. Proof of Proposition 3

### Proof of CVaR characterization.

*T* he result follows by applying Corollary 4 of Dorn and Guo (2022) to Theorem 1.

## 7.2. Realizability Counterexample

We'll consider a highly simplified empirical distribution with only a single state. We'll drop all dependences on $S$ and $t$ for simplicity. The possible outcomes $Y$ lie in a discrete set and each have equal probability. We have three actions, the first with 4 data points, the second with 8 data points, and the last with 12 data points:

$$N = 24$$
$$P(A=0) = 4/24, P(A=1) = 8/24, P(A=2) = 12/24$$

Let the outcomes for the four $A = 0$ datapoints be $\{Y_i = i : i \text{ from 1 to 4}\}$. Similarly $Y_j$ and $Y_k$ for $A = 1$ and $A = 2$ respectively. Then:

$$P(Y_i|A=0) = 1/4, P(Y_j|A=1) = 1/8, P(Y_k|A=2) = 1/12$$

Let $\Lambda = 3$, so that $1 - \tau = 1/4$. Denote the relevant lower bounds on the weights as $\alpha(A) = P(A) + \frac{1}{\Lambda}(1 - P(A))$ and $\beta(A) = P(A) + \Lambda(1 - P(A))$. Then from the Dorn and Guo result, we have unique weights that achieve the infimum over the MSM ambiguity set:

For $A = 0, w = \{\beta(0), \alpha(0), \alpha(0), \alpha(0)\}$,

For $A = 1, w = \{\beta(1), \beta(1), \alpha(1), \alpha(1), \alpha(1), \alpha(1), \alpha(1), \alpha(1)\}$,

For $A = 2, w = \{\beta(2), \beta(2), \beta(2), \alpha(2), \alpha(2), \alpha(2), \alpha(2), \alpha(2), \alpha(2), \alpha(2), \alpha(2), \alpha(2)\}$

Consider the first weight for $A = 0$, $w = \beta(0)$. We know that there exists some arbitrary $u$ such that $P(A=0)/P(A=0|U=u) = \beta(0)$. Bayes rule then implies that:

$$P(U=u) = P(U=u|A=0)\beta(0)$$

Then we have:

$$P(U=u) = P(U=u|A=0)\beta(0) = \sum_a p(A=a)p(U=u|A=a)$$
$$\implies P(U=u|A=0)\beta(0) - P(U=a|A=0)P(A=0) = \sum_{a\neq 0} P(A=a)P(U=u|A=a)$$

and since $\beta(0) > p(A=0)$, the probability of $u$ occurring in the other actions must be non-zero. We therefore know that $P(A=1|U=u) \in \{P(A=1)/\alpha(A=1), P(A=1)/\beta(a=1)\}$ and similarly for $P(A=2|U=u)$. But there does not exist any choice such that $\sum_a P(A=a|U=u)=1$ given our choices of $\Lambda$ and $P(A)$.

#### 7.2.1. Auxiliary lemmas for robust FQE/FQI

LEMMA 3 **(Higher-order quantile error terms)**. *Assume Assumption 5 (i.e. bounded conditional density by $M_P$), and that $Z_t^{1-\tau}$ is differentiable with respect to s and its gradient is Lipschitz continuous. Then, for $f_t = R_t + \hat{\overline{Q}}_{t+1}$, if $\hat{Z}_t^{1-\tau}$ is $O_p(w_n)$ sup-norm consistent, i.e.* $\sup_{s\in\mathcal{S}} |Z_t^{1-\tau} - \hat{Z}_t^{1-\tau}| = O_p(w_n)$*, uniformly over $s \in \mathcal{S}$,*

$$|\mathbb{E}[(f_t - Z_t^{1-\tau})(\mathbb{I}[f_t \le \hat{Z}_t^{1-\tau}] - \mathbb{I}[f_t \le Z_t^{1-\tau}]) \mid S=s, A=1]| = O_p(w_n^2), \tag{14}$$

*and*

$$\mathbb{E}[(Z_t^{1-\tau} - \hat{Z}_t^{1-\tau})\left(\mathbb{I}[f \le Z_t^{1-\tau}] - (1-\tau)\right) \mid A=1] \le M_P \mathbb{E}[(Z_t^{1-\tau} - \hat{Z}_t^{1-\tau})^2 \mid A=a]. \tag{15}$$

Lemma 3 is a technical lemma which summarizes the properties of the orthogonalized target which lead to quadratic bias in the first-stage estimation error of $\hat{Z}_t$. Equation (14) is a slight modification of (Olma 2021)/(Kato 2012, A.3); eq. (15) is a slight modification of Semenova (2023, Lemma 4.1).

LEMMA 4 **(Bernstein concentration for least-squares loss (under approximate realizability))**. *Suppose Assumption 9 and that:*

1. *Approximate realizability: $\mathcal{Q}$ approximately realizes $\overline{\mathcal{T}}\mathcal{Q}$ in the sense that $\forall f \in \mathcal{Q}, z \in \mathcal{Z}$, let $q_f^\star = \arg\min_{q\in\mathcal{Q}} \|q - \overline{\mathcal{T}} f\|_{2,\mu}$, then $\left\| q_f^\star - \overline{\mathcal{T}} f \right\|_{2,\mu}^2 \le \epsilon_{\mathcal{Q},\mathcal{Z}}$.*
2. *The dataset $\mathcal{D}$ is generated from $P_{obs}$ as follows: $(s,a) \sim \mu, r = R(s,a), s' \sim P(s' \mid s,a)$.*

*We have that $\forall f \in \mathcal{Q}$, with probability at least $1-\delta$,*

$$\mathbb{E}_\mu[\ell(\widehat{\mathcal{T}}_{\mathcal{Z}} f; f)] - \mathbb{E}_\mu[\ell(g_f^\star; f)] \le \frac{56 V_{\max}^2 \ln \frac{|\mathcal{Q}||\mathcal{Z}|}{\delta}}{3n} + \sqrt{\frac{32 V_{\max}^2 \ln \frac{|\mathcal{Q}||\mathcal{Z}|}{\delta}}{n} \epsilon_{\mathcal{Q},\mathcal{Z}}}$$

LEMMA 5 **(Stability of covering numbers)**. *We relate the covering numbers of the squared loss function class, denoted as $\mathcal{L}_{q(z'),z}(q_{t+1})$, to the covering numbers of the function classes $\mathcal{Q}, \mathcal{Z}$. Define the squared loss function class as:*

$$\mathcal{L}_{q(z'),z}(q_{t+1}) = \left\{ \ell(q(z'), q_{t+1}; z) - \ell(\overline{Q}_{t,Z_t}^\dagger, q_{t+1}; z) : q(z') \in \{\mathcal{Q} \otimes \mathcal{Z}\}, z \in \mathcal{Z} \right\}$$

*Then*

$$N_{[]}\left(2\epsilon L, \mathcal{L}_{q(z'),z}, \|\cdot\|\right) \le N(\epsilon, \mathcal{Q} \times \mathcal{Z}, \|\cdot\|).$$

LEMMA 6 (**Difference of indicator functions**). *Let* $\widehat{f}$ *and* $f$ *take any real values. Then* $\left|\mathbb{I}[\widehat{f} > 0] - \mathbb{I}[f > 0]\right| \leq \mathbb{I}[|f| \leq |\widehat{f} - f|]$

### 7.3. Proofs of theorems

*Proof of Theorem 2* The squared loss with respect to a given conditional quantile function $Z$ is:

$$\ell(q, q_{t+1}; Z)$$
$$= \left(\alpha(R + q_{t+1}) + (1-\alpha)\left(Z_t^{1-\tau} + \frac{1}{1-\tau}\left((R + q_{t+1} - Z_t^{1-\tau})_- - Z_t^{1-\tau} \cdot (\mathbb{I}\left[R + q_{t+1} \le Z_t^{1-\tau}\right] - (1-\tau))\right)\right) - q_t\right)^2$$

We let $\hat{Z}_{t,Q_{t+1}}$ and $Z_{t,Q_{t+1}}$ denote estimated and oracle conditional quantile functions, respectively, with respect to a target function that uses the $Q_{t+1}$ estimate. Where the next-timestep $Q_{t+1}$ function is fixed (as it is in the following analysis) we drop the $Q_{t+1}$ from the subscript.

Define

$$\hat{\overline{Q}}_{t,Z_t} \in \arg\min_q \mathbb{E}_n[\ell(q, \hat{\overline{Q}}_{t+1}; Z_t)]$$

and for $z \in \{\hat{Z}_t, Z_t\}$, define the following *oracle* Bellman error projections $\overline{Q}^\dagger_{t,z}$ of the iterates of the algorithm:

$$\overline{Q}^\dagger_{t,z} = \arg\min_{q_t \in Q_t} \|q_t - \overline{\mathcal{T}}^*_{t,z} \hat{\overline{Q}}_{t+1}\|_{\mu_t}.$$

*Relating the Bellman error to FQE loss.* The bias-variance decomposition implies if $U, V$ are conditionally uncorrelated given $W$, then

$$\mathbb{E}[(U-V)^2 \mid W] = \mathbb{E}[(U - \mathbb{E}[V \mid W])^2 \mid W] + Var[V \mid W].$$

Hence a similar relationship holds for the robust Bellman error as for the Bellman error:

$$\mathbb{E}[\ell(q, \overline{Q}_{t+1}; Z)^2] = \|q - \overline{\mathcal{T}}^* \overline{Q}_{t+1}\|_\mu + Var[W_t^{*,\pi}(Z)(R_t + \bar{V}_{\overline{Q}_{t+1}}(S_{t+1})) \mid S_t, A].$$

which is used to decompose the Bellman error as follows:

$$\|\hat{\overline{Q}}_{t,\hat{Z}_t} - \overline{\mathcal{T}}^*_{t,Z_t} \hat{\overline{Q}}_{t+1}\|^2_{\mu_t} = \mathbb{E}_\mu[\ell(\hat{\overline{Q}}_{t,\hat{Z}_t}, \hat{\overline{Q}}_{t+1}; Z_t)] - \mathbb{E}_\mu[\ell(\overline{Q}^\dagger_{t,Z_t}, \hat{\overline{Q}}_{t+1}; Z_t)] + \|\overline{Q}^\dagger_{t,Z_t} - \overline{\mathcal{T}}^*_t \hat{\overline{Q}}_{t+1}\|^2_{\mu_t}.$$

Then,

$$\begin{aligned} &\|\hat{\overline{Q}}_{t,\hat{Z}_t} - \overline{\mathcal{T}}^*_{t,Z_t} \hat{\overline{Q}}_{t+1}\|^2_{\mu_t} \\ &= \mathbb{E}_\mu[\ell(\hat{\overline{Q}}_{t,\hat{Z}_t}, \hat{\overline{Q}}_{t+1}; Z_t)] - \mathbb{E}_\mu[\ell(\hat{\overline{Q}}_{t,Z_t}, \hat{\overline{Q}}_{t+1}; Z_t)] && (16) \\ &+ \mathbb{E}_\mu[\ell(\hat{\overline{Q}}_{t,Z_t}, \hat{\overline{Q}}_{t+1}; Z_t)] - \mathbb{E}_\mu[\ell(\overline{Q}^\dagger_{t,Z_t}, \hat{\overline{Q}}_{t+1}; Z_t)] && (17) \\ &+ \|\overline{Q}^\dagger_{t,Z_t} - \overline{\mathcal{T}}^*_t \hat{\overline{Q}}_{t+1}\|^2_{\mu_t} && (18) \end{aligned}$$

We bound eq. (16) by orthogonality and eq. (17) by Bernstein inequality arguments.

We bound the first term. Let $f$ denote the Bellman residual. Let $x = f$, $(a-x) = Q - f$, $b = Q'$. Since, by expanding the square and Cauchy-Schwarz, we obtain the following elementary inequality:

$$\begin{aligned}(a-x)^2 - (b-x)^2 &= (a-b)^2 + 2(a-b)(b-x) \\ &\leq (a-b)^2 + \sqrt{\mathbb{E}[(a-b)^2]\mathbb{E}[(b-x)^2]}\end{aligned}$$

Applying the above, we have that

$$\mathbb{E}_\mu[\ell(\hat{\overline{Q}}_{t,Z_t}, \hat{\overline{Q}}_{t+1}; Z_t)] - \mathbb{E}_\mu[\ell(\overline{Q}^\dagger_{t,Z_t}, \hat{\overline{Q}}_{t+1}; Z_t)] \leq \underbrace{\|(\hat{\overline{Q}}_{t,Z_t} - \overline{Q}^\dagger_{t,Z_t})\|_2^2}_{o_p(n^{-1}) \text{ by } Proposition\ 7} + \|(\hat{\overline{Q}}_{t,Z_t} - \overline{Q}^\dagger_{t,Z_t})^2\| \underbrace{\|\hat{\overline{Q}}_{t,Z_t} - \tilde{Y}_t(\hat{\overline{Q}}_{t+1}; Z_t)\|}_{=O_p(n^{-1/2}) \text{ by realizability}}$$

Therefore

$$\mathbb{E}_\mu[\ell(\hat{\overline{Q}}_{t,Z_t}, \hat{\overline{Q}}_{t+1}; Z_t)] - \mathbb{E}_\mu[\ell(\overline{Q}^\dagger_{t,Z_t}, \hat{\overline{Q}}_{t+1}; Z_t)] = o_p(n^{-1}).$$

We bound eq. (17) by Lemma 4 directly.

Supposing Assumption 9, we obtain that

$$\left\|\hat{Q}_t - \overline{\mathcal{T}}^\star_t \hat{Q}_{t+1}\right\|^2_{\mu_t} \leq \epsilon_{Q,\mathcal{Z}} + \frac{56V^2_{\max} \ln \frac{|Q||\mathcal{Z}|}{\delta}}{3n} + \sqrt{\frac{32V^2_{\max} \ln \frac{|Q||\mathcal{Z}|}{\delta}}{n} \epsilon_{Q,\mathcal{Z}}} + o_p(n^{-1}).$$

Instead, supposing Assumption 10, instantiate the covering numbers choosing $\epsilon = O(n^{-1})$. Lemma 5 bounds the bracketing numbers of the (Lipschitz over a bounded domain) loss function class with the covering numbers of the primitive function classes $Q, \mathcal{Z}$. Supposing that Bellman completeness holds with respect to $Q, \mathcal{Z}$, approximate Bellman completeness holds over the $\epsilon$-net implied by the covering numbers with $\epsilon_{Q,\mathcal{Z}} = O(n^{-1})$ and we obtain that:

$$\begin{aligned}\left\|\hat{Q}_t - \overline{\mathcal{T}}^\star_t \hat{Q}_{t+1}\right\|^2_{\mu_t} &\leq \epsilon_{Q,\mathcal{Z}} + \frac{56V^2_{t,\max} \log\{N(\epsilon, Q, \|\cdot\|)N(\epsilon, \mathcal{Z}, \|\cdot\|)/\delta\}}{3n} \\ &\quad + \sqrt{\frac{32V^2_{t,\max} \log\{N(\epsilon, Q, \|\cdot\|)N(\epsilon, \mathcal{Z}, \|\cdot\|)/\delta\}}{n} \epsilon_{Q,\mathcal{Z}}} + o_p(n^{-1}). \\ &\leq \epsilon_{Q,\mathcal{Z}} + \frac{56V^2_{t,\max} \log\{N(\epsilon, Q, \|\cdot\|)N(\epsilon, \mathcal{Z}, \|\cdot\|)/\delta\}}{3n}\end{aligned}$$

*Proof of Theorem 3* Note that Lemma 13, (Chen and Jiang 2019) establishes the Bellman error as an upper bound to the policy suboptimality. It states: Let $f : \mathcal{S} \times \mathcal{A} \to \mathbb{R}$ and $\hat{\pi} = \pi_f$ be the policy of interest, we have

$$\bar{V}^\star - \bar{V}^{\hat{\pi}} \leq \sum_{t=1}^{\infty} \gamma^{t-1} \left( \|\bar{Q}^\star - f\|_{2,\eta_t^{\hat{\pi}} \times \pi^\star} + \|\bar{Q}^\star - f\|_{2,\eta_t^{\hat{\pi}} \times \hat{\pi}} \right).$$

Choosing $f = \hat{\overline{Q}}_k$ and $f' = \hat{\overline{Q}}_{k-1}$ in (Chen and Jiang 2019, Lemma 15) gives

$$\left\|\hat{\overline{Q}}_k - \bar{Q}^\star\right\|_{2,\nu} \leq \sqrt{C}\left\|\hat{\overline{Q}}_k - \mathcal{T}\hat{\overline{Q}}_{k-1}\right\|_{2,\mu} + \gamma\left\|\hat{\overline{Q}}_{k-1} - \bar{Q}^\star\right\|_{2,P(\nu)\times\pi_{\hat{\overline{Q}}_{k-1},\bar{Q}^\star}}. \tag{19}$$

Note that we can apply the same analysis with $P(\nu)\times\pi_{\hat{\overline{Q}}_{k-1},\bar{Q}^\star}$ replacing the $\nu$ distribution on the left hand side, and expand the inequality $k$ times. Then it remains to upper bound $\left\|\hat{\overline{Q}}_k - \mathcal{T}\hat{\overline{Q}}_{k-1}\right\|_{2,\mu}$, which we can do via the same analysis of eqs. (16) to (18). Following the analysis of the proof of Theorem 2, we then obtain, with probability $\geq 1-\delta$,

$$\left\|\hat{\overline{Q}}_k - \overline{\mathcal{T}}^\star_t\hat{\overline{Q}}_{k-1}\right\|^2_{\mu_t} \leq \epsilon_{\mathcal{Q},\mathcal{Z}} + \epsilon_1 + o_p(n^{-1}),$$

where

$$\epsilon_1 = \frac{56V^2_{t,\max}\log\{N(\epsilon,\mathcal{Q},\|\cdot\|)N(\epsilon,\mathcal{Z},\|\cdot\|)/\delta\}}{3n} + \sqrt{\frac{32V^2_{\max}\log\{N(\epsilon,\mathcal{Q},\|\cdot\|)N(\epsilon,\mathcal{Z},\|\cdot\|)/\delta\}}{n}\epsilon_{\mathcal{Q},\mathcal{Z}}}.$$

Since $\epsilon_1$ and $\epsilon_{\mathcal{Q},\mathcal{Z}}$ are independent of $k$, and the bound holds uniformly over $k$, we have that, plugging the above back into the recursive expansion of Equation (19):

$$\left\|\hat{\overline{Q}}_k - \overline{Q}^\star\right\|_{2,\nu} \leq \frac{1-\gamma^k}{1-\gamma}\sqrt{C\left(\epsilon_1 + \epsilon_{\mathcal{Q},\mathcal{Z}}\right)} + \gamma^k\bar{V}_{\max}.$$

## 7.4. Proofs of intermediate results

### 7.4.1. Orthogonality

*Proof of Proposition 7* We first focus on the case of a single action, $a = 1$. First recall that in the population, $\mathbb{E}[Z_t^{1-\tau} + \frac{1}{1-\tau}(f_t - Z_t^{1-\tau}) \mid s,a] = \frac{1}{1-\tau}\mathbb{E}[f_t\mathbb{I}\left[f_t \leq Z_t^{1-\tau}\right] \mid s,a]$. In the analysis below we study this truncated conditional expectation representation.

$$\|\hat{\overline{Q}}_t(S,1) - \overline{Q}_t(S,1)\| \lesssim \|\mathbb{E}[\tilde{Y}_t(\hat{Z}_t,\hat{\overline{Q}}_{t+1}) - \tilde{Y}_t(Z_t,\hat{\overline{Q}}_{t+1}) \mid S, A=1]\| + \|\hat{\overline{Q}}_t(S,1) - \overline{Q}_t(S,1)\|$$
$$\text{by Prop. 1 of Kennedy (2020) (regression stability)}$$

Prop. 1 of Kennedy (2020) provides bounds on how regression upon pseudooutcomes with estimated nuisance functions relates to the case with known nuisance functions.

It remains to relate $\|\mathbb{E}[\tilde{Y}_t(\hat{Z}_t,\hat{\overline{Q}}_{t+1}) - \tilde{Y}_t(Z_t,\hat{\overline{Q}}_{t+1}) \mid S, A=1]\|$ to the terms comprising the pointwise bias, which are bounded by Lemma 3. We define these terms as:

$$B_1^1(S) = \mathbb{E}\left[\frac{1-\tilde{\alpha}}{1-\tau}\left\{(f_t - Z_t^{1-\tau})\left(\mathbb{I}\left[f_t \leq \hat{Z}_t^{1-\tau})\right] - \mathbb{I}\left[f_t \leq Z_t^{1-\tau})\right]\right)\right\} \mid S, A=1\right]$$
$$B_2^1(S) = \mathbb{E}\left[\frac{1-\tilde{\alpha}}{1-\tau}\left\{(Z_t^{1-\tau} - \hat{Z}_t^{1-\tau})\left(\mathbb{I}\left[f \leq Z_t^{1-\tau}\right] - (1-\tau)\right)\right\} \mid S, A=1\right].$$

Lemma 3 bounds these terms as quadratic in the first-stage estimation error of $\hat{Z}_t$.

We have that

$$\mathbb{E}[\tilde{Y}_t(\hat{Z}_t, \hat{\overline{Q}}_{t+1}) - \tilde{Y}_t(Z_t, \hat{\overline{Q}}_{t+1}) \mid S, 1] = B_1^1(S) + B_2^1(S).$$

To see this, note:

$$\begin{aligned}
&\mathbb{E}[\tilde{Y}_t(\hat{Z}_t, \hat{\overline{Q}}_{t+1}) - \tilde{Y}_t(Z_t, \hat{\overline{Q}}_{t+1}) \mid S, 1] \\
&= \mathbb{E}\Bigg[\frac{1-\tilde{\alpha}}{1-\tau}\Big\{\Big(f_t \mathbb{I}\left[f_t \leq \hat{Z}_t^{1-\tau})\right] - f_t \mathbb{I}\left[f_t \leq Z_t^{1-\tau})\right]\Big) \\
&\qquad - \Big(\hat{Z}_t^{1-\tau} \cdot (\mathbb{I}\left[f \leq \hat{Z}_t^{1-\tau}\right] - (1-\tau)) - Z_t^{1-\tau} \cdot (\mathbb{I}\left[f \leq Z_t^{1-\tau}\right] - (1-\tau))\Big) \pm Z_t^{1-\tau} \cdot \mathbb{I}\left[f \leq \hat{Z}_t^{1-\tau}\right]\Big\} \mid S, A = 1\Bigg] \\
&= \mathbb{E}\Bigg[\frac{1-\tilde{\alpha}}{1-\tau}\Big\{(f_t - Z_t^{1-\tau})\mathbb{I}\left[f_t \leq \hat{Z}_t^{1-\tau})\right] - (f_t - Z_t^{1-\tau})\mathbb{I}\left[f_t \leq Z_t^{1-\tau})\right] \\
&\qquad + (Z_t^{1-\tau} - \hat{Z}_t^{1-\tau})\mathbb{I}\left[f \leq Z_t^{1-\tau}\right] - (Z_t^{1-\tau} - \hat{Z}_t^{1-\tau})(1-\tau)\Big\} \mid S, A = 1\Bigg] \\
&= \mathbb{E}\Bigg[\frac{1-\tilde{\alpha}}{1-\tau}\Big\{(f_t - Z_t^{1-\tau})\Big(\mathbb{I}\left[f_t \leq \hat{Z}_t^{1-\tau})\right] - \mathbb{I}\left[f_t \leq Z_t^{1-\tau})\right]\Big) \\
&\qquad + (Z_t^{1-\tau} - \hat{Z}_t^{1-\tau})\Big(\mathbb{I}\left[f \leq Z_t^{1-\tau}\right] - (1-\tau)\Big)\Big\} \mid S, A = 1\Bigg] \\
&= B_1^1(S) + B_2^1(S)
\end{aligned}$$

Finally, we relate the root mean-squared conditional bias,

$$\|\mathbb{E}[\tilde{Y}_t(\hat{Z}_t, \hat{\overline{Q}}_{t+1}) - \tilde{Y}_t(Z_t, \hat{\overline{Q}}_{t+1}) \mid S, A = 1]\|,$$

to the above quadratic error as follows. Using the inequalities $(a+b)^2 \leq 2(a^2+b^2)$ and $\sqrt{a+b} \leq \sqrt{a} + \sqrt{b}$ (for nonnegative $a, b$), we obtain that

$$\begin{aligned}
\|\mathbb{E}[\tilde{Y}_t(\hat{Z}_t, \hat{\overline{Q}}_{t+1}) - \tilde{Y}_t(Z_t, \hat{\overline{Q}}_{t+1}) \mid S, A = 1]\| &= \sqrt{\mathbb{E}[(B_1^1(S) + B_2^1(S))^2 \mid A = 1]} \\
&\leq \sqrt{\mathbb{E}[2\{(B_1^1(S))^2 + (B_2^1(S))^2\} \mid A = 1]} \\
&\leq \sqrt{2\mathbb{E}[(B_1^1(S))^2 \mid A = 1]} + \sqrt{2\mathbb{E}[(B_2^1(S))^2 \mid A = 1]}.
\end{aligned}$$

The result follows by the uniform bounds of Lemma 3.

*Proof of Lemma 3* Proof of eq. (14):

For $l > 0$, define

$$\mathcal{M}_n^a(l) = \left\{g : \mathcal{S} \to \mathbb{R} \text{ s.t. } \sup_{s \in \mathcal{S}} |g(s) - Z_t^{1-\tau}(s, a)| \leq l w_n\right\}$$

Define

$$U_n(g,s) := |\mathbb{E}[(f_t - Z_t^{1-\tau})(\mathbb{I}\left[f_t \le \hat{Z}_t^{1-\tau}\right] - \mathbb{I}\left[f_t \le Z_t^{1-\tau}\right]) \mid S=s, A=1]|$$

We will show that for every $l > 0, s \in \mathcal{S}$:

$$\sup_{g \in \mathcal{M}_n(l)} U_n(g,s) = O_p\left(w_n^2\right)$$

Breaking up the absolute value,

$$U_n(g,s) \le \mathbb{E}[(f_t - Z_t^{1-\tau})(\mathbb{I}\left[Z_t^{1-\tau} \le f_t \le g\right]) \mid S=s, A=1] + \mathbb{E}[(Z_t^{1-\tau} - f_t)(\mathbb{I}\left[g \le f_t \le Z_t^{1-\tau}\right]) \mid S=s, A=1]$$

We will bound the first term, bounding the second term is analogous. Define

$$U_{1,n}(g,s) := \mathbb{E}[(f_t - Z_t^{1-\tau})(\mathbb{I}\left[Z_t^{1-\tau} \le f_t \le g\right]) \mid S=s, A=1]$$

Observe that

$$\begin{aligned}\sup_{g \in \mathcal{M}_n(l)} U_{1,n}(g,s) &= \mathbb{E}[(f_t - Z_t^{1-\tau})(\mathbb{I}\left[Z_t^{1-\tau} \le f_t \le Z_t^{1-\tau} + l w_n\right]) \mid S=s, A=1] \\ &\le M_P l^2 w_n^2\end{aligned}$$

The result follows.

Proof of eq. (15):

The argument follows that of Semenova (2017). The difference of indicators is nonzero on the events:

$$\mathcal{E}^- := \{f_t - \hat{Z}_t^{1-\tau} < 0 < f_t - Z_t^{1-\tau}\}$$
$$\mathcal{E}^+ := \{f_t - Z_t^{1-\tau} < 0 < f_t - \hat{Z}_t^{1-\tau}\}$$

On these events, the estimation error upper bounds the exceedance

$$\{\mathcal{E}^- \cup \mathcal{E}^+\} \implies \{|Z - f| < |Z_t^{1-\tau} - \hat{Z}_t^{1-\tau}|\} \tag{20}$$

(since $\mathcal{E}^- \implies \{f - \hat{Z}_t^{1-\tau} < 0 < f - Z_t^{1-\tau}\}$ and $\mathcal{E}^+ \implies \{0 < Z_t^{1-\tau} - f < Z_t^{1-\tau} - \hat{Z}_t^{1-\tau}\}$.)

Then

$$\begin{aligned}\mathbb{E}[(f_t - Z_t^{1-\tau})\mathbb{I}\left[\mathcal{E}^- \cup \mathcal{E}^+\right] \mid S=s, A=1] &= \int_{-|Z_t^{1-\tau}-\hat{Z}_t^{1-\tau}|}^{|Z_t^{1-\tau}-\hat{Z}_t^{1-\tau}|} (f_t(s,a,s') - Z_t^{1-\tau}) P(s' \mid s,a) ds' \\ &\le M_P \mathbb{E}[(Z_t^{1-\tau} - \hat{Z}_t^{1-\tau})^2 \mid S=s, A=1]\end{aligned}$$

Assumption 7 ensures the result holds for state distributions that could arise during policy fitting. The above results hold conditionally on some action $A = 1$ but hold for all actions.

### 7.4.2. Other lemmas

*Proof of Lemma 4* Recall that

$$\ell(q, q_{t+1}; Z)$$
$$= \left(\alpha(R + q_{t+1}) + (1-\alpha)\left(Z_t^{1-\tau} + \frac{1}{1-\tau}\left((R + q_{t+1} - Z_t^{1-\tau})_- - Z_t^{1-\tau} \cdot (\mathbb{I}\left[R + q_{t+1} \leq Z_t^{1-\tau}\right] - (1-\tau))\right)\right) - q_t\right)^2$$

Define $f_{q',z} =$ Define $X$ to be the difference of the integrands.

Step 1:

$$Var(X(g, f, z, g_f^*)) \leq 4V_{\max}^2 \|\hat{\overline{Q}}_{t,Z_t} - \overline{Q}_{t,Z_t}^{\dagger}\|_2^2$$

(by similar arguments as in the original paper). By the same arguments (i.e. adding and subtracting $\overline{\mathcal{T}} f$) we obtain that

$$\|\hat{\overline{Q}}_{t,Z_t} - \overline{Q}_{t,Z_t}^{\dagger}\|_2^2 \leq 2(\mathbb{E}[X(g, f, z, g_f^*)] + 2\epsilon_{Q,\mathcal{Z}})$$

Therefore,

$$Var(X(g, f, z, g_f^*)) \leq 8V_{\max}^2(\mathbb{E}[X(g, f, z, g_f^*)] + 2\epsilon_{Q,\mathcal{Z}}).$$

Applying (one-sided) Bernstein's inequality uniformly over $Q, \mathcal{Z}$, we obtain:

$$\mathbb{E}\left[X(g, f, z, g_f^*)\right] - \mathbb{E}_n[X(g, f, z, g_f^\star)]$$
$$\leq \sqrt{\frac{16V_{\max}^2\left(\mathbb{E}\left[X(g, f, z, g_f^\star)\right] + 2\epsilon_{\mathcal{F},\mathcal{Z}}\right)\ln\frac{|Q||\mathcal{Z}|}{\delta}}{n}} + \frac{4V_{\max}^2 \ln\frac{|Q||\mathcal{Z}|}{\delta}}{3n}$$

Note that $\hat{\overline{Q}}_{t,Z_t}$ minimizes both $\mathbb{E}_n[\ell(q, \hat{\hat{Q}}_{t+1}; Z_t)]$ and $\mathbb{E}[(q, \hat{\hat{Q}}_{t+1}, Z_t, \overline{Q}_{\hat{\hat{Q}}_{t+1}}^*)]$ with respect to $q$. Therefore, by completeness since the Bayes-optimal predictor is realizable,

$$\mathbb{E}[\ell(\hat{\overline{Q}}_{t,Z_t}, \hat{\hat{Q}}_{t+1}; Z_t)] \leq \mathbb{E}[\ell(\overline{Q}_{t,Z_t}^{\dagger}, \hat{\hat{Q}}_{t+1}; Z_t)] = 0$$

Therefore (solving for the quadratic formula),

$$\mathbb{E}[X(\hat{\overline{Q}}_{t,Z_t}, \hat{Q}_{t+1}, Z_t, \overline{Q}_{t,Z_t}^{\dagger})] \leq \frac{56V_{\max}^2 \ln\frac{|Q||\mathcal{Z}|}{\delta}}{3n} + \sqrt{\frac{32V_{\max}^2 \ln\frac{|Q||\mathcal{Z}|}{\delta}}{n}\epsilon_{\mathcal{F},\mathcal{Z}}}$$

*Proof of Lemma 5* We show this result by establishing Lipschitz-continuity of the squared loss function class (with respect to the product function class of $Q \times \mathcal{Z}$).

We use a stability result on the bracketing number under Lipschitz transformation. Classes of functions $x \mapsto f_\theta(x)$ that are Lipschitz in the index parameter $\theta \in \Theta$ have bracketing numbers readily related to the covering numbers of $\Theta$. Suppose that

$$|f_{\theta'}(x) - f_\theta(x)| \leq d(\theta', \theta)F(x),$$

for some metric $d$ on the index set, function $F$ on the sample space, and every $x$. Then $(\operatorname{diam}\Theta)F$ is an envelope function for the class $\{f_\theta - f_{\theta_0} : \theta \in \Theta\}$ for any fixed $\theta_0$. We invoke Theorem 2.7.11 of van de Vaart and Wellner (1996) which shows that the bracketing numbers of this class are bounded by the covering numbers of $\Theta$.

Theorem 5 **((van de Vaart and Wellner 1996), Theorem 2.7.11).** *Let $\mathcal{F} = \{f_\theta : \theta \in \Theta\}$ be a class of functions satisfying the preceding display for every $\theta'$, $\theta$ and some fixed function $F$. Then, for any norm $\|\cdot\|$,*

$$N_{[]}\left(2\epsilon\|F\|, \mathcal{F}, \|\cdot\|\right) \le N(\epsilon, \Theta, d).$$

This shows that the bracketing numbers of the loss function class can be expressed via the covering numbers of the estimated function classes $\mathcal{Q}, \mathcal{Z}$, which are the primitive function classes of estimation, for which results are given in various references for typical function classes.

Denote

$$\begin{aligned} g(q_{t+1}) &= \alpha(s,a)(R + q_{t+1}) \\ h(z) &= (1-\alpha)\left(\frac{1}{1-\tau}\left(z + (R + q_{t+1} - z)_{-} - z \cdot \left(\mathbb{I}\left[R + q_{t+1} \le z\right] - (1-\tau)\right)\right)\right) \end{aligned}$$

and notate

$$\ell(q, q_{t+1}; z) = (q - g(q_{t+1}) + h(q_{t+1}, z))^2.$$

Note that $\frac{1}{1-\tau} = (1+\Lambda)$. Assuming bounded rewards, define $D_{z,t}, D_{q,t}$ as the diameters of $\mathcal{Q}_t, \mathcal{Z}_t$, respectively and note that $D_{z,t} \approx D_{q,t}$. Note that $h(q_{t+1}, z)$ is $(1-\alpha_{\min})(3(1+\Lambda)+1)$-Lipschitz in $z$ (since the sum of Lipschitz continuous functions is Lipschitz) and it is $(1-\alpha_{\min})\left(1 + (1+\Lambda)(\frac{D_{z,t}}{D_{q,t}} + 1)\right)$-Lipschitz in $q_{t+1}$. Further, $g(q_{t+1})$ is $\alpha_{\max}$-Lipschitz in $q_{t+1}$. Therefore, $\ell\left(q, q_{t+1}; z\right)$ is $D_{q,t}$ Lipschitz in $q$, $L^C_{q,t+1}$-Lipschitz in $q_{t+1}$ and $L^C_{z,t}$-Lipschitz in $z$, with $L^C_{q,t+1}, L^C_{z,t}$ defined as follows:

$$\begin{aligned} L^C_{q,t+1} &= (2D_{q,t+1} + D_{z,t})(1-\alpha_{\min})\left(\left(1 + (1+\Lambda)(\frac{D_{z,t}}{D_{q,t+1}} + 1)\right) + \alpha_{\max}\right) \\ L^C_{z,t} &= (2D_{q,t+1} + D_{z,t})(1-\alpha_{\min})(3(1+\Lambda)+1). \end{aligned}$$

Therefore we have shown that restrictions of $\ell\left(q, q_{t+1}; z\right)$ to the $q_{t+1}, z$ coordinates are individually Lipschitz. We leverage the fact that a function $f : \mathbb{R}^n \to \mathbb{R}$ is Lipschitz if and only if there exists a constant $L$ such that the restriction of $f$ to every line parallel to a coordinate axis is Lipschitz with constant $L$. Choosing

$$L_t = \sqrt{3}\max\{D_q, L^C_{q,t+1}, L^C_{z,t}\}$$

gives that $\ell(q, q_{t+1}; z)$ is $L_t$-Lipschitz.

*Proof of Corollary 1* Lemma 5 gives that $\ell(q, q_{t+1}; z)$ is $L_t$-Lipschitz with $L_t = \sqrt{3}\max\{D_q, L^C_{q,t+1}, L^C_{z,t}\}$.

To interpret the scaling of the result, we can appeal to van de Vaart and Wellner (1996, Thm. 2.6.4) which upper bounds the (log) covering numbers by the VC-dimension. Namely, van de Vaart and Wellner (1996, Thm. 2.6.4) states that there exists a universal constant $K$ such that

$$N(\epsilon, \mathcal{F}, L_r(Q)) \leq KV(\mathcal{F})(4e)^{V(\mathcal{F})}\left(\frac{1}{\epsilon}\right)^{r(V(\mathcal{F})-1)}.$$

Therefore, achieving an $\epsilon = cn^{-1}$ approximation error on the bracketing numbers of robust $Q$ functions results in an $\log(2L_t n)$ dependence.

Lastly we remark on instantiating $L_t$. Note that under the assumption of bounded rewards, $D_{q,t+1} = B_r(T-t+1)$. Focusing on leading-order dependence in problem-dependent constants, we have that $L_t = O(B_r(T-t)\Lambda)$. Then $\hat{\mathcal{E}}(\hat{Q}) \leq \epsilon + \sum_{t=1}^T K\frac{\log(2B_r(T-t)\Lambda n)}{n}$. Upper bounding the left Riemann sum by the integral, we obtain that

$$\sum_{t=1}^T K\frac{\log(2KB_r(T-t)\Lambda n/\epsilon)}{n} \leq \int_1^T K\frac{\log(2KB_r(T-x)\Lambda n/\epsilon)}{n}dx = \frac{(T-1)}{n}(\log(2KB_r\Lambda(T-1)n/\epsilon)-1).$$

### 7.5. Confounding with infinite data

First, we prove the following useful result for confounded regression with conditional Gaussian tails:

LEMMA 7. *Define:*

$$C(\Lambda) := \left(\frac{\Lambda^2-1}{\Lambda}\right)\phi\left(\Phi^{-1}\left(\frac{1}{1+\Lambda}\right)\right),$$

*where $\phi$ and $\Phi$ are the standard Gaussian density and CDF respectively. Let $Y_t(Q)$ be conditionally Gaussian given $S_t = s$ and $A_t = a$ with mean $\mu_t(s,a)$ and standard deviation $\sigma_t(s,a)$. Then,*

$$(\bar{\mathcal{T}}_t^* Q)(s,a) = \mu_t(s,a) - [1-\pi_t^b(a|s)]C(\Lambda)\sigma_t(s,a).$$

*Proof of Lemma 7* The CVaR for Gaussians has a closed-form (Norton et al. 2021):

$$\frac{1}{1-\tau}\mathbb{E}_{\pi^b}\left[Y_t(Q)\mathbb{I}\left[Y_t(Q) < Z_t^{1-\tau}\right]|S_t = s, A_t = a\right] = \mu_t(s,a) - \sigma_t(s,a)\frac{\phi(\Phi^{-1}(1-\tau))}{1-\tau}.$$

Applying this to Proposition 5 gives the desired result.

*Proof of Proposition 8* First, note that $R_t$ is conditionally Gaussian given $S_t$ and $A_t$ with mean $\theta_R\theta_P s$ and standard deviation $\theta_R\sigma_T$. Define $\beta_i := \theta_R \sum_{k=1}^{i} \theta_P^k$. Using value iteration, we can show that $V_{T-i}^{\pi^e}(s) = \beta_i s$ for $i \geq 1$. E.g. by induction, $V_{T-1}^{\pi^e}(s) = \theta_R\theta_P s = \beta_1$ and if $V_{T-t+1}^{\pi^e}(s) = \beta_{t-1}s$, then

$$V_{T-t}^{\pi^e}(s) = \theta_P(\theta_R + \gamma\beta_{t-1})s = \beta_t s.$$

Next we will derive the form of the robust value function by induction. For the base case, $t = T-1$, we have:

$$Y_{T-1} = \theta_R s'.$$

Therefore, $Y_{T-1}$ is conditionally gaussian with mean $\theta_R\theta_P s$ and standard deviation $\theta_R\sigma_P$. Applying Lemma 7, we have:

$$\bar{V}_{T-1}^{\pi^e}(s) = \theta_R\theta_P s - 0.5C(\Lambda)\theta_R\sigma_P.$$

Now assume that $\bar{V}_{t+1}^{\pi^e}(s) = \theta_V s + \alpha_V$. Then

$$\begin{aligned} Y_t &= \theta_R s' + (\theta_V s' + \alpha_V) \\ &= (\theta_R + \theta_V)s' + \alpha_V. \end{aligned}$$

Therefore, $Y_t$ is conditionally gaussian with mean $(\theta_R + \theta_V)\theta_P s + \alpha_V$ and standard deviation $(\theta_R + \theta_V)\sigma_P$. Applying Lemma 7, we have:

$$\bar{V}_t^{\pi^e}(s) = (\theta_R + \theta_V)\theta_P s + \alpha_V - 0.5C(\Lambda)(\theta_R + \theta_V)\sigma_P, \tag{21}$$

which is linear in $s$ with new coefficients $\theta_V' := (\theta_R + \theta_V)\theta_T$ and $\alpha_V' := \alpha_V - 0.5C(\Lambda)(\theta_R + \theta_V)\sigma_P$.

By rolling out the recursion defined in Equation (21), consolidating the coefficients into $\beta_i$ terms, and then simplifying we get:

$$\bar{V}_0^{\pi^e}(s) = V_0^{\pi^e}(s) - \frac{1}{2\theta_P}\left(\sum_{i=0}^{T-1}\beta_i\right)\sigma_P C(\Lambda).$$

Finally, that $C(\Lambda) \leq \frac{1}{8}\log(\Lambda)$ can be verified numerically.

### 7.6. Proofs for warm-starting

*Proof of Theorem 4* We prove this via backwards induction.

We show asymptotic linearity, which follows from orthogonality. Define the following:

$$\theta^*_{t,a} = \mathbb{E}[\phi_{t,a}\phi_{t,a}^\top]^{-1}\mathbb{E}[\phi_{t,a}^\top \overline{Q}_t(S_t,a)] = \mathbb{E}[\phi_{t,a}\phi_{t,a}^\top]^{-1}\mathbb{E}[\phi_{t,a}^\top \tilde{Y}_{t,a}(\zeta_t^*, \overline{\theta}^*_{t+1,a})]$$

$$\tilde{\theta}_{t,a} = \mathbb{E}_n[\phi_{t,a}\phi_{t,a}^\top]^{-1}\mathbb{E}_n[\phi_{t,a}^\top \tilde{Y}_{t,a}(\zeta_t^*, \overline{\theta}^{(k)}_{t+1,a})] = \mathbb{E}_n[\phi_{t,a}\phi_{t,a}^\top]^{-1}\sum_{k=1}^K \mathbb{E}_k[\phi_{t,a}^\top \tilde{Y}_{t,a}(\zeta_t^*, \overline{\theta}^{(k)}_{t+1,a})]$$

$$\hat{\theta}_{t,a} = \mathbb{E}_n[\phi_{t,a}\phi_{t,a}^\top]^{-1}\sum_{k=1}^K \mathbb{E}_k[\phi_{t,a}^\top \tilde{Y}_{t,a}(\zeta_t^{(k)}, \overline{\theta}^{(k)}_{t+1,a})]$$

Note that

$$\sqrt{n}(\hat{\theta}_{t,a} - \theta^*_{t,a}) = \sqrt{n}(\hat{\theta}_{t,a} - \tilde{\theta}_{t,a}) + \sqrt{n}(\tilde{\theta}_{t,a} - \theta^*_{t,a})$$

Orthogonality and cross-fitting in Proposition 7 establish that the first term is $o_p(1)$. The second term includes $\overline{\theta}^{(k)}_{t+1,a}$ as a generated regressor term, and we establish its asymptotic variance by GMM.

Note that

$$\begin{aligned}
\sqrt{n}(\hat{\theta}_{t,a} - \tilde{\theta}_{t,a}) &= \mathbb{E}_n[\phi_{t,a}\phi_{t,a}^\top]^{-1}\sum_{k=1}^K \left\{\mathbb{E}_k[\phi_{t,a}^\top \tilde{Y}_{t,a}(\hat{\zeta}_t^{(k)}, \hat{\overline{\theta}}^{(k)}_{t+1,a})] - \mathbb{E}_k[\phi_{t,a}^\top \tilde{Y}_{t,a}(\zeta_t^*, \overline{\theta}^*_{t+1,a})]\right\} \\
&= \mathbb{E}_n[\phi_{t,a}\phi_{t,a}^\top]^{-1}\sum_{k=1}^K \left\{\mathbb{E}\left[\phi_{t,a}^\top\left(\tilde{Y}_{t,a}(\hat{\zeta}_t^{(k)}, \hat{\overline{\theta}}^{(k)}_{t+1,a}) - \tilde{Y}_{t,a}(\zeta_t^*, \overline{\theta}^*_{t+1,a})\right)\right]\right\} \\
&\quad + \mathbb{E}_n[\phi_{t,a}\phi_{t,a}^\top]^{-1}\sum_{k=1}^K \left\{(\mathbb{E}_k - \mathbb{E})\left[\phi_{t,a}^\top \tilde{Y}_{t,a}(\hat{\zeta}_t^{(k)}, \hat{\overline{\theta}}^{(k)}_{t+1,a}) - \phi_{t,a}^\top \tilde{Y}_{t,a}(\zeta_t^*, \overline{\theta}^*_{t+1,a})\right]\right\}
\end{aligned}$$

We will show the first term is $o_p(n^{-\frac{1}{2}})$ by orthogonality. Define

$$S_{1,k} := \mathbb{E}\left[\phi_{t,a}^\top\left(\tilde{Y}_{t,a}(\hat{\zeta}_t^{(k)}, \hat{\overline{\theta}}^{(k)}_{t+1,a}) - \tilde{Y}_{t,a}(\zeta_t^*, \overline{\theta}^*_{t+1,a})\right)\right]$$

We consider elements of the vector-valued moment condition: for each $j = 1, \ldots, p$ :

$$\begin{aligned}
&= \mathbb{E}\left[(\phi_{t,a}^\top)_j \mathbb{E}\left[\tilde{Y}_{t,a}(\hat{\zeta}_t^{(k)}, \hat{\overline{\theta}}^{(k)}_{t+1,a}) - \tilde{Y}_{t,a}(\zeta_t^*, \overline{\theta}^*_{t+1,a}) \mid S_t, a\right]\right] \\
&\leq \|(\phi_{t,a}^\top)_j\| \|\mathbb{E}[\tilde{Y}_{t,a}(\hat{\zeta}_t^{(k)}, \hat{\overline{\theta}}^{(k)}_{t+1,a}) - \tilde{Y}_{t,a}(\zeta_t^*, \overline{\theta}^*_{t+1,a}) \mid S_t, a]\| \\
&\leq \overline{C}\|\mathbb{E}[\tilde{Y}_{t,a}(\hat{\zeta}_t^{(k)}, \hat{\overline{\theta}}^{(k)}_{t+1,a}) - \tilde{Y}_{t,a}(\zeta_t^*, \overline{\theta}^*_{t+1,a}) \mid S_t, a]\| && \text{by Assumption 15} \\
&= o_p(n^{-\frac{1}{2}}) && \text{by Proposition 7}
\end{aligned}$$

Next we study the sampling/cross-fitting terms:

$$S_{2,k} := \left\{ (\mathbb{E}_k - \mathbb{E}) \left[ \phi_{t,a}^\top \left( \tilde{Y}_{t,a}(\hat{\zeta}_t^{(k)}, \hat{\overline{\theta}}_{t+1,a}^{(k)}) - \tilde{Y}_{t,a}(\zeta_t^*, \overline{\theta}_{t+1,a}^*) \right) \right] \right\}$$

Since $|\mathcal{I}_k| \simeq n/K$, by the concentration of iid terms, by Cauchy-Schwarz inequality, we have that

$$S_{2,k} = o_p \left( n^{-1/2} \sum_{i=1}^p \mathbb{E} \left[ \left( \tilde{Y}_{t,a}(\hat{\zeta}_t^{(k)}, \hat{\overline{\theta}}_{t+1,a}^{(k)}) - \tilde{Y}_{t,a}(\zeta_t^*, \overline{\theta}_{t+1,a}^*) \right)^2 \left( (\phi_{t,a})_j \right)^2 \right]^{1/2} \right)$$

Further,

$$\begin{aligned}
&\sum_{i=1}^p \mathbb{E} \left[ \left( \tilde{Y}_{t,a}(\hat{\zeta}_t^{(k)}, \hat{\overline{\theta}}_{t+1,a}^{(k)}) - \tilde{Y}_{t,a}(\zeta_t^*, \overline{\theta}_{t+1,a}^*) \right)^2 \left( (\phi_{t,a})_j \right)^2 \right]^{1/2} \\
&\leq \overline{C} \| \tilde{Y}_{t,a}(\hat{\zeta}_t^{(k)}, \hat{\overline{\theta}}_{t+1,a}^{(k)}) - \tilde{Y}_{t,a}(\zeta_t^*, \overline{\theta}_{t+1,a}^*) \| \\
&= o_p(1) \qquad \text{by consistency of nuisances}
\end{aligned}$$

Therefore, by continuous mapping theorem, Slutsky's theorem, and Assumption 15, $\sqrt{n}(\hat{\theta}_{t,a} - \tilde{\theta}_{t,a}) = o_p(1)$.

Next we study $\sqrt{n}(\tilde{\theta}_{t,a} - \theta_{t,a}^*)$. One approach for establishing asymptotic variance under generated regressors is via GMM, which we do so in this setting (Newey and McFadden 1994). We can write $\overline{\theta}$ as the parameter vector satisfying the "stacked" moment conditions (over timesteps and actions) at the true quantile parameter $\zeta$ (via our previous orthogonality analysis).

The moment functions for the robust $Q$-function parameters of interest, $\overline{\theta}_{t,\cdot}$, satisfy:

$$\left\{ 0 = \mathbb{E} \left[ \left\{ \tilde{Y}_{t,a}(\zeta_t^*, \overline{\theta}_{t+1}) - \overline{\theta}_{t,a}^\top \phi_{t,a} \right\} \phi_{t,a}^\top \mid A = a \right] \right\}_{a \in \mathcal{A}, t=1,\ldots,T} \tag{22}$$

We let these stacked moments be denoted as $\{0 = \mathbb{E}[g_{t,a}(\zeta^*, \bar{\theta})]\}_{a \in \mathcal{A}, t=0,\ldots,T-1}$.

For GMM, the asymptotic covariance matrix is given by

$$\sqrt{n}(\tilde{\theta}_{t,a} - \theta_{t,a}^*) \xrightarrow{d} -(G'G)^{-1} G' N(0, I) = N(0, V)$$

where $G = \partial g(\zeta^*, \bar{\theta}) / \partial \theta$ and a consistent estimator of the asymptotic variance is given by $\hat{V} = \left( \hat{G}'\hat{G} \right)^{-1}, \hat{G} = \partial \hat{g}(\zeta^*, \bar{\theta}) / \partial \bar{\theta}$.

Note that $G$ is a block upper triangular matrix. The (blockwise) entries on the time diagonal are given by the covariance matrix $\phi_{t,a}\phi_{t,a}^\top$ (i.e., from linear regression). The lower entries, i.e. $\partial g_{t,a}(\zeta^*, \overline{\theta}) / \partial \overline{\theta}_{t+1,a'}$ are given below, by differentiating under the integral:

$$\frac{\partial g_{t,a}(\zeta^*,\overline{\theta})}{\partial \overline{\theta}_{t+1,a'}} \left\{\mathbb{E}\left[\left\{\left(\alpha_{t,a}Y_{t,a}(\overline{\theta}_{t+1}) + (1-\alpha_{t,a})\cdot \tfrac{1}{1-\tau}\left(Y_{t,a}(\overline{\theta}_{t+1})\mathbb{I}(Y_{t,a}(\overline{\theta}_{t+1}) \le \zeta_{t,a}^\top \phi_{t,a}) \right.\right.\right.\right.\right.$$
$$\left.\left.\left.\left.\left. -\zeta_{t,a}^\top\phi_{t,a}\cdot[\mathbb{I}\{Y_{t,a}(\overline{\theta}_{t+1}) \le \zeta_{t,a}^\top\phi_{t,a}\} - (1-\tau)]\right)\right) - \overline{\theta}_{t,a}{}^\top\phi_{t,a}\right\}\phi_{t,a}^\top \mid A=a\right]\right\}$$
$$= \mathbb{E}\left[\alpha_{t,a}(\phi(S_{t+1},a')\phi(S_t,A_t)^\top) + \frac{1-\alpha_{t,a}}{1-\tau}\left(\int_{-\infty}^{q}(\phi(S_{t+1},a_{t+1}))dP_{S_{t+1}|S_t,A_t}\right)\phi^\top(S_t,A_t) \mid A_t = a\right]$$
$$= \mathbb{E}[\alpha_{t,a}(\phi(S_{t+1},a')\phi(S_t,a)^\top)] + \mathbb{E}\left[(1-\alpha_{t,a})(\mathbb{E}[\phi(S_{t+1},a_{t+1}) \mid Y_{t+1} \le \zeta_{t,a}^\top\phi_{t,a}, S_t, A_t = a]\phi^\top(S_t,a)\right]$$

Denote $Z_{a_{t+1}}^{\phi_{t+1}}(S_t,a) = \mathbb{E}[\phi(S_{t+1},a_{t+1}) \mid Y_{t+1} \le \zeta_{t,a}^\top\phi_{t,a}, S_t, A_t = a]$, then

$$\begin{aligned}\frac{\partial g_{t,a}(\zeta^*,\overline{\theta})}{\partial \overline{\theta}_{t+1,a'}} &= \mathbb{E}\left[\alpha_{t,a}(\phi(S_{t+1},a')\phi(S_t,a)^\top) + (1-\alpha_{t,a})(Z_{a'}^{\phi}(S_t,a)\phi^\top(S_t,a))\right]\\ &= \mathbb{E}\left[\alpha_{t,a}(\phi_{t+1,a'}\phi_{t,a}^\top) + (1-\alpha_{t,a})(Z_{a',t,a}^{\phi}\phi_{t,a}^\top)\right]\\ &= \tilde{\Sigma}_{t,a}^{t+1,a'} + \tilde{\Omega}_{t,a}^{a',t,a}\end{aligned}$$

So, $G$ is a block upper triangular matrix:

$$\begin{bmatrix} \ddots & \cdots & & & & \\ 0 & \mathbb{E}[\phi_{t,a}\phi_{t,a}^\top] & \{\tilde{\Sigma}_{t,a}^{t+1,a'} + \tilde{\Omega}_{t,a}^{a',t,a}\}_{a'\in\mathcal{A}} & & & \\ 0 & 0 & \ddots & \{\tilde{\Sigma}_{t,a_k}^{t+1,a'} + \tilde{\Omega}_{t,a_k}^{a',t,a_k}\}_{a'\in\mathcal{A}} & & \\ 0 & 0 & 0 & \mathbb{E}[\phi_{t,a_K}\phi_{t,a_K}^\top] & \{\tilde{\Sigma}_{t,a_K}^{t+1,a'} + \tilde{\Omega}_{t,a_K}^{a',t,a_K}\}_{a'\in\mathcal{A}} \end{bmatrix}$$

| | $\Lambda = 1$ | $\Lambda = 2$ | $\Lambda = 5.25$ | $\Lambda = 8.5$ | $\Lambda = 11.75$ | $\Lambda = 15$ |
|---|---|---|---|---|---|---|
| Method | FQI | Non-orth/Orth | Non-orth/Orth | Non-orth/Orth | Non-orth/Orth | Non-orth/Orth |
| MSE($\bar{V}_0^*$) | 0.2 | 0.5 / 0.5 | 3.2 / 1.7 | 7.7 / 2.7 | 15.2 / 3.4 | 30.2 / 3.8 |
| $\ell_2$ Param. Error | 3.4 | 4.1 / 3.5 | 11.5 / 3.9 | 24.0 / 3.9 | 48.9 / 3.7 | 88.0 / 3.5 |
| % Wrong Action | 28% | 31% / 28% | 43% / 31% | 45% / 31% | 47% / 31% | 48% / 30% |

**Table 3 Simulation results with $d = 100$ and $n = 600$, reporting the value function MSE, Q function parameter error, and the portion of the time a sub-optimal action is taken. Each cell shows Non-Orthogonal / Orthogonal results for each $\Lambda$.**

## Appendix 8: Details on experiments

### 8.1. Simulation (evaluation)

*Additional high-dimensional simulated experiments* The results for the high-dimensional setting are in Table 3. In this setting, policy optimization is substantially harder — even the nominal policy estimate only picks the true optimal action 72% of the time. However, we still see almost identical behavior as in the low-dimensional setting when comparing the orthogonal and non-orthogonal estimators. Without orthogonalization, performance drops off dramatically as $\Lambda$ increases, such that for $\Lambda = 15$, the policy is only slightly better than random choice. Our orthogonalized algorithm has MSE that decays more gracefully with $\Lambda$, and picks the correct action at essentially the same rate as the nominal algorithm, even as $\Lambda$ increases.

*Low-Dimensional Parameter Values* $\theta_A = -0.05$, $\sigma = 0.36$, $\gamma = 0.9$, $H = 4$.

The matrices $A$ and $B$ were chosen randomly with a fixed random seed:

```
np.random.seed(1)
B_sparse0 = np.random.binomial(1,0.3,size=d)
B = 2.2*B_sparse0 * np.array( [ [ 1/(j+k+1) for j in range(d) ]
                                             for k in range(d) ] )


np.random.seed(2)
A_sparse0 = np.random.binomial(1,0.6,size=d)
A = 0.48*A_sparse0 * np.array( [ [ 1/(j+k+10) for j in range(d) ]
                                               for k in range(d) ] )
```

Likewise for $\theta_R$:

```
theta_R = 3 * np.random.normal(size=d) * np.random.binomial(1,0.3,size=d)
```

*High-Dimensional Parameter Values* $\theta_A = -0.05$, $\sigma = 0.1$, $\gamma = 0.9$, $H = 4$.

The matrices $A$ and $B$ were chosen randomly with a fixed random seed:

```
np.random.seed(1)
B_sparse0 = np.random.binomial(1,0.3,size=d)
B = 2.2*B_sparse0 * np.array( [ [ 1/(j+k+1) for j in range(d) ]
                                 for k in range(d) ] )/1.2


np.random.seed(2)
A_sparse0 = np.random.binomial(1,0.6,size=d)
A = 0.48*A_sparse0 * np.array( [ [ 1/(j+k+10) for j in range(d) ]
                                  for k in range(d) ] )/20
```

Likewise for $\theta_R$:

```
theta_R = 2 * np.random.normal(size=d) * np.random.binomial(1,0.3,size=d)
```

*Function Approximation* Conditional expectations were approximated with the Lasso using `scikit-learn`'s implementation, with regularization hyperparameter $\alpha$ = 1e-4. Conditional quantiles were approximated with `scikit-learn`'s $\ell_1$-penalized quantile regression, regularization hyperparameter $alpha$ = 1e-2, using the `highs` solver.

*Calculating Ground Truth* To provide ground truth for our sparse linear setting, we analytically derive the form of the robust Bellman operator. Consider the candidate $Q$ function, $Q(s,0) = \beta^\top s + a_0, Q(s,1) = \beta^\top s + a_1$. Then,

$$\begin{aligned} Y_t &= \theta_R^\top S_{t+1} + \gamma\beta^\top S_{t+1} + \theta_A \gamma \max\{1_d^\top \theta_R, 0\} \\ &= \theta_R^\top S_{t+1} + \gamma\beta^\top S_{t+1} + \theta_A \gamma 1_d^\top \theta_R \end{aligned}$$

where we chose simulation parameters such that $\theta_A \gamma \max\{1_d^\top \theta_R, 0\} > 0$. Therefore:

$$Y_t | S_t, A_t \sim \mathcal{N}\left( (\theta_R + \gamma\beta)^\top (BS_t + \theta_A A_t) + \theta_A \gamma 1_d^\top \theta_R, \sqrt{\sum_{i=1}^d (\theta_R + \gamma\beta)_i^2 (AS_t + \sigma)_i^2} \right)$$

Since $Y_t$ is conditionally Gaussian, we apply Lemma 7:

$$\begin{aligned} (\bar{\mathcal{T}}_t^* Q)(s,a) &= \mathbb{E}[Y_t | S_t = s, A_t = a] - 0.5C(\Lambda)\sqrt{\text{Var}[Y_t | S_t = s, A_t = a]} \\ &= (\theta_R + \gamma\beta)^\top (BS_t + \theta_A A_t) + \theta_A \gamma 1_d^\top \theta_R - 0.5C(\Lambda)\sqrt{\sum_{i=1}^d (\theta_R + \gamma\beta)_i^2 (AS_t + \sigma)_i^2} \end{aligned}$$

First, note that the slope w.r.t. $S_t$ is not a function of $A_t$ validating our choice of an action-independent $\beta$. Second, note that only the last term is non-linear in $S_t$. So the ground truth for FQI with Lasso adds the first two terms to the closest linear approximation of this last term. Since our object of interest is the average optimal value function at the initial state, we perform this linear approximation in terms of mean squared error at the initial state. In practice, we compute this by drawing $200,000$ samples i.i.d. from the initial state distribution and then doing linear regression on this last term. Plugging the slope and intercept back in is extremely close to the best linear approximation of $(\bar{\mathcal{T}}_t^* Q)(s,a)$.

### 8.2. Policy learning simulation: Healthcare-inspired

*State and Action Spaces* The state at time $t$ is a tuple $(L_t, S_t, U_t)$ and the action is $A_t$.

- **Latent Health State:** $L_t \in \{0,1,2,3,4,5\}$
- **Continuous State:** $S_t \in \mathbb{R}^4$
- **Unobserved Confounder:** $U_t \in \{-1,1\}$
- **Action Space:** $A_t \in \{0,1,2\}$, corresponding to "Do Nothing," "Low Drug," and "High Drug."

The confounder $U_t$ can be generated in two ways, based on the simulation parameters.

**State-dependent generation, $S_t \to U_t$.**

The probability of a favorable confounder ($U_t = 1$) depends on the first component of the continuous state, $S_{t,1}$.

$$p(U_t = 1|S_t) = \text{clip}\left(\left(p_{\text{min}} + (p_{\text{max}} - p_{\text{min}})\frac{1.5 - |S_{t,1}|}{1.5}\right) + \mathcal{N}(0, \sigma^2_{\text{noise}}), c_{\text{min}}, c_{\text{max}}\right) \tag{23}$$

where $p_{\text{min}}, p_{\text{max}}, \sigma^2_{\text{noise}}, c_{\text{min}}, c_{\text{max}}$ are fixed scalar parameters.

**Autoregressive generation, $U_{t-1} \to U_t$.**

The probability is a blend of the previous confounder's value and the IID probability.

$$p(U_t = 1|S_t, U_{t-1}) = \text{clip}\left(\rho \cdot \frac{U_{t-1}+1}{2} + (1-\rho)\cdot p_{\text{base}}(S_t) + \mathcal{N}(0, \sigma^2_{\text{noise}}), c_{\text{min}}, c_{\text{max}}\right) \tag{24}$$

where $\rho$ is the autoregressive coefficient and $p_{\text{base}}(S_t)$ is the un-noised probability from the IID case.

*State Transition Dynamics*

*Latent Health State ($L_{t+1}$)* The transition from $L_t$ to $L_{t+1}$ is determined by a random logit vector, $\mathbf{Z}_t \in \mathbb{R}^3$, corresponding to (improve, stay, worsen). The probability of each outcome is $\text{softmax}(\mathbf{Z}_t)$. The logits are calculated as:

$$\mathbf{Z}_t = \mathbf{b}_{A_t,U_t} + \mathbf{w}_{A_t}(S_t - \mathbf{c}_{L_t}) + \beta_{L_t} \tag{25}$$

The components are:

- $A_t = 0$ (Do Nothing):

$$\mathbf{b}_{0,U_t} = \begin{cases} (0.10 + 2\delta, 0.30, 0.60 - 2\delta)^T & \text{if } U_t = 1 \text{ (Favorable)} \\ (0.10 - 1\delta, 0.30, 0.60 + 1\delta)^T & \text{if } U_t = -1 \text{ (Unfavorable)} \end{cases}$$

- $A_t = 1$ (Low Drug):

$$\mathbf{b}_{1,U_t} = \begin{cases} (0.40 + 1\delta, 0.40, 0.20 - 1\delta)^T & \text{if } U_t = 1 \text{ (Favorable)} \\ (0.40 - 2\delta, 0.40, 0.20 + 2\delta)^T & \text{if } U_t = -1 \text{ (Unfavorable)} \end{cases}$$

- $A_t = 2$ (High Drug):

$$\mathbf{b}_{2,U_t} = \begin{cases} (0.02 + 4\delta, 0.03, 0.95 - 4\delta)^T & \text{if } U_t = 1 \text{ (Favorable)} \\ (0.02 - 6\delta, 0.03, 0.95 + 6\delta)^T & \text{if } U_t = -1 \text{ (Unfavorable)} \end{cases}$$

This transition may be overridden by *random deterioration* (with 4% probability, $L_{t+1} = \min(L_t + 1, 5)$) or *time-dependent deterioration*, for $t > 6$, $L_{t+1} = \min(L_t + 1, 5)$ with probability $0.02 \cdot (t - 6)$.

*Continuous State evolution ($S_{t+1}$)* To generate contextual transitions that still reflect the dynamics from the underlying latent state, we generate continuous states via a hybrid approach. The continuous state evolves via an $\alpha$-weighted mixture of a vector autoregressive (VAR) model and certain "contextual center vectors" $c_{L_{t+1}}$, one for each value of the latent state $L_{t+1}$.

$$S_{t+1} = (1 - \alpha)\mathbf{b}S_t + \alpha\mathbf{c}_{L_{t+1}} + \mathcal{E}_t \tag{26}$$

where $\mathbf{b} \in \mathbb{R}^{4\times4}$ is a stable autoregressive parameter matrix, $\alpha \in (0, 1)$ is a scalar blending parameter, $\mathbf{c}_{L_{t+1}}$ is the parameter vector corresponding to the next latent health state, $\mathcal{E}_t \sim \mathcal{N}(\mathbf{0}, \sigma^2_{t+1}\mathbf{I})$ is a random noise vector, where the variance $\sigma^2_{t+1}$ is a parameter that increases for worse health states and more aggressive actions.

- $\mathbf{b} \in \mathbb{R}^{4\times4}$ **(Autoregressive Matrix)**:with elements drawn from a standard normal distribution, then normalized by $1.1\times$ its spectral radius to ensure stability.

- $\mathcal{E}_t$ **(Noise Vector)**: A random noise vector drawn from a normal distribution $\mathcal{N}(\mathbf{0}, \sigma^2_{t+1}\mathbf{I})$, where the variance $\sigma^2_{t+1}$ increases with more aggressive actions and worse health states.
- $\mathbf{c}_{L_t}$ **(Center Vectors)**: These are the fixed 4-dimensional parameter vectors corresponding to each latent health state $L_t$.

$$\begin{aligned}
\mathbf{c}_0 \text{ (Healthiest)} &= [\ 0.0, -0.0,\ 0.0, -0.0]^T \\
\mathbf{c}_1 \text{ (Stable)} &= [\ 0.5, -0.5,\ 0.5, -0.5]^T \\
\mathbf{c}_2 \text{ (Unstable)} &= [\ 1.0, -1.0,\ 1.0, -1.0]^T \\
\mathbf{c}_3 \text{ (Serious)} &= [\ 1.5, -1.5,\ 1.5, -1.5]^T \\
\mathbf{c}_4 \text{ (Critical)} &= [\ 2.0, -2.0,\ 2.0, -2.0]^T \\
\mathbf{c}_5 \text{ (Mortality)} &= [-3.0, -3.0, -3.0, -3.0]^T
\end{aligned}$$

*Reward Function* The reward $R_t$ is a random variable calculated as:

$$R_t = r(L_t, A_t, U_t) = r_{\text{base}}(L_t) - c(A_t, L_t) - p(A_t, L_t) - h(L_t) + b(A_t, L_t, U_t) \tag{27}$$

where the component functions depend on fixed parameters:

- Base Reward $r_{base}(L_t)$: Values are $15, 12, 9, 4, 2, -5$ for $L_t \in 0, \ldots, 5$, respectively.
- Action Cost $c(A_t, L_t)$ : A fraction of the base reward, with costs of $\{0\%, 5\%, 50\%\}$ of $r_{\text{base}}(L_t)$ for actions $A_t \in \{0, 1, 2\}$, respectively.
- Risk Penalty $p(A_t, L_t)$ :

$$p(A_t, L_t) = -0.1\mathbb{I}[A_t = 2] \times r_{\text{base}}(L_t)$$

- Health State Penalty $h(L_t)$: A fraction of the base reward, with penalties of $\{5\%, 10\%, 20\%\}$ of $r_{\text{base}}(L_t)$ for states $L_t \in \{2, 3, 4\}$, respectively.
- Confounder Effect $b(A_t, L_t, U_t)$: Action-confounder dependent change.

$$\begin{aligned}
b(A_t, L_t, U_t) &= r_{\text{base}}(L_t) \times \\
&(\mathbb{I}[U_t = 1]\{0.15\mathbb{I}[A_t = 1] + 0.6\mathbb{I}[A_t = 2]\} \cdot r_{\text{base}}(L_t) + \mathbb{I}[U_t = -1]\{-0.25\mathbb{I}[A_t = 1] - 0.7\mathbb{I}[A_t = 2]\})
\end{aligned}$$

### 8.3. MIMIC-III case study

**8.3.1. Calibrating** $\Lambda$ See Figure 15 for a plot of odds-ratio values obtained by dropping each covariate. (Note that we use a preprocessing of (Killian et al. 2020), so that features are actually dimensions of a representation, and therefore not inherently interpretable.). The 90% quantile of the lower bound on $\Lambda$ is given by $\Lambda = 1.42$, and the 99% quantile is given by $\Lambda = 2.48$.

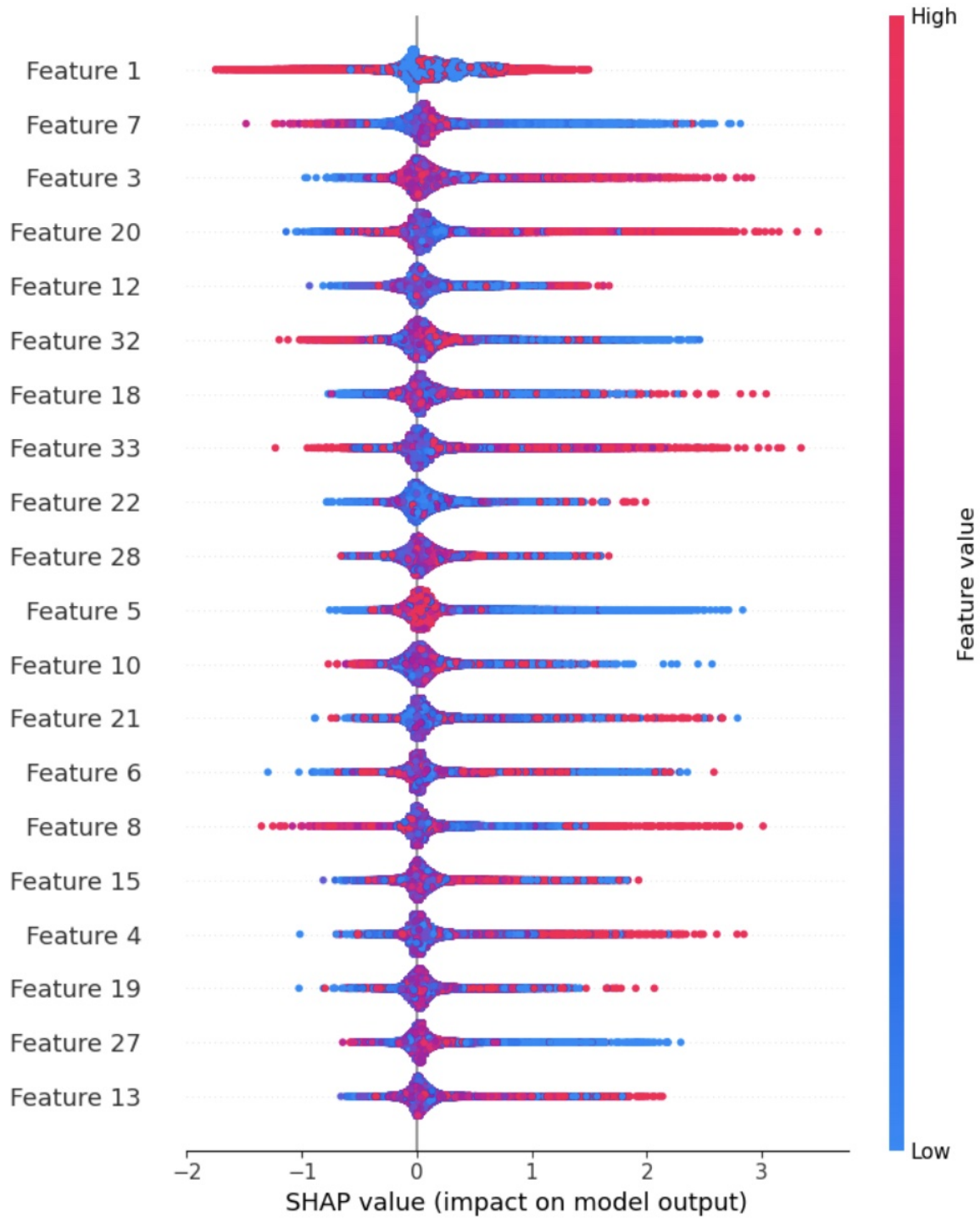

**Figure 15** **Distribution plots of odds-ratios obtained by dropping each covariate – obtained via SHAP package.**

#### 8.3.2. Additional analysis

### 8.4. LSVI-UCB Warmstarting Hyperparameter Choices

In this section, we provide additional experimental results, varying the LSVI-UCB hyperparameter, $\xi$. In general, we would like $\xi$ to be as small as possible so that we switch from exploring to exploiting the optimal arm quickly. However, for standard LSVI-UCB, if $\xi$ gets too small, then we no longer have a valid upper confidence bound, and the algorithm can get stuck on a sub-optimal arm, resulting in possibly linear regret. You can see this trend in Figure 16, which compares standard LSVI-UCB and our robust warm-started LSVI-UCB for various values of $\xi$. The "No Warm-start" plots achieve their optimal performance at around $\xi = 0.1$ or $0.15$. At $\xi = 0.2$, the intervals are slightly wider than necessary and regret increases. But as $\xi$ gets smaller than $0.1$, the intervals start to become too small, and average regret steadily becomes linear.

By contrast, the robust bounds from the offline data are always valid, and so the smallest values of $\xi = 0.02$ and $0.05$ tend to perform the best. Regret increases for the higher values of $\xi$, $0.1, 0.15, 0.2$. In Figure 17, we perform an experiment where we tune the hyperparameters for the two procedures separately, choosing $\xi = 0.1$ for No Warm-start and $\xi = 0.05$ for Warm-start, so that we can compare the best achievable performance of the two algorithms.

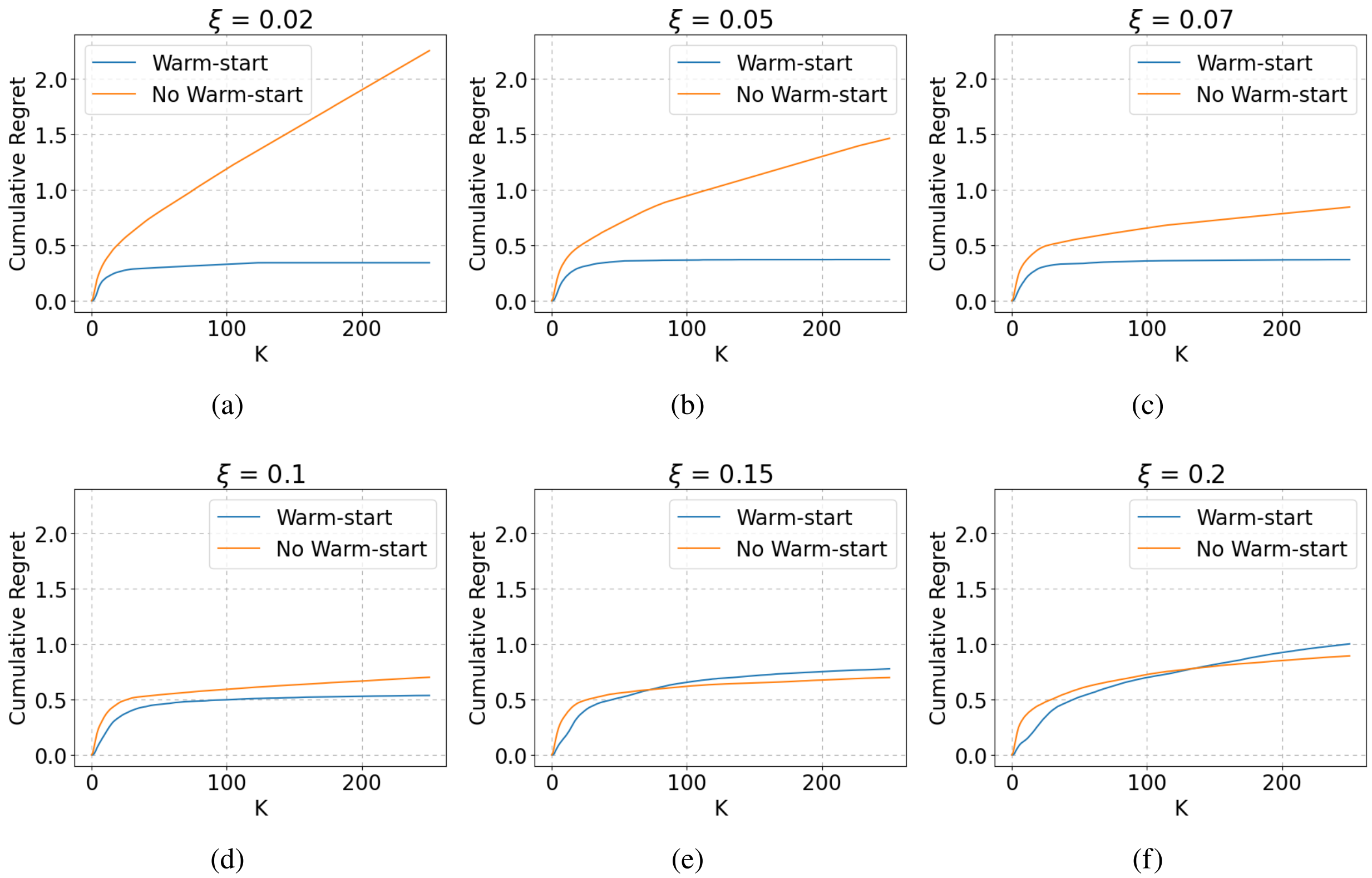

**Figure 16** **We repeat the LSVI-UCB simulations for various values of $\xi$. All cumulative regrets are an average of over $200$ trials.**

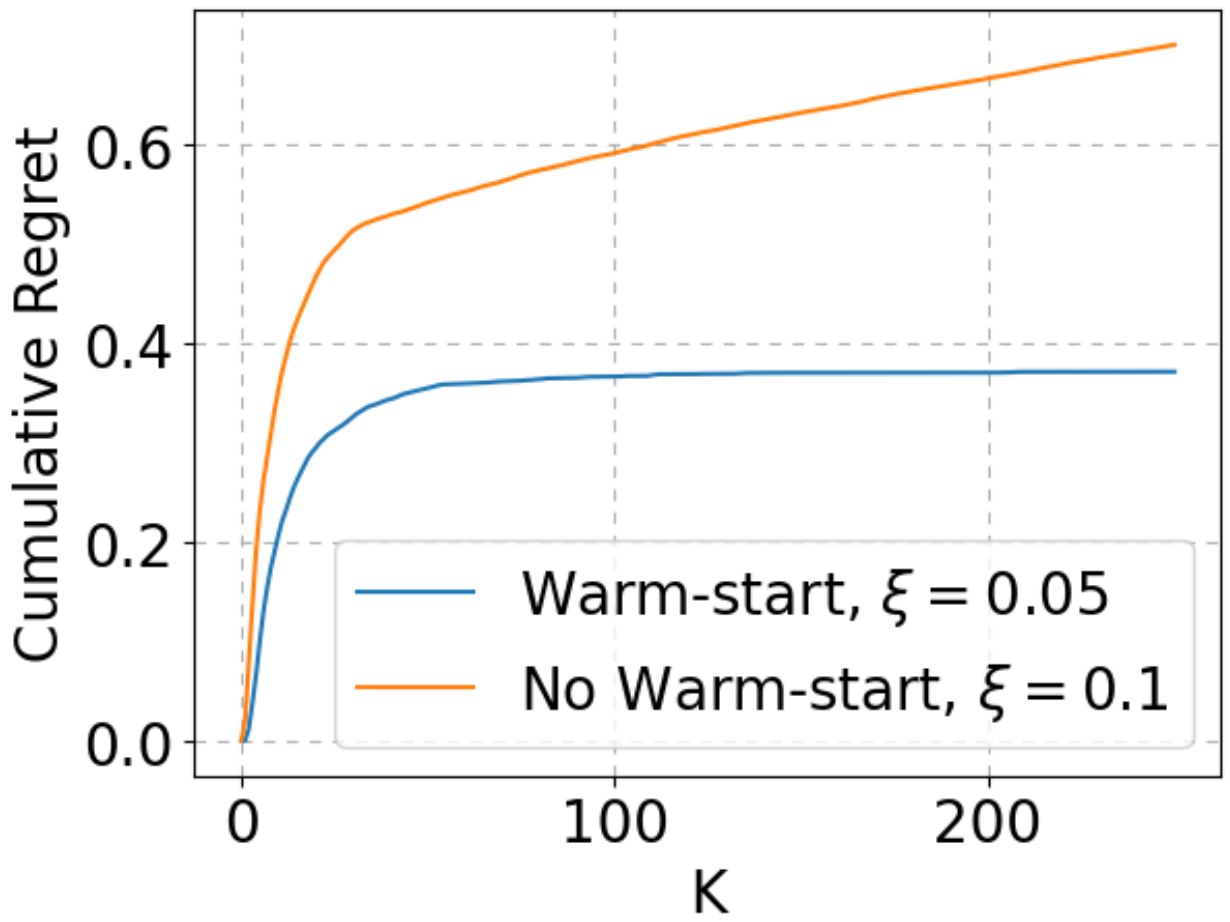

**Figure 17** **Comparison of LSVI-UCB with and without warm-starting for the best values of $\xi$, chosen from $\{0.02, 0.05, 0.07, 0.1, 0.15, 0.2\}$. Cumulative regrets is an average of over $200$ trials.**

# Acknowledgments